\documentclass{article}

\usepackage{microtype}
\usepackage{graphicx}
\usepackage{subfigure}
\usepackage{booktabs} % for professional tables
\usepackage{multirow}
\usepackage{enumitem}
\usepackage{float}
\usepackage{placeins}     % Add this line
\usepackage{hyperref}
\usepackage{chemfig}
\usepackage{xcolor}
\usepackage{amsmath}

\usepackage{wrapfig}

\newcommand{\spacedhline}{\noalign{\vskip 1.1pt}\hline\noalign{\vskip 1.1pt}}
\usepackage[accepted]{icml2024}

\usepackage{amsmath}
\usepackage{amssymb}
\usepackage{mathtools}
\usepackage{amsthm}

\usepackage[capitalize,noabbrev]{cleveref}
\usepackage{tablefootnote}
\theoremstyle{plain}

\theoremstyle{definition}

\theoremstyle{remark}

\usepackage[textsize=tiny]{todonotes}

\icmltitlerunning{\textsc{GraphBPE}: Molecular Graphs Meet Byte-Pair Encoding}

\begin{document}

\twocolumn[
\icmltitle{\textsc{GraphBPE}: Molecular Graphs Meet Byte-Pair Encoding}




\begin{icmlauthorlist}
\icmlauthor{Yuchen Shen}{lti}
\icmlauthor{Barnabás Póczos}{mld}

\end{icmlauthorlist}

\icmlaffiliation{lti}{Language Technologies Institute, Carnegie Mellon University}
\icmlaffiliation{mld}{Machine Learning Department, Carnegie Mellon University}

\icmlcorrespondingauthor{Barnabás Póczos}{bapoczos@cs.cmu.edu}

\icmlkeywords{Machine Learning, ICML}

\vskip 0.3in
]



\printAffiliationsAndNotice{}  % leave blank if no need to mention equal contribution

\begin{abstract}
With the increasing attention to molecular machine learning, various innovations have been made in designing better models or proposing more comprehensive benchmarks. However, less is studied on the data preprocessing schedule for molecular graphs, where a different view of the molecular graph could potentially boost the model's performance. Inspired by the Byte-Pair Encoding (BPE) algorithm, a subword tokenization method popularly adopted in Natural Language Processing, we propose \textsc{GraphBPE}, which tokenizes a molecular graph into different substructures and acts as a preprocessing schedule independent of the model architectures. Our experiments on 3 graph-level classification and 3 graph-level regression datasets show that data preprocessing could boost the performance of models for molecular graphs, and \textsc{GraphBPE} is effective for small classification datasets and it performs on par with other tokenization methods across different model architectures. 
\end{abstract}
\section{Introduction}
Tokenization~\cite{sennrich2016neural, Schuster2012JapaneseAK, kudo-2018-subword, kudo-richardson-2018-sentencepiece} is an important building block that contributes to the success of modern Natural Language Processing (NLP) applications such as Large Language Models (LLMs)~\cite{brown2020language, touvron2023llama, almazrouei2023falcon, touvron2023llama2}. Before being fed into a model, each word in the input sentence is first tokenized into subwords (e.g., $\text{``lowest''}\rightarrow\text{``low'', ``est''}$), which may not necessarily convey meaningful semantics but facilitates the learning of the model. Among different tokenization methods, Byte-Pair Encoding (BPE)~\cite{Gage1994ANA, sennrich2016neural} is a popularly adopted mechanism. Given a text corpus containing numerous sentences and thus words, BPE counts the appearance of two consecutive tokens (e.g., a subword ``es'', an English letter ``t'') in each word at each iteration, and merges the token pair with the highest frequency and treats it as the new token (e.g., $\text{``es'', ``t''}\rightarrow\text{``est''}$) for next round. A vocabulary containing a variety of subwords is then learned after some iterations, and later used to tokenize sentences fed to the model.

It is easy to observe that this ``\textbf{count-and-merge}'' schedule has the potential to generalize beyond texts into arbitrary structures such as molecular graphs. Indeed, we can view words as line graphs, where each character in the word is the node, and the edges are defined by whether two characters are contiguous in the word. This observation naturally motivates us to explore the following questions: a). ``\textit{Can graphs be tokenized similarly to that of texts?}'' b). ``\textit{Will the tokenized graphs improve the model performance?}''

To investigate whether molecular graphs can be tokenized similarly to texts, we develop \textsc{GraphBPE}, a variant of the BPE algorithm for molecular graphs, which counts the co-occurrence of contextualized (e.g., neighborhood-aware) node pairs (e.g., defined by edges) and merges the most frequent pair as the new node for next round. Compared with other methods~\cite{jin2020hierarchical, 10.1093/bib/bbad398} that require external knowledge (e.g., functional groups, a trained neural network) to mine substructures, our algorithm relies solely on a given molecular graph corpus and is model agnostic. After each round of tokenization, the resulting new graph is still connected with its nodes being subsets of the nodes of the previous graph, which provides a view to construct both simple graphs and hypergraphs (Section~\ref{sec: hypergraph}) that can be used by Graph Neural Networks (GNNs)~\cite{kipf2017semisupervised, veličković2018graph, xu2019powerful, hamilton2018inductive} and Hypergraph Neural Networks (HyperGNNs)~\cite{feng2019hypergraph, bai2020hypergraph, dong2020hnhn, 9795251}.

To explore whether tokenization helps with model performance, we compare \textsc{GraphBPE} with other tokenization methods on various datasets with different types of GNNs and HyperGNNs. We observe that tokenization in general helps across different model architectures, however, there exists no tokenization method that performs universally well over different datasets, models, and configurations. Our \textsc{GraphBPE} algorithm tends to provide more improvements on smaller datasets with a fixed number of tokenization steps (i.e., 100), as the structures to be learned are proportional to the size of the datasets; thus, larger datasets might need more tokenization steps to observe significant performance boost compared to no tokenization. We summarize our contribution as follows.
\begin{itemize}
    \item We proposed \textsc{GraphBPE}, an iterative tokenization method for molecular graphs that requires no external knowledge and is agnostic to any model architectures, which provides a view of the original graph to construct a new (simple) graph or a hypergraph that can be used by both GNNs and HyperGNNs.
    \item We compare \textsc{GraphBPE} to different graph tokenization methods on six datasets for both classification and regression tasks. The experiment results show that tokenization will affect the performance of both GNNs and HyperGNNs, and \textsc{GraphBPE} can boost the performance on small datasets for different architectures, while performing on par with other tokenization methods on larger datasets.
\end{itemize}
\section{Related Work}
\textbf{Graph tokenization} The idea of graph tokenization is similar to frequency subgraph mining~\cite{10.5555/3000292.3000298, 989534, DBLP:conf/icdm/HeS07, Ranu2009GraphSigAS}, and is popularly explored in molecular generation, where a set of rules is learned to generate novel molecules. Specifically, \citet{kong2022molecule} use BPE to tokenize graphs and develop Principal Subgraph Extraction (PSE), which learns a vocabulary for novel molecule generation. Similar to \citet{kong2022molecule}, \citet{geng2023novo} focus on de nove molecule generation and propose connection-aware vocabulary extraction. Instead of relying on the statistics of substructures, \citet{guo2022dataefficient, lee2024drug} use neural networks to learn tokenization rules for molecule generation. Compared with~\citet{kong2022molecule, geng2023novo}, our algorithm is context-aware; thus by modifying the contextualizer, we can tokenize graphs more flexibly.

\textbf{Substructures for molecular machine learning} Explicitly modeling substructures has shown promising results \cite{yu2022molecular, luong2023fragmentbased, liu2024rethinking} for molecular representation learning. \citet{yu2022molecular} model both molecular nodes and motif nodes to learn good representations. Similarly, \citet{luong2023fragmentbased} use PSE to extract substructures that are later encoded by a fragment encoder for molecular graph pre-training and finetuning, together with another encoder that embeds regular molecular graphs. \citet{liu2024rethinking} discuss different types of graph tokenizers and propose SimSGT, which uses a simple GNN-based tokenizer to help pre-training on molecules.
\section{Preliminary}
In this section, we introduce the Byte-Pair Encoding~\cite{Gage1994ANA, sennrich2016neural} algorithm, which is widely used for NLP tasks, and the notion of hypergraphs.
\subsection{Byte-Pair Encoding} 
Byte-Pair Encoding (BPE) is first developed by \citet{Gage1994ANA} as a data compression technique, where the most frequent byte pair is replaced with an unused ``placeholder'' byte in an iterative fashion. \citet{sennrich2016neural} introduce BPE for machine translation, which improves the translation quality by representing rare and unseen words with subwords from a vocabulary produced by BPE. 

The core of BPE can be summarized as a ``\textbf{count-and-merge}'' paradigm. Starting from a character-level vocabulary derived from a given corpus, it \textbf{counts} the co-occurrence of two contiguous tokens\footnote{Token here refers to a character, a subword, or a word.}, and \textbf{merges} the most frequent pair into a new token. Such a process is carried out iteratively until a desired vocabulary size is reached or there are no tokens to be merged\footnote{It means the corpus is effectively compressed, with the size of the vocabulary equal to the number of unique words in the corpus.}.

An example of BPE on the corpus \{``low'', ``low'', ``lowest'', ``widest''\} is shown in Table~\ref{tab: bpe_example}, where at each round the most frequent contiguous pair is merged into a new token for next round. Note that BPE is \textit{order-sensitive}, meaning the definition of contiguity is always left-to-right, and such an order is preserved for the tokens (e.g., ``l'' and ``o'' are merged and continue to appear as ``lo'' instead of ``ol'').

\begin{table}[h]
\centering
\begin{tabular}{crrc}
\hline
\multicolumn{1}{c}{\textbf{corpus}} & \multicolumn{1}{c}{low$\times 2$} & \multicolumn{1}{c}{lowest} & \multicolumn{1}{l}{widest} \\ 
\hline
\textbf{count}$_1$ & \multicolumn{3}{c}{\{`lo'$\times 3$, `ow'$\times 3$, `es'$\times 2$ ...\}} \\ 
\textbf{merge}$_1$ & \multicolumn{1}{c}{\textcolor[HTML]{FC819E}{\textbf{lo}}w$\times 2$} & \multicolumn{1}{c}{\textcolor[HTML]{FC819E}{\textbf{lo}}west} & \multicolumn{1}{l}{widest} \\ 
\spacedhline
\textbf{count}$_2$ & \multicolumn{3}{c}{\{`\textcolor[HTML]{FC819E}{\textbf{lo}}w'$\times 3$, `es'$\times 3$, `st'$\times 2$ ...\}} \\ 
\textbf{merge}$_2$ & \multicolumn{1}{c}{\textcolor[HTML]{41C9E2}{\textbf{low}}$\times 2$} & \multicolumn{1}{c}{\textcolor[HTML]{41C9E2}{\textbf{low}}est} & \multicolumn{1}{l}{widest} \\
\multicolumn{1}{c}{\rotatebox{90}{...}} & \multicolumn{3}{c}{\rotatebox{90}{$\sim$}}\\
\textbf{count}$_8$ & \multicolumn{3}{c}{\{`\textcolor[HTML]{54B435}{\textbf{wid}}\textcolor[HTML]{C0D6E8}{\textbf{est}}'$\times 1$\}} \\ 
\textbf{merge}$_8$ & \multicolumn{1}{c}{\textcolor[HTML]{41C9E2}{\textbf{low}}$\times 2$} & \multicolumn{1}{c}{\textcolor[HTML]{E9C874}{\textbf{lowest}}} & \multicolumn{1}{l}{\textcolor[HTML]{E5B8F4}{\textbf{widest}}} \\
\hline
\end{tabular}
\caption{A example of BPE with the most frequent token pairs at each round \textbf{bold} \textcolor[HTML]{FC819E}{\textbf{c}}\textcolor[HTML]{41C9E2}{\textbf{o}}\textcolor[HTML]{54B435}{\textbf{l}}\textcolor[HTML]{C0D6E8}{\textbf{o}}\textcolor[HTML]{E9C874}{\textbf{r}}\textcolor[HTML]{E5B8F4}{\textbf{ed}}. After the 8-th round there are no pairs to be merged and every word in the corpus is efficiently compressed.}
\label{tab: bpe_example}
\end{table}

\subsection{Hypergraph}\label{sec: hypergraph}
Compared with a $N$-node simple graph $G=(V, E)$, with $V=\{v_1, v_2, ..., v_N\}$ and $E\subseteq V\times V$ denoting the vertex set and edge set, $\mathcal{N}(v)$ representing the 1-hop neighbors of $v$, a $N$-node $M$-hyperedge hypergraph is defined as $G_h=(V, \mathcal{E}, W)$, including a vertex set $|V|=N$, a hyperedge set $|\mathcal{E}|=M$, and a diagonal weight matrix $W\in\mathbb{R}^{M\times M}$ with $W_{mm}$ for hyperedge $\mathcal{E}_m$. The hypergraph $G_h$ can be represented by a incident matrix $\mathcal{H}\in\mathbb{R}^{N\times M}$, where
\begin{align}
    \mathcal{H}_{nm}=\begin{cases}
        1 & \text{if } v_n\in \mathcal{E}_m \\
        0 & \text{otherwise}\\
    \end{cases}.
\end{align}

Hypergraphs are natural in citation or co-authorship networks, where all the documents cited by a document or co-authored by an author are in one hyperedge. For other domains where the hyperedge relation is less explicit, one can construct the hyperedge around a node with its 1-hop neighbors~\cite{feng2019hypergraph}, or use external domain knowledge~\cite{jin2020hierarchical, 10.1093/bib/bbad398}.

\section{\textsc{GraphBPE}} In this section, we motivate our algorithm by showing a performance boost via ring contraction compared with no tokenization on molecules, followed by the details of the proposed \textsc{GraphBPE} tokenization algorithm.
\subsection{A Motivating Example}
To show that tokenization can potentially yield better performance for molecules, we compare the performance of GNNs learned on the original molecules and tokenized ones. Specifically, we contract rings in the original molecules into hypernodes\footnote{The connectivity of the tokenized graphs are specified by our algorithm in Section \ref{sec: algorithm}} (e.g., a benzene ring is viewed as 1 hypernode instead of 6 carbons), and use the summation \cite{xu2019powerful} of the node features within a hypernode as its representation to be fed into GNNs.

We evaluate on two graph-level tasks, with \textsc{Mutag}~\cite{morris2020tudataset} for classification and \textsc{Freesolv}~\cite{wu2018moleculenet} for regression, and choose GCN~\cite{kipf2017semisupervised}, GAT~\cite{veličković2018graph}, GIN~\cite{xu2019powerful}, and GraphSAGE~\cite{hamilton2018inductive} as the GNNs, with the implementations detailed in Appendix~\ref{app: implementation}.
\vskip -0.1in
\begin{table}[h!]
    \centering
    \resizebox{0.49\textwidth}{!}{%
    \begin{tabular}{lcccc}
  \hline
   \textbf{dataset}        & \textbf{GCN}              & \textbf{GAT}              & \textbf{GIN}              & \textbf{GraphSAGE}        \\
  \hline
   \textsc{Mutag}   & \(0.649_{\pm 0.089}\) & \(0.589_{\pm 0.052}\) & \(0.703_{\pm 0.068}\) & \(0.568_{\pm 0.000}\) \\
    \,\, w. $\scalebox{0.15}{\chemfig{*6(=-=-=-)}}\rightarrow\bullet$                & \(\textbf{0.724}_{\pm 0.052}\) & \(\textbf{0.697}_{\pm 0.069}\) & \(\textbf{0.730}_{\pm 0.017}\) & \(\textbf{0.692}_{\pm 0.041}\) \\
    \spacedhline
  \textsc{Freesolv} & \(4.237_{\pm 0.087}\) & \(4.263_{\pm 0.114}\) & \(4.231_{\pm 0.055}\) & \(4.231_{\pm 0.109}\) \\
   \,\, w. $\scalebox{0.15}{\chemfig{*6(=-=-=-)}}\rightarrow\bullet$                & \(\textbf{4.168}_{\pm 0.030}\) & \(\textbf{4.142}_{\pm 0.065}\) & \(\textbf{4.108}_{\pm 0.069}\) & \(\textbf{4.173}_{\pm 0.046}\) \\
  \hline
  \end{tabular}%
}
    \caption{The performance of different models on \textsc{Mutag} (classification, with accuracy $\uparrow$) and \textsc{Freesolv} (regression, with RMSE $\downarrow$). ``w. $\scalebox{0.15}{\chemfig{*6(=-=-=-)}}\rightarrow\bullet$'' means we contract rings (e.g., a benzene ring) into hypernodes. We report the mean $\mu$ and standard deviation $\sigma$ over 5 random runs as $\mu_{\pm \sigma}$ and the best results are \textbf{bold}.}
    \label{tab:motivating_example}
\end{table}
% \vskip -0.1in
As the results shown in Table~\ref{tab:motivating_example}, the tokenization specified by ring contraction already yields better performance compared with learning from untokenized molecules, with better means and smaller standard deviations for both classification and regression tasks, which suggests that tokenization can \textit{indeed} bring potential performance boosts for molecules.

\begin{figure}[t] % Use [!t] to try placing the figure at the top of the page
    \centering
    \includegraphics[width=0.99\linewidth]{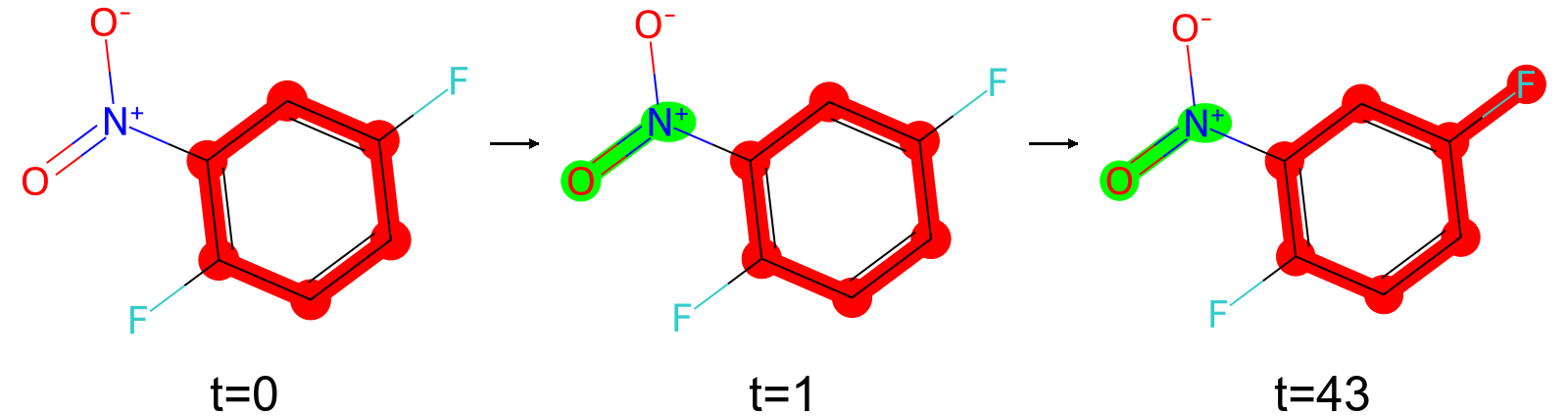}
    \caption{The tokenization of a molecule from \textsc{Mutag} with its SMILES being ``c1cc(c(cc1F)[N+](=O)[O-])F''. We color the identified node sets at iteration $t=0, 1, 43$.} 
\label{fig:mutag_5_tok}
\end{figure}

\subsection{Algorithm}\label{sec: algorithm}
Given a collection of graphs $D=\{G_i=(V_i, E_i)\}_{i=1}^{|D|}$, at each iteration $t$, our algorithm aims to tokenize each graph $G^{t-1}_i$ into a collection of node sets $\mathcal{V}_i^t=\{N_j^{t}|N_j^{t}\in 2^{V_i}\}$, where $2^{V_i}$ denotes the power set over $V_i$ and each node set $N_j^{t}$ is viewed as a hypernode, and constructs the next tokenized graph as $G_i^t=(\mathcal{V}_i^t, E_i^t)$, with $E_i^t=\{(N_j^t, N_k^t)|\exists v_m\in N_j^t, v_n\in N_k^t, (v_m, v_n)\in E_i\}$. A visualization of the tokenization process of our algorithm is presented in Figure~\ref{fig:mutag_5_tok}.

Algorithm~\ref{alg:graph_bpe} shows the proposed \textsc{GraphBPE}, which consists of a \textbf{preprocessing} stage and the \textbf{tokenization} stage. We use $\mathcal{G}$ to denote a general space for graphs, $\mathcal{T}$ to represent the space for different types of topology (e.g., rings), and $\mathcal{S}$ as the space for text strings. We explain the functions used in Algorithm~\ref{alg:graph_bpe} in detail as follows.

\begin{itemize}[topsep=0pt, partopsep=0pt, itemsep=-1pt]
    \item \texttt{Find()}$:\mathcal{G}\times\mathcal{T}\rightarrow 2^V$, a function that finds a certain topology $\tau\in\mathcal{T}$ of a graph $G=(V, E)\in\mathcal{G}$, and returns the node set $N^\tau\in 2^V$ presenting that topology. We abuse the notation of $2^V$ which represents the power set of a specific vertex set $V$ henceforth.
    \item \texttt{Context()}$:\mathcal{G}\times 2^V\rightarrow \mathcal{S}$, a function that contextualizes a node set $N^\tau\in 2^V$ of a graph $G=(V, E)$, mapping it to a identifiable string $s\in \mathcal{S}$.
    \item \texttt{Contract()}$:\mathcal{G}\times2^V\times \mathcal{S}\rightarrow \mathcal{G}$, a function that contracts a graph $G=(V, E)\in\mathcal{G}$ on a node set $N\in 2^V$ and its identifiable string $s\in \mathcal{S}$, and returns a new graph $G^\prime=(V^\prime, E^\prime)\in\mathcal{G}$ with $N$ being its hypernode\footnote{For simplicity we introduce the scenario where one node set is contracted and one hypernode is constructed, in practice we can contract multiple node sets at the same time.}, and construct the edge set $E^\prime$ such that $E^\prime=\{(N, N^\prime)|\exists v_m\in N, v_n\in N^\prime, (v_m, v_n)\in E\}$.
    \item \texttt{S$_\texttt{map}$()}$:\mathcal{S}\times \mathcal{G}\rightarrow 2^V$, a function that keeps track of the mapping between a graph $G=(V, E)\in\mathcal{G}$, an identifiable string $s\in \mathcal{S}$ and the corresponding node set $N\in2^V$.
\end{itemize}

\vskip -0.1in
\begin{algorithm}[H]
   \caption{\textsc{GraphBPE}}
   \label{alg:graph_bpe}
\begin{algorithmic}
   \STATE {\bfseries Input:} a collection of graphs $D=\{G_i\}_{i=1}^{|D|}$, the number of iterations $T$, a topology identifier \texttt{Find()}$:\mathcal{G}\times\mathcal{T}\rightarrow 2^V$, a contextualizer \texttt{Context()}$:\mathcal{G}\times 2^V\rightarrow \mathcal{S}$, a structure contractor \texttt{Contract()}$:\mathcal{G}\times2^V\times \mathcal{S}\rightarrow \mathcal{G}$, a frequency recorder \texttt{Counter()}$:\mathcal{S}\rightarrow\mathbb{Z}^+$, a structure mapper \texttt{S$_\texttt{map}$()}$:\mathcal{S}\times \mathcal{G}\rightarrow 2^V$, and a specific topology $\tau\in\mathcal{T}$ for preprocessing
   \STATE {\bfseries Output:} the tokenized datasets $D^0, D^1, ..., D^T$
   \STATE $D^0\leftarrow \{\}$
   \STATE \# \textit{preprocessing}
   \FOR{$G_i$ {\bfseries in} $D$}
   \IF{$\mathcal{T}=\emptyset$}
   \STATE $D^0\leftarrow D^0\cup\{G_i\}$
   \ELSE
   \STATE $\mathcal{V}_i^{\tau}\leftarrow\texttt{Find($G_i, \tau$)}$
   \STATE $s_i^{\tau}\leftarrow\texttt{Context($G_i^t, \mathcal{V}_i^\tau$)}$
   \STATE $D^0\leftarrow D^0\cup\{G_i^0:=\texttt{Contract($G_i, \mathcal{V}_i^\tau, s_i^{\tau}$)}\}$
   \ENDIF
   \ENDFOR
   \STATE \# \textit{tokenization}
   \STATE $t\leftarrow 0$
   \REPEAT
   \STATE $D^{t+1}\leftarrow \{\}$
   \STATE $\texttt{Counter()}\leftarrow \{\}$
   \FOR{$G_i^{t}(V_i^{t}, E_i^{t})$ {\bfseries in} $D^{t}$}
   \FOR{$e$ {\bfseries in} $E_i^{t}$}
   \STATE $s\leftarrow \texttt{Context($G_i^t, e$)}$
   \STATE $\texttt{Counter($s$)}\leftarrow\texttt{Counter($s$)}+1$
   \STATE $\texttt{S$_\texttt{map}$($s, G^t_i$)}\leftarrow e$
   \ENDFOR
   \ENDFOR
   \STATE $s^*=\underset{s^\prime}{\arg\max\,}\texttt{Counter($s^\prime$)}$
   \FOR{$G_i^{t}(V_i^{t}, E_i^{t})$ {\bfseries in} $D^{t}$}
   \STATE $e\leftarrow \texttt{S$_\texttt{map}$($s^*, G^t_i$)}$
   \STATE $G_i^{t+1}\leftarrow \texttt{Contract($G_i^t, e, s^*$)}$
   \STATE $D^{t+1}\leftarrow D^{t+1}\cup\{G_i^{t+1}\}$
   \ENDFOR
   \STATE $t\leftarrow t+1$
   \UNTIL{$t>T$}
\end{algorithmic}
\end{algorithm}
\vskip -0.2in
\begin{algorithm}[H]
   \caption{\textsc{Contextualizer}}
   \label{alg:contextualizer}
\begin{algorithmic}
   \STATE {\bfseries Input:} a graph $G=(V, E)$, a set of nodes $V_c\in 2^V$, a name mapper \texttt{N$_\texttt{map}$()}$:V\times2^V\rightarrow \mathcal{S}$
   \STATE {\bfseries Output:} a string representation $s$ for $V_c$ 
   \STATE \# \textit{initialize $s$ to be an empty list}
   \STATE $s\leftarrow [\,\,]$ 
   \FOR{$v$ {\bfseries in} $V_c$}
   \STATE $s\leftarrow s+\texttt{N$_\texttt{map}$($v, \mathcal{N}(v)$)}$
   \ENDFOR
   \STATE $s\leftarrow\texttt{Sort($s$)}$
   \STATE $s\leftarrow\texttt{Concat($s$)}$
\end{algorithmic}
\end{algorithm}
% \vskip -0.25in
\textbf{Preprocessing}. Given a topology $\tau$ (e.g., ring or clique) of interest, we first preprocess the dataset $D$ by contracting the structure $\tau$ for each graph. Specifically, after the node sets for $\tau$ in $G$ are identified by \texttt{Find()}, we contract $G$ into a new graph $G^0$ with \texttt{Contract()}, based on the node sets and their contextualized representations. In practice, we only consider $\tau$ being rings or cliques, and $\tau=\emptyset$ means the preprocessing is omitted.

\textbf{Tokenization}. Given a graph $G_i^{t-1}\in D^{t-1}$, whose vertices are node sets in $2^{V_i}$, we aim to contract $G_i^{t-1}$ and build $D^{t}$ following a ``\textbf{count-and-merge}'' paradigm similar to BPE (as illustrated in Table~\ref{tab: bpe_example}). Specifically, node pairs (i.e., edges) in graphs are the natural analog of paired tokens in texts, and \textsc{GraphBPE} first contextualizes each edge in $D^{t-1}$ into an identifiable string using \texttt{Context()}, and \textbf{counts} its frequency, recorded with \texttt{S$_\texttt{map}$()}. The mostly co-occurred node pair, represented by $s^*$, is then selected to \textbf{merge}, where we iterate $D^{t-1}$ again to contract graphs that contain the identification $s^*$, and construct $D^t$ for the next round of tokenization.

We provide an example implementation of \texttt{Context()} in Algorithm~\ref{alg:contextualizer}. Despite the resemblance between edges and token pairs, one should note that edges in \textsc{GraphBPE} should be treated \textit{orderless}, meaning as long as two edges contain the same two identifiable strings, they should be viewed as the same (e.g., ``$s_1\text{-}s_2$'' is the same as ``$s_2\text{-}s_1$''), which is different from BPE on texts, where the token pairs are \textit{order-sensitive} (e.g., ``lo'' is different from ``ol''). By customizing the contextualizer, \textsc{GraphBPE} can produce different tokenization strategies, and we present a detailed discussion on how it connects \textsc{GraphBPE} with other tokenization algorithms in Appendix~\ref{app: discussion_on_contextualizer}.\\
Note that the tokenized graph $G^t=(V^t, E^t)$ produced by \textsc{GraphBPE} can be viewed as both a simple graph and a hypergraph. Since $V^t\subseteq2^V$ and $E^t$ is constructed by \texttt{Contract()} such that $G^t$ and $G$ have the same number of connected components, with the (untokenized) simple graph being $G=(V, E)$, a simple graph can be derived from $G^t$, with each vertex defined by the node ensemble of vertices of $G^t$, and its topology defined by $E^t$. Naturally, $G^t$ defines a hypergraph with hyperedges specified by $V^t$ and $E$, where for vertex $v_s\in V^t, |v_s|=1$ that remains a single node from $V$, we construct the hyperedges based on the edges $(v_s, v_n)\in E, v_n\in\mathcal{N}(v_s)$.
\section{Experiment}
In this section, we introduce the datasets, tokenization methods for comparison, and models for simple graphs and hypergraphs, and then present the experiment results.
\subsection{Dataset}
We conduct experiments on graph-level classification and regression datasets, and show their statistics in Table~\ref{tab:dataset_stat}. We detail the train-validation-test split in Appendix~\ref{app: implementation}.\\
\textbf{Classification} For graph classification tasks, we choose \textsc{Mutag, Enzymes}, and \textsc{Proteins} from the TUDataset~\cite{morris2020tudataset}. \textsc{Mutag} is for binary classification where the goal is to predict the mutagenicity of compounds. \textsc{Enzymes} is a multi-class dataset that focuses on classifying a given enzyme into 6 categories, and \textsc{Proteins} aims to classify whether a protein structure is an enzyme or not.
\begin{table}[h!]
    \centering
    \small
    \resizebox{0.49\textwidth}{!}{%
    \begin{tabular}{lcccc}
  \hline
   \textbf{dataset}   & \textbf{\# molecule} & \textbf{\# class} & \textbf{label distri.} & \textbf{\# node type}   \\
  \hline
   \textsc{Mutag}   & 188 & 2 & 125:63 & 7\\
   \textsc{Enzymes}   & 600 & 6 & \text{balanced} & 3\\
   \textsc{Proteins}   & 1113 & 2 & 450:663 & 3\\
    \spacedhline
    \textsc{Freesolv}   & 642 & 1 & \text{none} & 9\\
    \textsc{Esol}   & 1128 & 1 & \text{none} & 9\\
    \textsc{Lipophilicity} & 4200 & 1 & \text{none} & 9\\
  \hline
  \end{tabular}%
}
    \caption{The statistics of different datasets, where \textbf{label distri.} stands for label distribution. The 1st and 2nd blocks are for graph classification and regression, respectively. }
    \label{tab:dataset_stat}
\end{table}\\
\textbf{Regression} For graph regression tasks, we use \textsc{Freesolv, Esol}, and \textsc{Lipophilicity} from the MoleculeNet~\cite{wu2018moleculenet}, where \textsc{Freesolv} aims to predict free energy of small molecules in water, \textsc{Esol} targets at predicting water solubility for common organic small molecules, and \textsc{Lipophilicity} focuses on octanol/water distribution coefficient.
\begin{figure*}[t] % Use [!t] to try placing the figure at the top of the page
    \centering
    \includegraphics[width=0.99\linewidth]{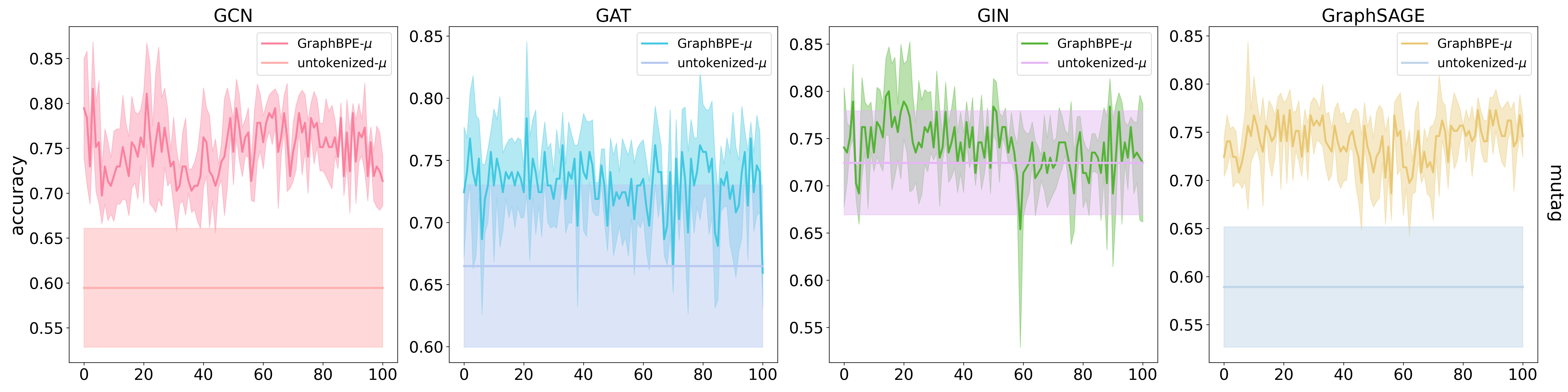}
    \includegraphics[width=0.99\linewidth]{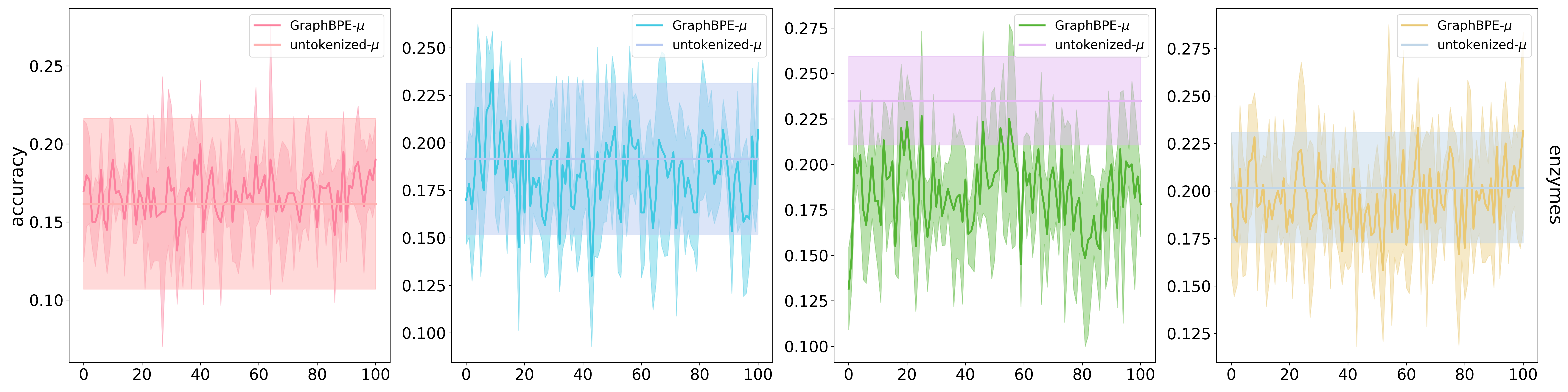}
    \includegraphics[width=0.99\linewidth]{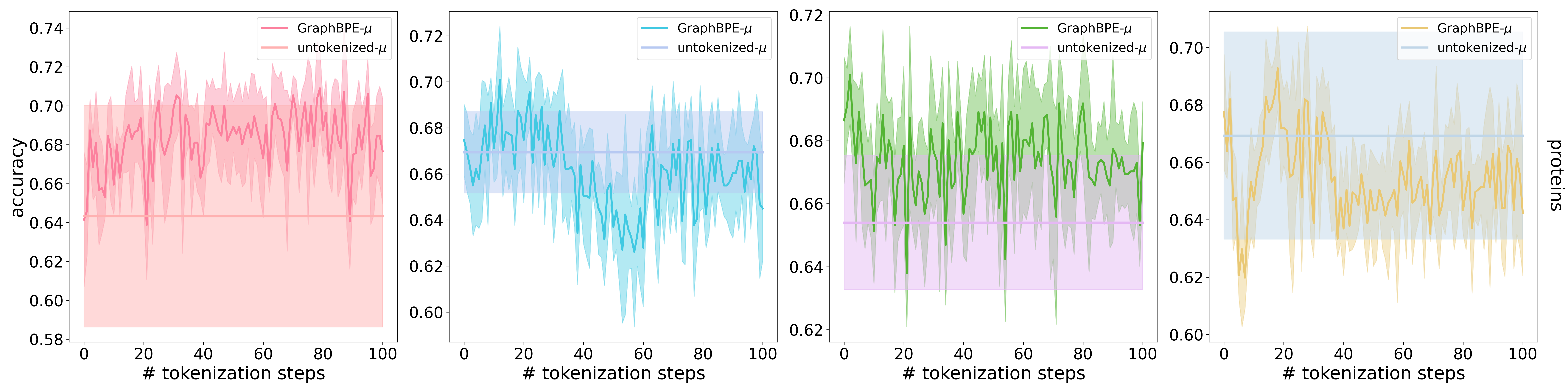}
    \caption{Results of a 3-layer GCN, GAT, GIN, and GraphSAGE with a learning rate of 0.01 and a hidden size of 32 on \textsc{Mutag, Enzymes,} and \textsc{Proteins} (1st, 2nd, 3rd row, respectively), with \textbf{accuracy} the \textit{higher} the better. The x-axis denotes the number of tokenization steps in our GraphBPE algorithm. We plot $\mu\pm\sigma$ over 5 runs for each configuration.} 
\label{fig:mutag_enzymes_proteins_all_gnns_lr0.01h32}
\end{figure*}
\begin{figure*}[t] % Use [!t] to try placing the figure at the top of the page
    \centering
    \includegraphics[width=0.99\linewidth]{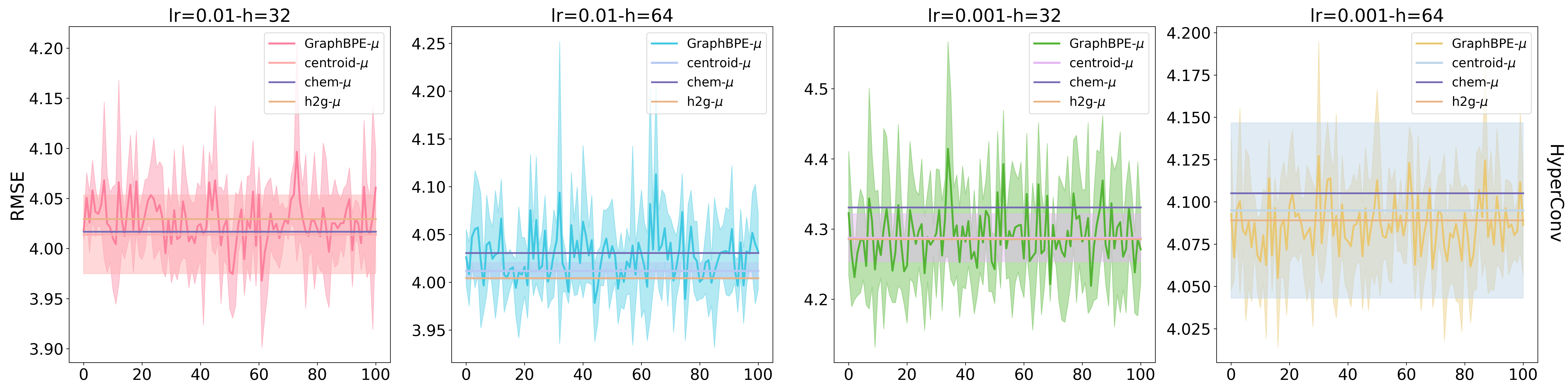}
    \includegraphics[width=0.99\linewidth]{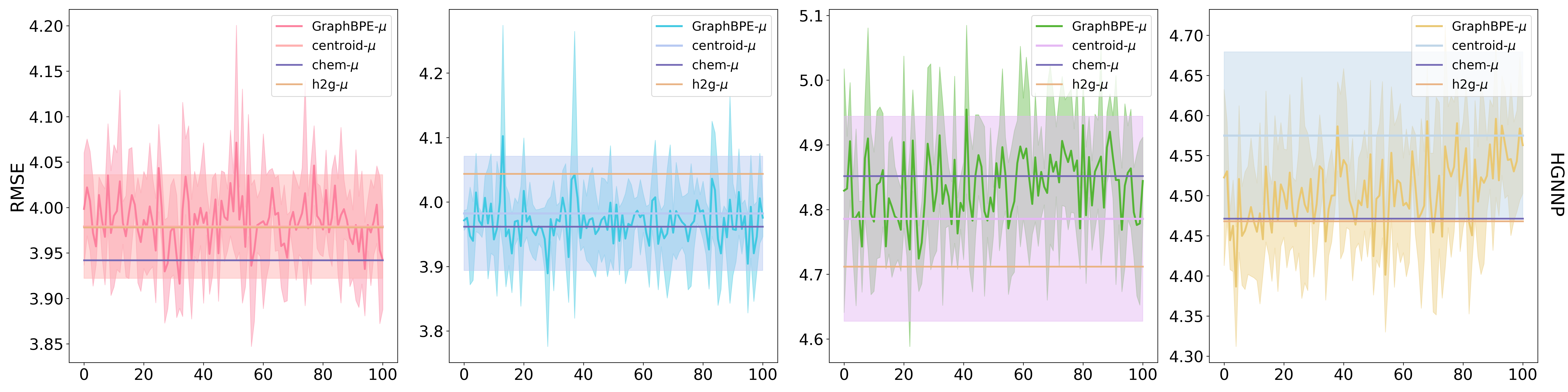}
    \includegraphics[width=0.99\linewidth]{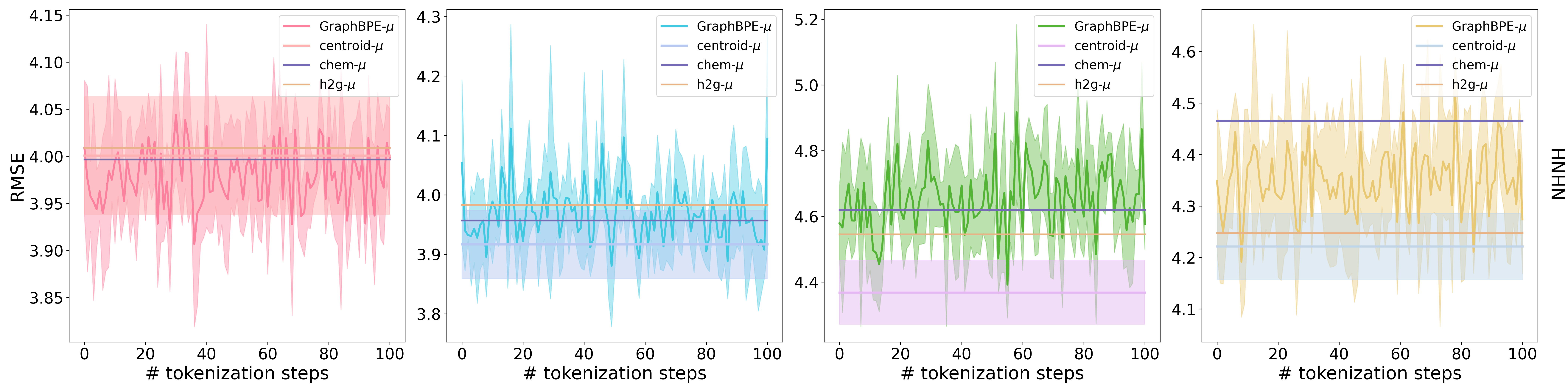}
    \caption{Results of a 3-layer HyperConv, HGNN++, and HNHN (1st, 2nd, 3rd row, respectively) with a learning rate of \{0.01, 0.001\} and a hidden size of \{32, 64\} on \textsc{Freesolv}, with \textbf{RMSE} the \textit{lower} the better. The x-axis denotes the number of tokenization steps in our GraphBPE algorithm. We plot $\mu\pm\sigma$ over 5 runs for \textsc{GraphBPE, Centroid}, and omit $\pm\sigma$ for \textsc{Chem, H2g} for better visualization.} 
\label{fig:freesolv_all_hypergnns_3layer_all_config}
\end{figure*}
\subsection{Tokenization}
Given a simple graph $G=(V, E)$, \textsc{GraphBPE} translates it into another graph $G^\prime$ whose vertices are node sets in $2^V$, which can be then used to construct a hypergraph as defined in Section~\ref{sec: hypergraph}. We introduce three other hypergraph construction strategies as follows.\\
\textbf{Centroid} Following \citet{feng2019hypergraph}, we construct the hyperedges by choosing each vertex together with its 1-hop neighbors. This is domain-agnostic and requires no extra knowledge, and we refer to it as \textsc{Centroid}.\\
\textbf{Chemistry-Informed} We can construct hyperedges such that the nodes within which represent functional groups~\cite{10.1093/bib/bbad398}. Specifically, we use RDKit~\cite{landrum2006rdkit} to extract functional groups\footnote{\href{http://rdkit.org/docs/source/rdkit.Chem.Fragments.html}{http://rdkit.org/docs/source/rdkit.Chem.Fragments.html}}, and construct each hyperedge based on the nodes that belong to the same functional group. For a node that does not belong to any functional groups, we treat its edges as the respective hyperedges. This method requires domain knowledge in chemistry and we refer to it as \textsc{Chem}.\\
\textbf{Hyper2Graph} \citet{jin2020hierarchical} introduce a motif extraction schedule for molecules based on chemistry knowledge and heuristics. We treat the extracted motifs, which are not necessarily meaningful substructures such as functional groups, as a type of tokenization and refer to this method as \textsc{H2g}.
\subsection{Model}
We choose two types of models for evaluation, with \textbf{GNN} for (untokenized) simple graphs, and graphs tokenized by \textsc{GraphBPE} at each iteration, and \textbf{HyperGNN} for hypergraphs defined by the tokenization of \textsc{GraphBPE} at each iteration, and constructed by other algorithms. We detail the model implementations in Appendix~\ref{app: implementation}.\\
\textbf{GNN} We choose GCN~\cite{kipf2017semisupervised}, GAT~\cite{veličković2018graph}, GIN~\cite{xu2019powerful}, and GraphSAGE~\cite{hamilton2018inductive} for (untokenized) simple graphs and graphs specified by the tokenization of \textsc{GraphBPE} at each iteration.\\
\textbf{HyperGNN} For hypergraphs constructed by \textsc{GraphBPE} and other tokenization methods, we choose HyperConv~\cite{bai2020hypergraph}, HGNN++~\cite{9795251}, which shows improved performances in metrics and standard deviation over HGNN~\cite{feng2019hypergraph} in our preliminary study, and HNHN~\cite{dong2020hnhn} as our three backbones.
\subsection{Result}
We present experiment results on both classification datasets, with accuracy reported, and regression datasets, with RMSE reported, as suggested by~\citet{wu2018moleculenet}, where for each configuration we run experiments 5 times and report the mean and standard deviation of the metrics. For \textsc{GraphBPE}, we present the results on preprocessing with 100 steps of tokenization. We also report the results on the number of times \textsc{GraphBPE} is statistically (with p-value $<0.05$) / numerically better / the same / worse compared with the baselines. Due to space limits, we present the rest of the results in Appendix~\ref{app: results}.\\
\textbf{GNN} We present the test accuracy for 3-layer GNNs on \textsc{Mutag, Enzymes}, and \textsc{Proteins} in Figure~\ref{fig:mutag_enzymes_proteins_all_gnns_lr0.01h32} and the results on performance comparison in Table~\ref{tab:p_value_metric_value_for_4gnns_on_3_classification_datasets}. 

In Figure~\ref{fig:mutag_enzymes_proteins_all_gnns_lr0.01h32}, we can observe that on the \textsc{Mutag} dataset, \textsc{GraphBPE} performs better in general across different GNN architectures, especially for GCN and GraphSAGE, where at different time steps our algorithm consistently outperforms the untokenized molecular graphs in terms of mean$\pm$std. This suggests that tokenization could potentially help the performance of GNNs on molecular graphs. For \textsc{Enzymes} and \textsc{Proteins}, \textsc{GrapgBPE} does not consistently perform better than untokenized graphs, where both the tokenization step and the choice of the model will affect the accuracy. For example, approximately the first 20 tokenization steps are favored by GAT on both \textsc{Enzymes} and \textsc{Proteins}, and the performance begins to degenerate as the tokenization step increases, while for GIN on \textsc{Enzymes}, our algorithm is outperformed in all time steps.
\vskip -0.06in
\begin{table}[!t]
    \centering
    \resizebox{0.49\textwidth}{!}{%
    \begin{tabular}{lccccc}
  \hline
   \textbf{dataset} & \textbf{strategy} & \textbf{GCN}              & \textbf{GAT}              & \textbf{GIN}              & \textbf{GraphSAGE}        \\
  \spacedhline
   \multirow{2}{*}{\textsc{Mutag}} & p-value  & \textcolor{red}{\textbf{93}}:8:0 & 17:\textcolor{red}{\textbf{84}}:0 & 1:\textcolor{red}{\textbf{100}}:0 & \textcolor{red}{\textbf{96}}:5:0 \\
   & metric & \textbf{101}:0:0 & \textbf{99}:0:2 & \textbf{78}:2:21 & \textbf{101}:0:0 \\
   \spacedhline
   \multirow{2}{*}{\textsc{Enzymes}} & p-value  & 0:\textcolor{red}{\textbf{101}}:0 & 0:\textcolor{red}{\textbf{100}}:1 & 0:\textcolor{red}{\textbf{54}}:47 & 0:\textcolor{red}{\textbf{101}}:0 \\
   & metric & \textbf{66}:1:34 & 38:0:\textbf{63} & 0:0:\textbf{101} & 37:1:\textbf{63} \\
   \spacedhline
   \multirow{2}{*}{\textsc{Proteins}} & p-value  & 1:\textcolor{red}{\textbf{100}}:0 & 1:\textcolor{red}{\textbf{94}}:6 & 16:\textcolor{red}{\textbf{84}}:1 & 0:\textcolor{red}{\textbf{101}}:0 \\
   & metric & \textbf{98}:0:3 & 32:1:\textbf{68} & \textbf{95}:0:6 & 16:0:\textbf{85} \\
  \hline
  \end{tabular}%
}
    \caption{Performance comparison on the accuracy of classification datasets for 3-layer GNNs with a learning rate of 0.01 and a hidden size of 32. For each triplet $a$:$b$:$c$, $a, b, c$ are the number of times \textsc{GraphBPE} is better / the same / worse compared with (untokenized) simple graphs. ``p-value'' stands for comparison based on p-value $<0.05$ from \textit{t}-test, and ``metric'' means numerical comparison of the metric values, where best within the triplet is \textbf{\textcolor{red}{bo}ld}.}
    \label{tab:p_value_metric_value_for_4gnns_on_3_classification_datasets}
    \vskip -0.2in
\end{table}
\vskip -0.06in
In terms of metric value comparison and statistical significance, we can observe from Table~\ref{tab:p_value_metric_value_for_4gnns_on_3_classification_datasets} that most of the time we can outperform untokenized graphs in terms of average accuracy, while performing as least the same under the lens of significance tests (e.g., \textit{t}-test with a p-value $<$ 0.05), which aligns with our findings from Figure~\ref{fig:mutag_enzymes_proteins_all_gnns_lr0.01h32}.

In general, we can observe that \textsc{GraphBPE} performs less satisfyingly as the dataset size increases, we suspect this might be because the vocabulary to be mined on large datasets is complex and diverse, such that the number of (limited) tokenization steps would affect the performance; thus, more tokenization steps might be favored to achieve better results on large datasets.\\
\vskip -0.1in
\textbf{HyperGNN} We present the test RMSE for 3-layer hyperGNNs on \textsc{Freesolv} over different configurations (learning rate $\times$ hidden size) in Figure~\ref{fig:freesolv_all_hypergnns_3layer_all_config} and the results on performance comparison for one configuration in Table~\ref{tab:p_value_metric_value_for_3hgnns_on_freesolv_lr0.01_h32}.
\vskip -0.06in
\begin{table}[!t]
    \centering
    \resizebox{0.49\textwidth}{!}{%
    \begin{tabular}{lcccc}
  \hline
   \textbf{method} & \textbf{strategy} & \textbf{HyperConv}              & \textbf{HGNN++}              & \textbf{HNHN}           \\
  \spacedhline
   \multirow{2}{*}{\textsc{Centroid}} & p-value  & 0:\textcolor{red}{\textbf{101}}:0 & 0:\textcolor{red}{\textbf{101}}:0 & 1:\textcolor{red}{\textbf{100}}:0  \\
   & metric & 26:0:\textbf{75} & 38:0:\textbf{63} & \textbf{75}:0:26  \\
   \spacedhline
   \multirow{2}{*}{\textsc{Chem}} & p-value  & 0:\textcolor{red}{\textbf{99}}:2 & 0:\textcolor{red}{\textbf{97}}:4 & 0:\textcolor{red}{\textbf{100}}:1 \\
   & metric & 28:0:\textbf{73} & 5:0:\textbf{96} & \textbf{71}:0:30\\
   \spacedhline
   \multirow{2}{*}{\textsc{H2g}} & p-value  & 1:\textcolor{red}{\textbf{100}}:0 & 0:\textcolor{red}{\textbf{99}}:2 & 0:\textcolor{red}{\textbf{101}}:0 \\
   & metric & \textbf{60}:0:41 & 35:0:\textbf{66} & \textbf{83}:0:18 \\
  \hline
  \end{tabular}%
}
    \caption{Performance comparison on the RMSE of 3-layer HyperGNNs with a learning rate of 0.01 and a hidden size of 32 on \textsc{Freesolv}. For each triplet $a$:$b$:$c$, $a, b, c$ are the number of times \textsc{GraphBPE} is better / the same / worse compared with other tokenization methods. ``p-value'' stands for comparison based on p-value $<0.05$ from \textit{t}-test, and ``metric'' means numerical comparison of the metric values, where best within the triplet is \textbf{\textcolor{red}{bo}ld}.}
    \label{tab:p_value_metric_value_for_3hgnns_on_freesolv_lr0.01_h32}
    \vskip -0.2in
\end{table}

As shown in Figure~\ref{fig:freesolv_all_hypergnns_3layer_all_config}, both learning rate and model architectures can largely affect the test performance, and no tokenization methods can perform universally well across different configurations. In terms of the average performance, there generally exists some steps for \textsc{GraphBPE} in different configurations, which have the lowest RMSE compared with other tokenization methods. However, there is yet no method that can determine such ``optimal'' tokenization steps ahead of training. 

We choose the configuration with a learning rate of 0.01 and a hidden size of 32 to further conduct performance comparison, as it achieves lower RMSE across different models. As detailed in Table~\ref{tab:p_value_metric_value_for_3hgnns_on_freesolv_lr0.01_h32}, we can observe that model architectures will affect the performance, and again, no tokenization method is the best among different configurations. For metric value comparison, \textsc{GraphBPE} shows good performance on HNHN and frequently outperforms other tokenization methods, while less is less satisfying on other models. However, we can observe that there always exists some steps (i.e., the number of times \textsc{GraphBPE} is better compared with other tokenization methods) that \textsc{GraphBPE} achieves a lower RMSE, similar to the findings from Figure~\ref{fig:freesolv_all_hypergnns_3layer_all_config}. In terms of the comparison based on p-values, our method performs the same compared with the baselines most of the time, unlike on \textsc{Mutag} where we can demonstrate statistical significance. We suspect that this might be because tokenization is in general less effective for regression tasks, which is supported by the results in Appendix~\ref{app: results}.

In general, we can observe that the choice of tokenization methods will largely affect the performance of hyperGNNs, suggesting that a well-designed hypergraph construction strategy would benefit hyperGNNs on molecular graphs. Although \textsc{GraphBPE} often has steps that achieve a smaller RMSE against the baselines for different configurations, it in general shows limited improvement for regression tasks compared to classification tasks.
\vskip -0.2in
\section{Conclusion}
In this work, we explore how tokenization would help molecular machine learning on classification and regression tasks, and propose \textsc{GraphBPE}, a \textbf{count-and-merge} algorithm that tokenize a simple graph into node sets, which are later used to construct a new (simple) graph or a hypergraph. Our experiment across various datasets and models suggests that tokenization will affect the test performance, and our proposed \textsc{GraphBPE} tends to excel on small classification datasets, given a limited number of tokenization steps. 

We explore the simple idea of how different views of molecular graphs would benefit graph-level tasks, and we hope our results can inspire more discussions and attract attention to the data preprocessing schedules for molecule machine learning, which is less studied compared with innovations on models and benchmarks.
\vskip -0.2in
\section*{Limitation}
\textbf{Types of Tokenization} We include two types of tokenization baselines where one is based on chemistry knowledge and the other is based on pre-defined rules. However, there exist more sophisticated tokenization methods, such as deriving tokenization rules from an off-the-shelf GNN, which are not discussed in this work.

\textbf{Types of Task \& Dataset} We focus on graph-level tasks and exclude node-level tasks. For the classification and regression task, although we include three datasets for each and consider both binary and multi-class classification datasets, the size of our datasets (e.g., $\sim 10^{3}$) is relatively small compared with those usually used for molecular graph pre-training (e.g., $\sim 10^6$).
\vskip -0.2in
\section*{Impact Statement}
This paper presents work whose goal is to advance the field of 
Machine Learning. There are many potential societal consequences 
of our work, none of which we feel must be specifically highlighted here.
\bibliography{graphBPE}
\bibliographystyle{icml2024}

%%%%%%%%%%%%%%%%%%%%%%%%%%%%%%%%%%%%%%%%%%%%%%%%%%%%%%%%%%%%%%%%%%%%%%%%%%%%%%%
%%%%%%%%%%%%%%%%%%%%%%%%%%%%%%%%%%%%%%%%%%%%%%%%%%%%%%%%%%%%%%%%%%%%%%%%%%%%%%%
% APPENDIX
%%%%%%%%%%%%%%%%%%%%%%%%%%%%%%%%%%%%%%%%%%%%%%%%%%%%%%%%%%%%%%%%%%%%%%%%%%%%%%%
%%%%%%%%%%%%%%%%%%%%%%%%%%%%%%%%%%%%%%%%%%%%%%%%%%%%%%%%%%%%%%%%%%%%%%%%%%%%%%%
\newpage
\appendix
\onecolumn
\section{Implementation}\label{app: implementation}
For all the models, we use \{1, 2, 3\}-layer architecture with a hidden size of \{32, 64\} and a learning rate of \{0.01, 0.001\}. For both classification and regression tasks, we apply a 1-layer MLP with a dropout rate of 0.1. We use a batch size of the form $2^N\cdot10^M$ where $N, M$ are chosen such that the batch size can approximately cover the entire training set, and we further apply BatchNorm~\cite{ioffe2015batch} to stabilize the training. We train the model for 100 epochs, and report the mean $\mu$ and standard deviation $\sigma$ over 5 runs for the test performance on the model with the best validation performance.

For datasets with a size smaller than 2000, we adopt a train-validation-test split of 0.6/0.2/0.2, and use 0.8/0.1/0.1 for larger datasets. We ignore the edge features and use the one-hot encodings as the node features. For the tokenized graphs from \textsc{GraphBPE}, we use the summation of the node features as the representation for that hypernode for our experiments on GNNs. For classification datasets, we make the validation and test set as balanced as possible, as suggested in our preliminary study, that using a random split validation set might favor models that are not trained at all (e.g., models always predict positive for binary classification task). For regression datasets, we follow \citet{wu2018moleculenet} and use random split. For \textsc{Mutag, Freesolv, Esol}, and \textsc{Lipophilicity}, we set the topology to be contracted as rings in the preprocessing stage, and we set that for \textsc{Enzymes} and \textsc{Proteins} as cliques.

Note that from untokenized graphs to the last iteration $T$, we can track how the nodes merge into node sets in the graph and thus develop a tokenization rule for unseen graphs. However, for simplicity and efficiency, we first tokenize the entire dataset before we split them into train/validation/test sets. Our code is available at \href{https://github.com/A-Chicharito-S/GraphBPE}{https://github.com/A-Chicharito-S/GraphBPE}.
\section{Discussion on Contextualizer}\label{app: discussion_on_contextualizer}
\begin{wrapfigure}{r}{0.5\textwidth} % 'r' for right, '0.5\textwidth' for half the text width
    \vspace{-20pt} % Adjust vertical space as needed
    \begin{minipage}{0.5\textwidth}
        \begin{algorithm}[H]
   \caption{\textsc{PSE-Contextualizer}}
   \label{alg:pse}
\begin{algorithmic}
   \STATE {\bfseries Input:} a graph $G=(V, E)$, a set of nodes $V_c\in 2^V$, a name mapper \texttt{N$_\texttt{map}$()}$:V\times2^V\rightarrow \emptyset$
   \STATE {\bfseries Output:} a string representation $s$ for $V_c$ 
   \STATE \# \textit{initialize $s$ to be an empty list}
   \STATE $s\leftarrow [\,\,]$ 
   \FOR{$v$ {\bfseries in} $V_c$}
   \STATE $s\leftarrow s+\texttt{N$_\texttt{map}$($v, \mathcal{N}(v)$)}$
   \ENDFOR
   \STATE $s\leftarrow\texttt{Sort($s$)}$
   \STATE $s\leftarrow\texttt{Concat($s$)}$
\end{algorithmic}
\end{algorithm}
    \end{minipage}
    \vspace{-10pt} % Adjust vertical space as needed
\end{wrapfigure}
For the Principal Subgraph Extraction (PSE) algorithm proposed by \citet{kong2022molecule}, we can recover it from \textsc{GraphBPE} by skipping the preprocessing stage, while setting the contextualizer as Algorithm~\ref{alg:pse}. The only difference between the PSE contextualizer and ours is that in Algorithm~\ref{alg:pse}, the name mapper \texttt{N$_\texttt{map}$()} returns an empty string for any node sets, while ours returns the string representation for the neighborhood, meaning PSE does not take the neighbors/context into consideration during tokenization. For the Connection-Aware Motif Mining algorithm proposed by \citet{geng2023novo}, where the connection among the nodes is considered to mine common substructures (e.g., as illustrated in Figure 2 of \citet{geng2023novo}, 3 hypernodes can be contracted at a time), we can recover it by increasing the number of tokenization steps, which mitigates the fact that \textsc{GraphBPE} always select one pair of nodes to contract.

Note that by customizing the \texttt{N$_\texttt{map}$()} function, we can further introduce external knowledge (e.g., include information about the chemistry properties), and constraints (e.g., limit the maximum size of the node set) in the tokenization process, and potentially extend our algorithm for non-molecular graphs that do not necessarily share common node types across different graphs, where instead of return the string representation of the neighborhood, \texttt{N$_\texttt{map}$()} can give out the structural information (e.g., the degree of the node) that reveals the neighborhood to facilitate tokenization.
\section{Result}\label{app: results}
We include the visualization of the test performance, and the performance comparison based on p-value and metric-value for \textsc{Mutag, Enzymes, Proteins, Freesolv, Esol,} and \textsc{Lipophilicity} in Section~\ref{app: mutag_result}, ~\ref{app: enzymes_result}, ~\ref{app: proteins_result}, ~\ref{app: freesolv_result}, ~\ref{app: esol_result} and ~\ref{app: lipo_result}. For better visualization, we plot both the mean and the standard deviation as $\mu\pm\sigma$ for our experiments on GNNs, and exclude the standard deviation for the \textsc{Chem, H2g} baselines on HyperGNNs. For the performance comparison, we use the triplet $a:b:c$ to denote the number of times our algorithms are better / the same / worse compared with the baseline, and use \textcolor{red}{\textbf{red}} to mark the best within the triplet based on p-value comparison, and \textbf{black} to mark that for metric value comparison. We can observe that in general, given the 100 tokenization steps, \textsc{GraphBPE} tend to perform well on small datasets, which we suspect is due to the reason that larger datasets contain richer substructures to learn; thus, may need more tokenization rounds. Compared with regression tasks, \textsc{GraphBPE} tends to provide more boosts for classification tasks.
\newpage
\subsection{\textsc{Mutag}}\label{app: mutag_result}
\begin{figure}[H]
\centering
\includegraphics[width=0.9\linewidth, height=0.59\textwidth]{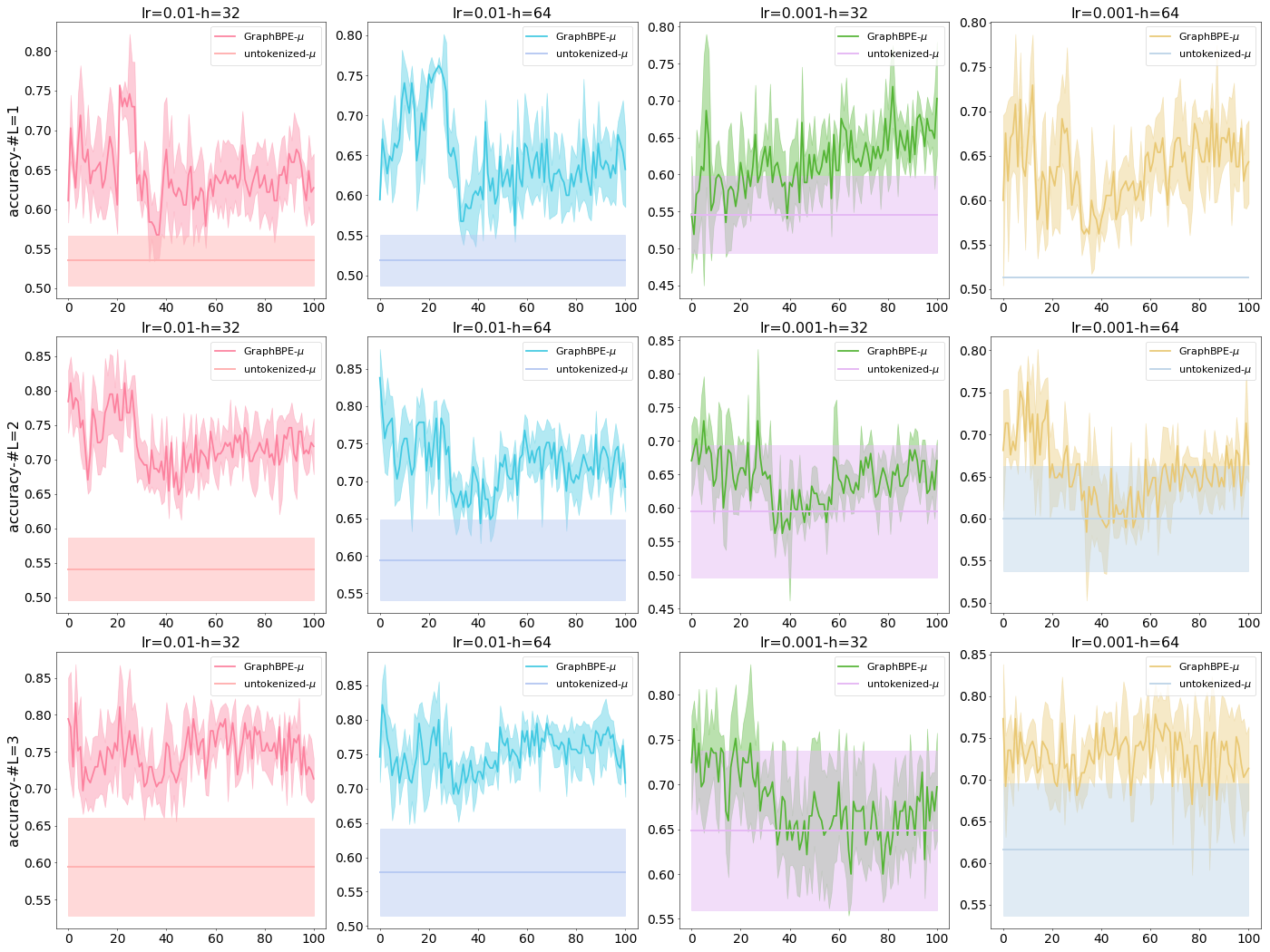}
\caption{Results of GCN on \textsc{Mutag}, with \textbf{accuracy} the \textit{higher} the better} 
\label{fig:mutag_gcn}
\end{figure}

\begin{figure}[H]
\centering
\includegraphics[width=0.9\linewidth, height=0.59\textwidth]{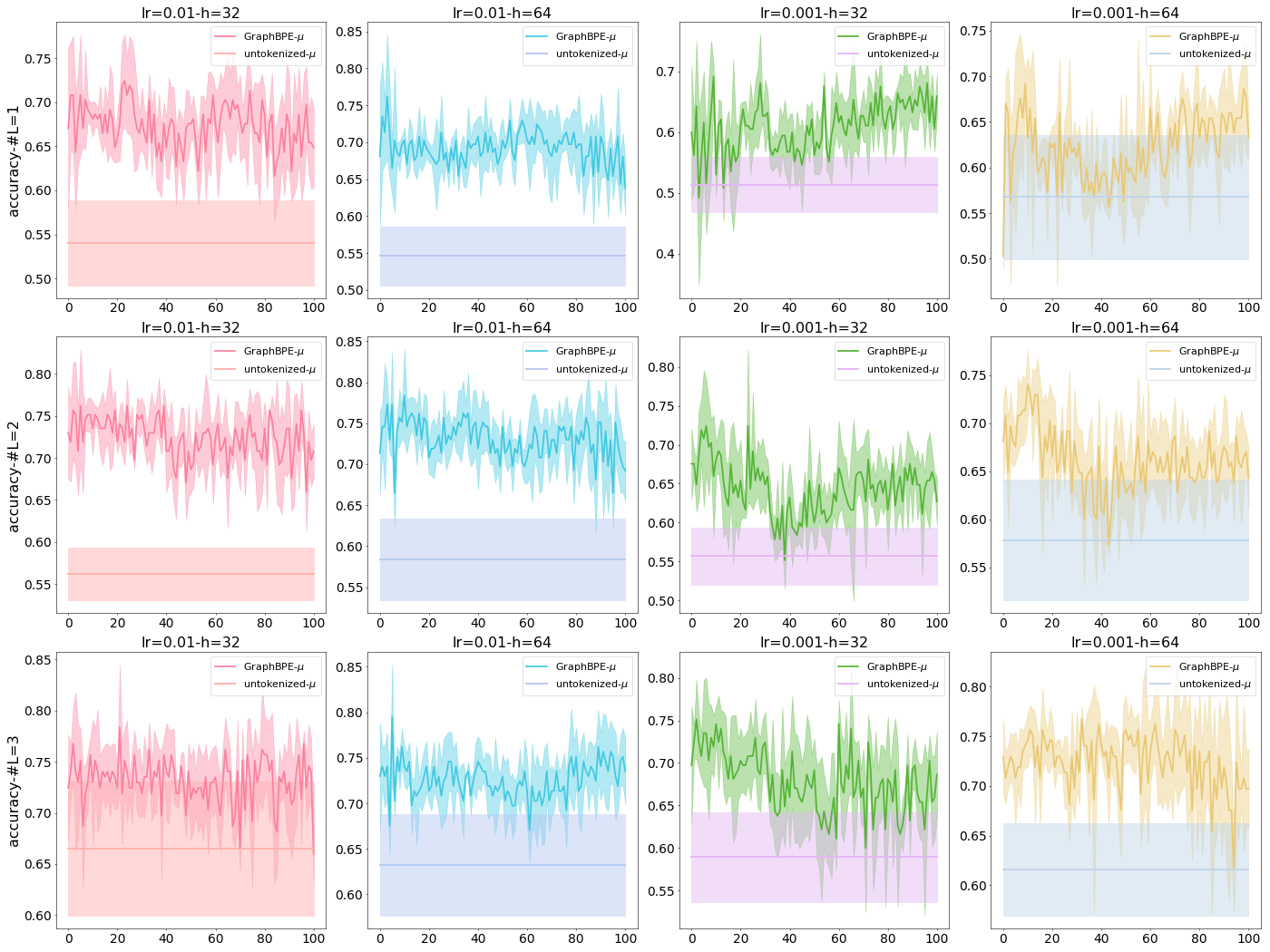}
\caption{Results of GAT on \textsc{Mutag}, with \textbf{accuracy} the \textit{higher} the better} 
\label{fig:mutag_gat}
\end{figure}

\begin{figure}[H]
\centering
\includegraphics[width=0.9\linewidth, height=0.59\textwidth]{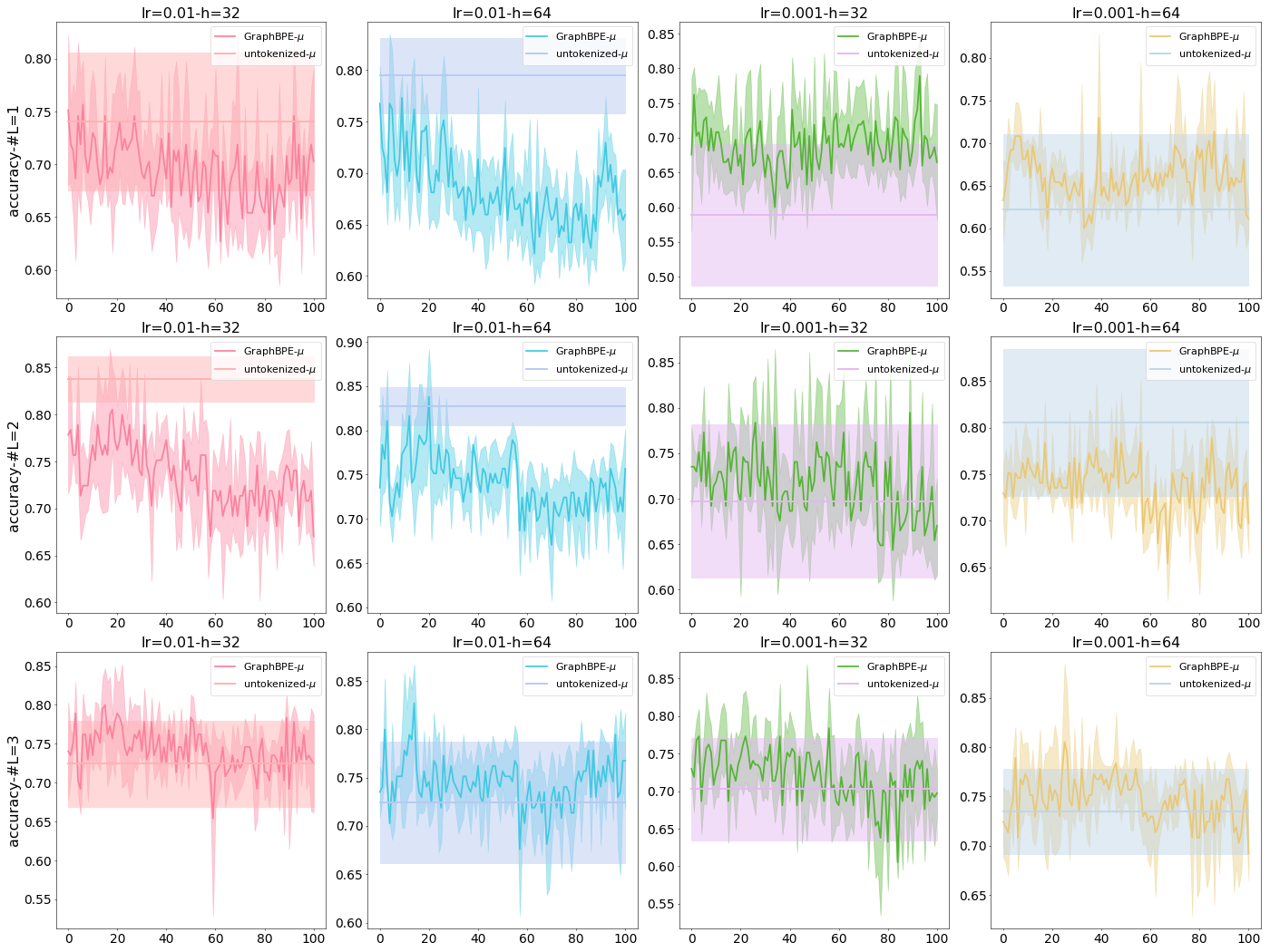}
\caption{Results of GIN on \textsc{Mutag}, with \textbf{accuracy} the \textit{higher} the better} 
\label{fig:mutag_gin}
\end{figure}

\begin{figure}[H]
\centering
\includegraphics[width=0.9\linewidth, height=0.59\textwidth]{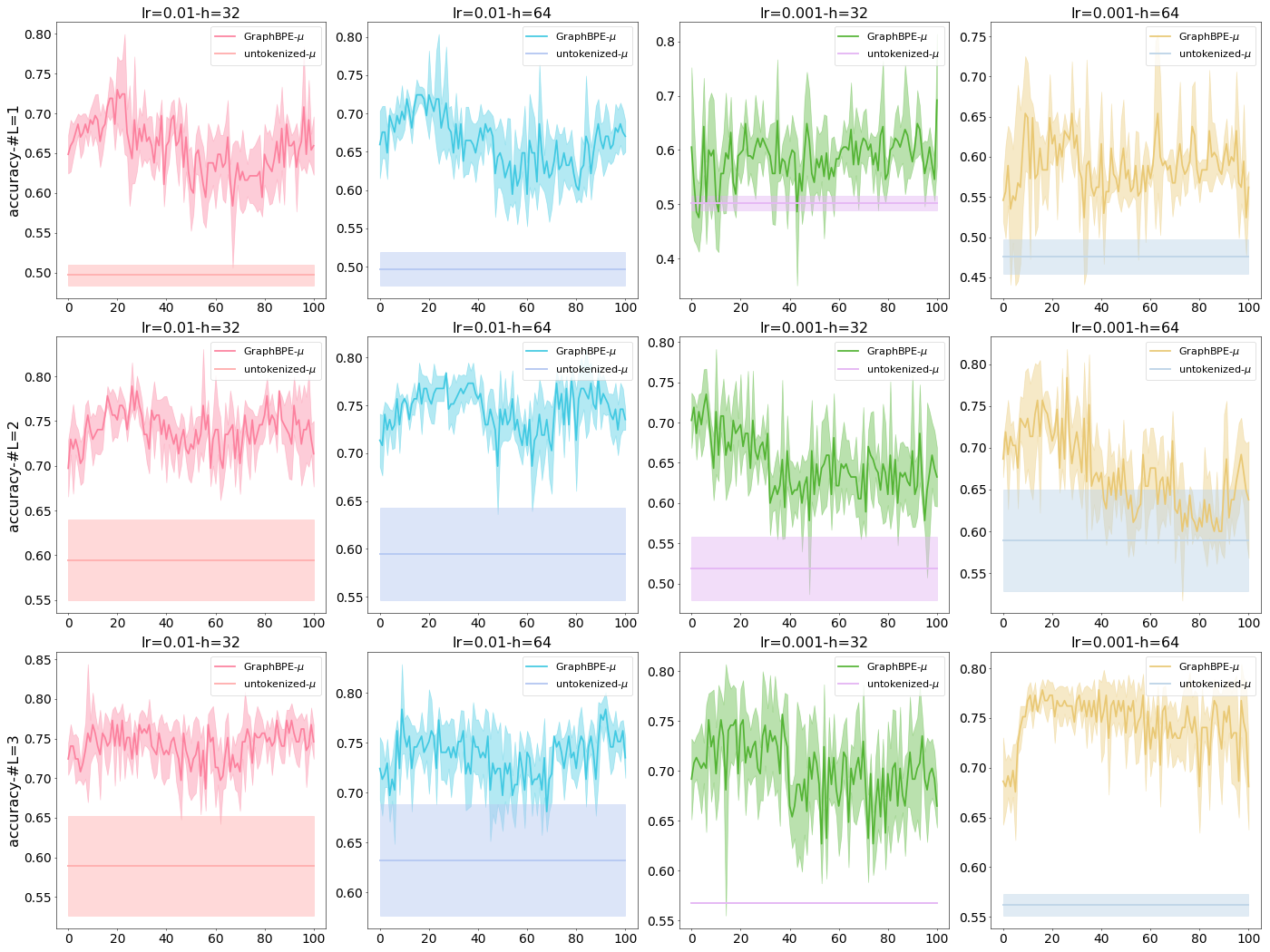}
\caption{Results of GraphSAGE on \textsc{Mutag}, with \textbf{accuracy} the \textit{higher} the better} 
\label{fig:mutag_graphsage}
\end{figure}

For GNNs, we include the performance comparison results in Table~\ref{tab: p_metric_4gnns_on_mutag}, and the visualization over different tokenization steps in Figure~\ref{fig:mutag_gcn}, ~\ref{fig:mutag_gat}, ~\ref{fig:mutag_gin}, and ~\ref{fig:mutag_graphsage} for GCN, GAT, GIN, and GraphSAGE.

\begin{figure}[H]
\centering
\includegraphics[width=0.9\linewidth, height=0.52\textwidth]{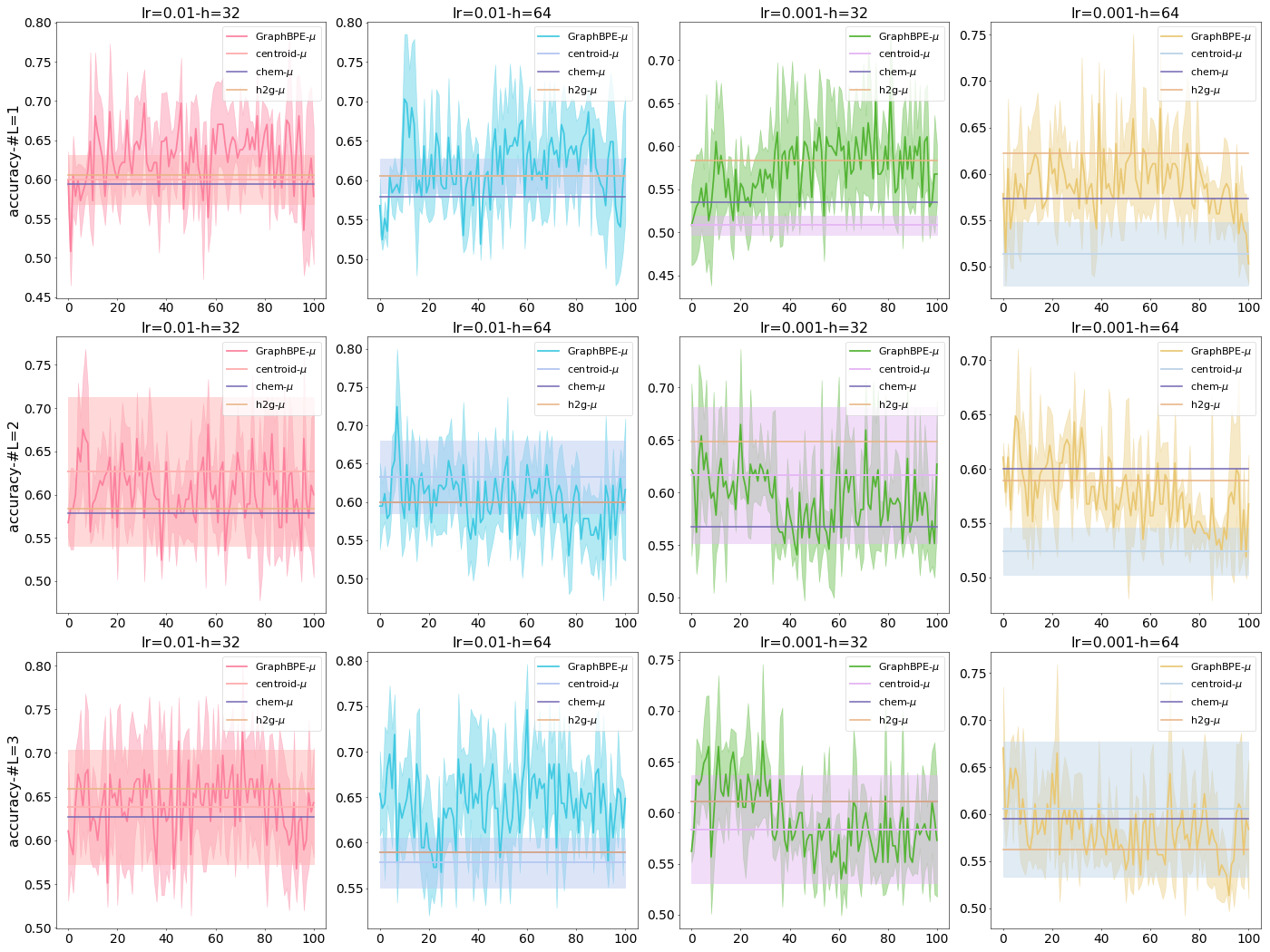}
\caption{Results of HyperConv on \textsc{Mutag}, with \textbf{accuracy} the \textit{higher} the better} 
\label{fig:mutag_hyperconv}
\end{figure}

\begin{figure}[H]
\centering
\includegraphics[width=0.9\linewidth, height=0.59\textwidth]{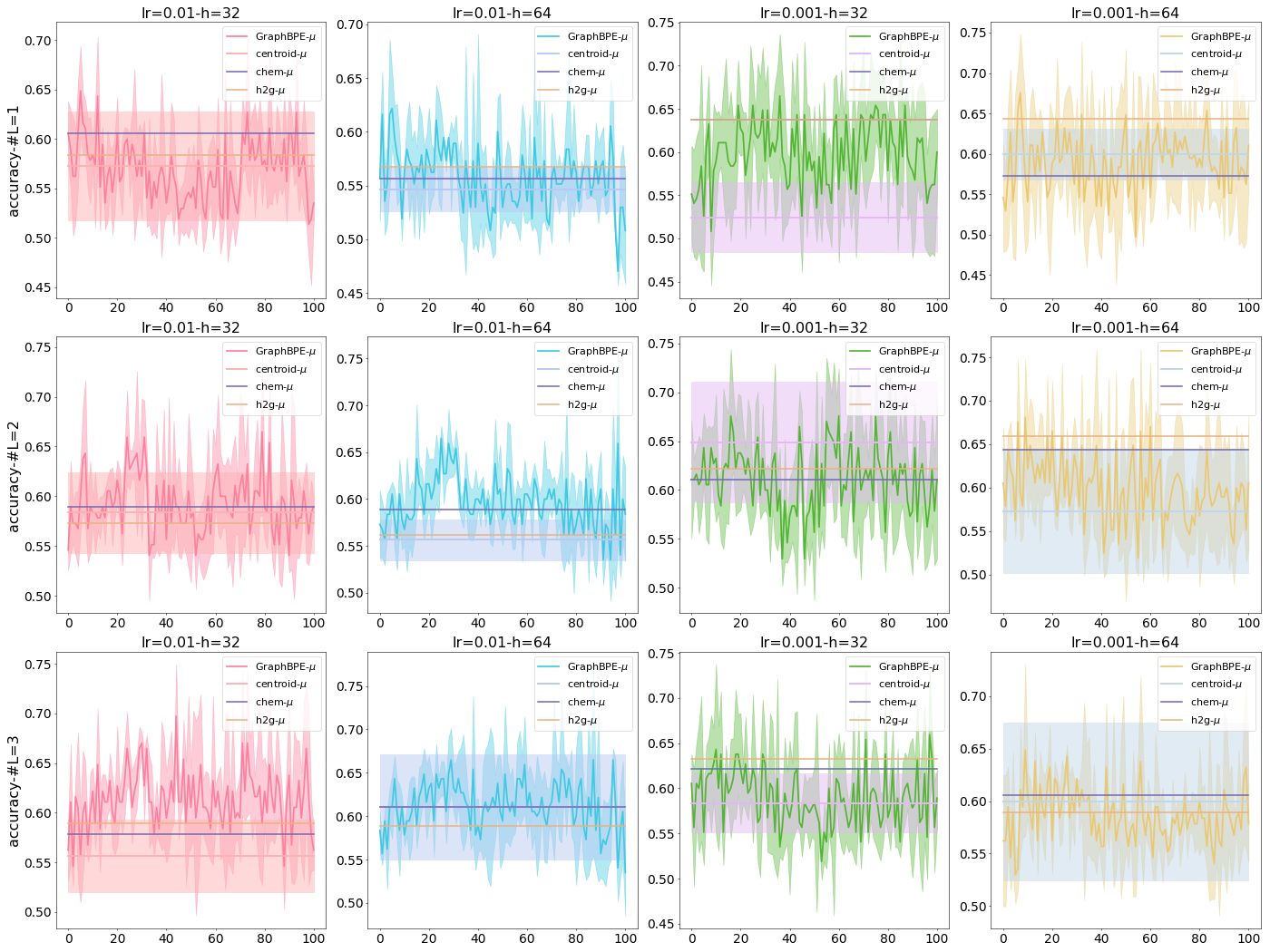}
\caption{Results of HGNN++ on \textsc{Mutag}, with \textbf{accuracy} the \textit{higher} the better} 
\label{fig:mutag_hgnnp}
\end{figure}

\begin{figure}[H]
\centering
\includegraphics[width=0.9\linewidth, height=0.59\textwidth]{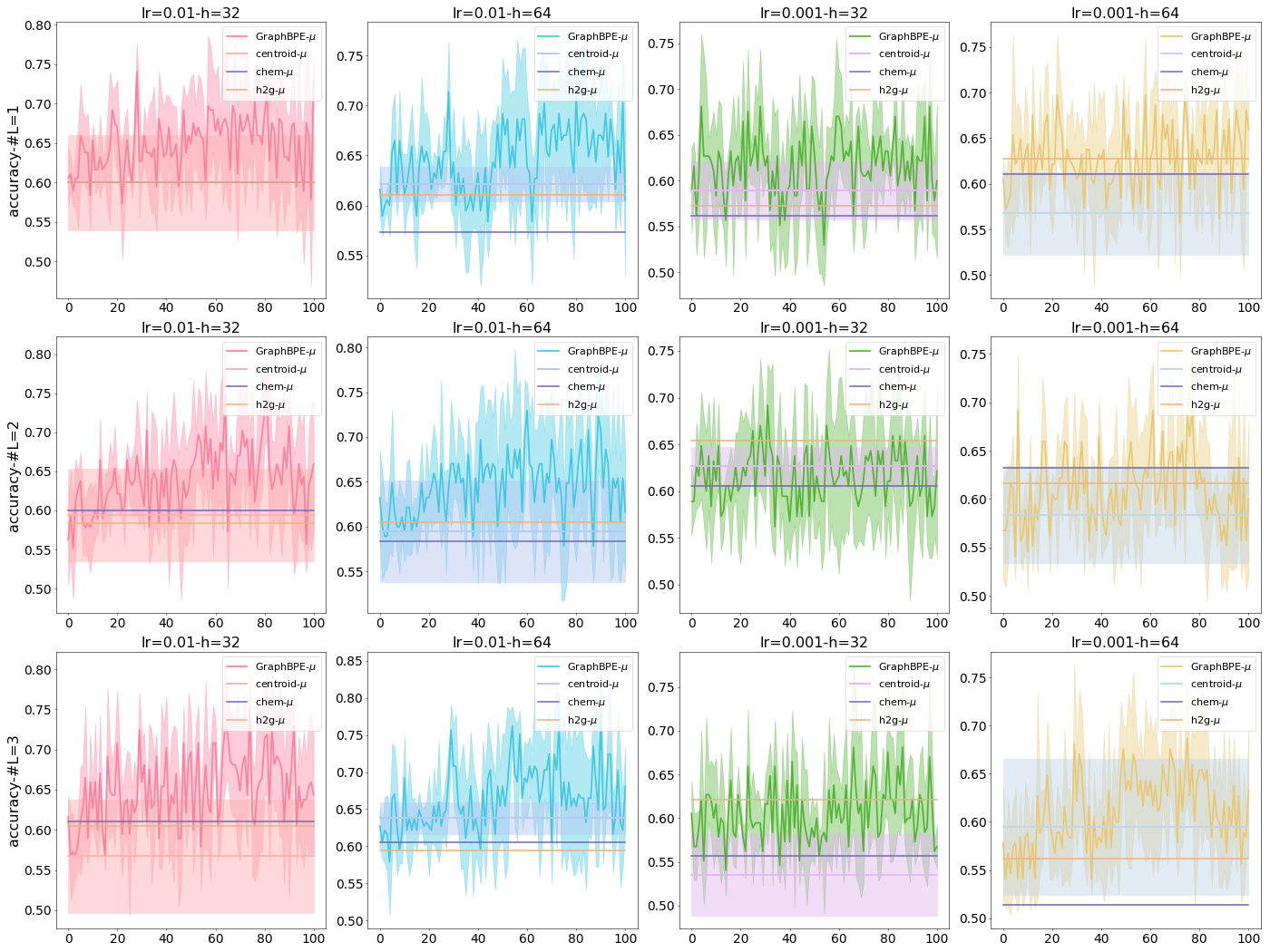}
\caption{Results of HNHN on \textsc{Mutag}, with \textbf{accuracy} the \textit{higher} the better} 
\label{fig:mutag_hnhn}
\end{figure}

For HyperGNNs, we include the performance comparison results in Table~\ref{tab: p_metric_3hgnns_on_mutag}, and the visualization over different tokenization steps in Figure~\ref{fig:mutag_hyperconv}, ~\ref{fig:mutag_hgnnp}, and ~\ref{fig:mutag_hnhn} for HyperConv, HGNN++, and HNHN.
\begin{table}[h]
    \centering
    \begin{minipage}[b]{0.45\linewidth}
        \centering
        \begin{subtable} % {.5\textwidth}
            \centering
	\resizebox{.99\textwidth}{!}{\begin{tabular}{lccccc}
	\toprule
	\multicolumn{2}{c}{\textbf{learning rate}} & \multicolumn{2}{c}{$10^{-2}$} & \multicolumn{2}{c}{$10^{-3}$} \\
	\multicolumn{2}{c}{\textbf{hidden size}} & $h=32$ & $h=64$ & $h=32$ & $h=64$ \\
	\midrule
	\multirow{2}{*}{$L={1}$} & p-value &\textcolor{red}{\textbf{86}}:15:0 & \textcolor{red}{\textbf{93}}:8:0 & 37:\textcolor{red}{\textbf{64}}:0 & \textcolor{red}{\textbf{95}}:6:0 \\
	 & metric &\textbf{101}:0:0 & \textbf{101}:0:0 & \textbf{97}:1:3 & \textbf{101}:0:0 \\
	\midrule
	\multirow{2}{*}{$L={2}$} & p-value &\textcolor{red}{\textbf{101}}:0:0 & \textcolor{red}{\textbf{89}}:12:0 & 1:\textcolor{red}{\textbf{100}}:0 & 11:\textcolor{red}{\textbf{90}}:0 \\
	 & metric &\textbf{101}:0:0 & \textbf{101}:0:0 & \textbf{91}:0:10 & \textbf{91}:2:8 \\
	\midrule
	\multirow{2}{*}{$L={3}$} & p-value &\textcolor{red}{\textbf{93}}:8:0 & \textcolor{red}{\textbf{100}}:1:0 & 1:\textcolor{red}{\textbf{100}}:0 & 36:\textcolor{red}{\textbf{65}}:0 \\
	 & metric &\textbf{101}:0:0 & \textbf{101}:0:0 & \textbf{77}:4:20 & \textbf{101}:0:0 \\
	\bottomrule
	\end{tabular}}
	\caption{Comparison with p-/metric value of GCN}
	\label{tab:p_metric_val_GCN_Mutag}
        \end{subtable}
        
        \vspace{0.3cm} % adjust vertical space between tables
        
        \begin{subtable} %{.5\textwidth}
            \centering
	\resizebox{.99\textwidth}{!}{\begin{tabular}{lccccc}
	\toprule
	\multicolumn{2}{c}{\textbf{learning rate}} & \multicolumn{2}{c}{$10^{-2}$} & \multicolumn{2}{c}{$10^{-3}$} \\
	\multicolumn{2}{c}{\textbf{hidden size}} & $h=32$ & $h=64$ & $h=32$ & $h=64$ \\
	\midrule
	\multirow{2}{*}{$L={1}$} & p-value &0:\textcolor{red}{\textbf{97}}:4 & 0:19:\textcolor{red}{\textbf{82}} & 17:\textcolor{red}{\textbf{84}}:0 & 2:\textcolor{red}{\textbf{99}}:0 \\
	 & metric &6:1:\textbf{94} & 0:0:\textbf{101} & \textbf{101}:0:0 & \textbf{94}:0:7 \\
	\midrule
	\multirow{2}{*}{$L={2}$} & p-value &0:18:\textcolor{red}{\textbf{83}} & 0:16:\textcolor{red}{\textbf{85}} & 0:\textcolor{red}{\textbf{101}}:0 & 0:\textcolor{red}{\textbf{93}}:8 \\
	 & metric &0:0:\textbf{101} & 1:0:\textbf{100} & \textbf{67}:0:34 & 0:0:\textbf{101} \\
	\midrule
	\multirow{2}{*}{$L={3}$} & p-value &1:\textcolor{red}{\textbf{100}}:0 & 2:\textcolor{red}{\textbf{99}}:0 & 2:\textcolor{red}{\textbf{98}}:1 & 0:\textcolor{red}{\textbf{101}}:0 \\
	 & metric &\textbf{78}:2:21 & \textbf{84}:4:13 & \textbf{74}:0:27 & \textbf{68}:4:29 \\
	\bottomrule
	\end{tabular}}
	\caption{Comparison with p-/metric value of GIN}
	\label{tab:p_metric_val_GIN_Mutag}
        \end{subtable}
    \end{minipage}
    \hspace{0.5cm}
    \begin{minipage}[b]{0.45\linewidth}
        \centering
        \begin{subtable} %{.5\textwidth}
            \centering
	\resizebox{.99\textwidth}{!}{\begin{tabular}{lccccc}
	\toprule
	\multicolumn{2}{c}{\textbf{learning rate}} & \multicolumn{2}{c}{$10^{-2}$} & \multicolumn{2}{c}{$10^{-3}$} \\
	\multicolumn{2}{c}{\textbf{hidden size}} & $h=32$ & $h=64$ & $h=32$ & $h=64$ \\
	\midrule
	\multirow{2}{*}{$L={1}$} & p-value &\textcolor{red}{\textbf{84}}:17:0 & \textcolor{red}{\textbf{99}}:2:0 & \textcolor{red}{\textbf{58}}:43:0 & 11:\textcolor{red}{\textbf{90}}:0 \\
	 & metric &\textbf{101}:0:0 & \textbf{101}:0:0 & \textbf{98}:1:2 & \textbf{96}:1:4 \\
	\midrule
	\multirow{2}{*}{$L={2}$} & p-value &\textcolor{red}{\textbf{101}}:0:0 & \textcolor{red}{\textbf{97}}:4:0 & \textcolor{red}{\textbf{63}}:38:0 & 29:\textcolor{red}{\textbf{72}}:0 \\
	 & metric &\textbf{101}:0:0 & \textbf{101}:0:0 & \textbf{100}:0:1 & \textbf{100}:0:1 \\
	\midrule
	\multirow{2}{*}{$L={3}$} & p-value &17:\textcolor{red}{\textbf{84}}:0 & \textcolor{red}{\textbf{57}}:44:0 & 45:\textcolor{red}{\textbf{56}}:0 & \textcolor{red}{\textbf{73}}:28:0 \\
	 & metric &\textbf{99}:0:2 & \textbf{101}:0:0 & \textbf{101}:0:0 & \textbf{100}:0:1 \\
	\bottomrule
	\end{tabular}}
	\caption{Comparison with p-/metric value of GAT}
	\label{tab:p_metric_val_GAT_Mutag}
        \end{subtable}
        
        \vspace{0.3cm} % adjust vertical space between tables
        
        \begin{subtable} %{.5\textwidth}
            \centering
	\resizebox{.99\textwidth}{!}{\begin{tabular}{lccccc}
	\toprule
	\multicolumn{2}{c}{\textbf{learning rate}} & \multicolumn{2}{c}{$10^{-2}$} & \multicolumn{2}{c}{$10^{-3}$} \\
	\multicolumn{2}{c}{\textbf{hidden size}} & $h=32$ & $h=64$ & $h=32$ & $h=64$ \\
	\midrule
	\multirow{2}{*}{$L={1}$} & p-value &\textcolor{red}{\textbf{100}}:1:0 & \textcolor{red}{\textbf{101}}:0:0 & \textcolor{red}{\textbf{58}}:43:0 & \textcolor{red}{\textbf{85}}:16:0 \\
	 & metric &\textbf{101}:0:0 & \textbf{101}:0:0 & \textbf{97}:0:4 & \textbf{101}:0:0 \\
	\midrule
	\multirow{2}{*}{$L={2}$} & p-value &\textcolor{red}{\textbf{101}}:0:0 & \textcolor{red}{\textbf{99}}:2:0 & \textcolor{red}{\textbf{89}}:12:0 & 36:\textcolor{red}{\textbf{65}}:0 \\
	 & metric &\textbf{101}:0:0 & \textbf{101}:0:0 & \textbf{101}:0:0 & \textbf{101}:0:0 \\
	\midrule
	\multirow{2}{*}{$L={3}$} & p-value &\textcolor{red}{\textbf{96}}:5:0 & \textcolor{red}{\textbf{79}}:22:0 & \textcolor{red}{\textbf{100}}:1:0 & \textcolor{red}{\textbf{101}}:0:0 \\
	 & metric &\textbf{101}:0:0 & \textbf{101}:0:0 & \textbf{101}:0:0 & \textbf{101}:0:0 \\
	\bottomrule
	\end{tabular}}
	\caption{Comparison with p-/metric value of GraphSAGE}
	\label{tab:p_metric_val_GraphSAGE_Mutag}
        \end{subtable}
    \end{minipage}
    \caption{Performance comparison on accuracy with p-value $<0.05$ and metric value on \textsc{Mutag}. For each triplet $a$:$b$:$c$, $a, b, c$ denote the number of times \textsc{GraphBPE} is \textcolor{red}{\textbf{statistically}}/\textbf{numerically} better/the same/worse compared with (untokenized) simple graph.}
    \label{tab: p_metric_4gnns_on_mutag}
\end{table}
\newpage
\FloatBarrier  % Prevent floats from moving past this point
%%%%%%%%%%%%%%%%%%%%%%%%%%%%%%%%%%%%%%%%%
\FloatBarrier  % Prevent floats from moving past this point
%%%%%%%%%%%%%%%%%%%%%%%%%%%%%%%%%%%%%%%%%
\FloatBarrier  % Prevent floats from moving past this point
%%%%%%%%%%%%%%%%%%%%%%%%%%%%%%%%%%%%%%%%%
\FloatBarrier  % Prevent floats from moving past this point
\newpage
\begin{table}[h]
    \centering
    \begin{minipage}[b]{0.32\linewidth}
        \centering
        \begin{subtable}
            \centering
	\resizebox{.99\textwidth}{!}{\begin{tabular}{lccccc}
	\toprule
	\multicolumn{2}{c}{\textbf{learning rate}} & \multicolumn{2}{c}{$10^{-2}$} & \multicolumn{2}{c}{$10^{-3}$} \\
	\multicolumn{2}{c}{\textbf{hidden size}} & $h=32$ & $h=64$ & $h=32$ & $h=64$ \\
	\midrule
	\multirow{2}{*}{$L={1}$} & p-value &12:\textcolor{red}{\textbf{88}}:1 & 5:\textcolor{red}{\textbf{89}}:7 & 44:\textcolor{red}{\textbf{57}}:0 & \textcolor{red}{\textbf{51}}:50:0 \\
	 & metric &\textbf{82}:2:17 & \textbf{56}:5:40 & \textbf{101}:0:0 & \textbf{100}:0:1 \\
	\midrule
	\multirow{2}{*}{$L={2}$} & p-value &0:\textcolor{red}{\textbf{100}}:1 & 0:\textcolor{red}{\textbf{91}}:10 & 0:\textcolor{red}{\textbf{101}}:0 & 47:\textcolor{red}{\textbf{54}}:0 \\
	 & metric &26:2:\textbf{73} & 17:0:\textbf{84} & 21:3:\textbf{77} & \textbf{98}:2:1 \\
	\midrule
	\multirow{2}{*}{$L={3}$} & p-value &0:\textcolor{red}{\textbf{99}}:2 & 40:\textcolor{red}{\textbf{61}}:0 & 4:\textcolor{red}{\textbf{97}}:0 & 0:\textcolor{red}{\textbf{101}}:0 \\
	 & metric &\textbf{58}:5:38 & \textbf{96}:2:3 & \textbf{59}:0:42 & 17:1:\textbf{83} \\
	\bottomrule
	\end{tabular}}
	\caption{\textsc{Centroid} on HyperConv}
	\label{tab:p_metric_val_HyperConv_Mutag_Centroid}
        \end{subtable}
        
        \vspace{0.25cm} % adjust vertical space between tables
        
        \begin{subtable}
            \centering
	\resizebox{.99\textwidth}{!}{\begin{tabular}{lccccc}
	\toprule
	\multicolumn{2}{c}{\textbf{learning rate}} & \multicolumn{2}{c}{$10^{-2}$} & \multicolumn{2}{c}{$10^{-3}$} \\
	\multicolumn{2}{c}{\textbf{hidden size}} & $h=32$ & $h=64$ & $h=32$ & $h=64$ \\
	\midrule
	\multirow{2}{*}{$L={1}$} & p-value &18:\textcolor{red}{\textbf{82}}:1 & 16:\textcolor{red}{\textbf{83}}:2 & 10:\textcolor{red}{\textbf{91}}:0 & 3:\textcolor{red}{\textbf{96}}:2 \\
	 & metric &\textbf{84}:1:16 & \textbf{89}:0:12 & \textbf{84}:3:14 & \textbf{71}:5:25 \\
	\midrule
	\multirow{2}{*}{$L={2}$} & p-value &8:\textcolor{red}{\textbf{92}}:1 & 2:\textcolor{red}{\textbf{97}}:2 & 4:\textcolor{red}{\textbf{97}}:0 & 0:\textcolor{red}{\textbf{87}}:14 \\
	 & metric &\textbf{76}:4:21 & \textbf{59}:1:41 & \textbf{83}:0:18 & 23:1:\textbf{77} \\
	\midrule
	\multirow{2}{*}{$L={3}$} & p-value &4:\textcolor{red}{\textbf{95}}:2 & 31:\textcolor{red}{\textbf{70}}:0 & 0:\textcolor{red}{\textbf{94}}:7 & 0:\textcolor{red}{\textbf{100}}:1 \\
	 & metric &\textbf{73}:0:28 & \textbf{93}:1:7 & 27:1:\textbf{73} & 29:6:\textbf{66} \\
	\bottomrule
	\end{tabular}}
	\caption{\textsc{Chem} on HyperConv}
	\label{tab:p_metric_val_HyperConv_Mutag_Chem}
        \end{subtable}

        \vspace{0.25cm} % adjust vertical space between tables
        
        \begin{subtable}
            \centering
	\resizebox{.99\textwidth}{!}{\begin{tabular}{lccccc}
	\toprule
	\multicolumn{2}{c}{\textbf{learning rate}} & \multicolumn{2}{c}{$10^{-2}$} & \multicolumn{2}{c}{$10^{-3}$} \\
	\multicolumn{2}{c}{\textbf{hidden size}} & $h=32$ & $h=64$ & $h=32$ & $h=64$ \\
	\midrule
	\multirow{2}{*}{$L={1}$} & p-value &7:\textcolor{red}{\textbf{92}}:2 & 7:\textcolor{red}{\textbf{89}}:5 & 1:\textcolor{red}{\textbf{94}}:6 & 0:\textcolor{red}{\textbf{93}}:8 \\
	 & metric &\textbf{75}:2:24 & \textbf{56}:5:40 & 42:0:\textbf{59} & 12:2:\textbf{87} \\
	\midrule
	\multirow{2}{*}{$L={2}$} & p-value &7:\textcolor{red}{\textbf{91}}:3 & 3:\textcolor{red}{\textbf{94}}:4 & 0:\textcolor{red}{\textbf{95}}:6 & 1:\textcolor{red}{\textbf{92}}:8 \\
	 & metric &\textbf{70}:4:27 & \textbf{56}:3:42 & 3:0:\textbf{98} & 35:3:\textbf{63} \\
	\midrule
	\multirow{2}{*}{$L={3}$} & p-value &1:\textcolor{red}{\textbf{96}}:4 & 34:\textcolor{red}{\textbf{67}}:0 & 1:\textcolor{red}{\textbf{85}}:15 & 3:\textcolor{red}{\textbf{97}}:1 \\
	 & metric &33:1:\textbf{67} & \textbf{93}:1:7 & 28:0:\textbf{73} & \textbf{75}:5:21 \\
	\bottomrule
	\end{tabular}}
	\caption{\textsc{H2g} on HyperConv}
	\label{tab:p_metric_val_HyperConv_Mutag_H2g}
        \end{subtable}
    \end{minipage}
    \hfill
    \begin{minipage}[b]{0.32\linewidth}
        \centering
        \begin{subtable}
            \centering
	\resizebox{.99\textwidth}{!}{\begin{tabular}{lccccc}
	\toprule
	\multicolumn{2}{c}{\textbf{learning rate}} & \multicolumn{2}{c}{$10^{-2}$} & \multicolumn{2}{c}{$10^{-3}$} \\
	\multicolumn{2}{c}{\textbf{hidden size}} & $h=32$ & $h=64$ & $h=32$ & $h=64$ \\
	\midrule
	\multirow{2}{*}{$L={1}$} & p-value &1:\textcolor{red}{\textbf{100}}:0 & 9:\textcolor{red}{\textbf{88}}:4 & 35:\textcolor{red}{\textbf{66}}:0 & 0:\textcolor{red}{\textbf{96}}:5 \\
	 & metric &39:3:\textbf{59} & \textbf{70}:2:29 & \textbf{99}:1:1 & 39:6:\textbf{56} \\
	\midrule
	\multirow{2}{*}{$L={2}$} & p-value &1:\textcolor{red}{\textbf{100}}:0 & 39:\textcolor{red}{\textbf{62}}:0 & 0:\textcolor{red}{\textbf{93}}:8 & 2:\textcolor{red}{\textbf{99}}:0 \\
	 & metric &\textbf{57}:5:39 & \textbf{97}:1:3 & 11:0:\textbf{90} & \textbf{77}:1:23 \\
	\midrule
	\multirow{2}{*}{$L={3}$} & p-value &25:\textcolor{red}{\textbf{76}}:0 & 0:\textcolor{red}{\textbf{100}}:1 & 1:\textcolor{red}{\textbf{95}}:5 & 0:\textcolor{red}{\textbf{101}}:0 \\
	 & metric &\textbf{98}:0:3 & \textbf{51}:5:45 & \textbf{54}:3:44 & 27:1:\textbf{73} \\
	\bottomrule
	\end{tabular}}
	\caption{\textsc{Centroid} on HGNN++}
	\label{tab:p_metric_val_HGNN++_Mutag_Centroid}
        \end{subtable}
        
        \vspace{0.32cm} % adjust vertical space between tables
        
        \begin{subtable}
            \centering
	\resizebox{.99\textwidth}{!}{\begin{tabular}{lccccc}
	\toprule
	\multicolumn{2}{c}{\textbf{learning rate}} & \multicolumn{2}{c}{$10^{-2}$} & \multicolumn{2}{c}{$10^{-3}$} \\
	\multicolumn{2}{c}{\textbf{hidden size}} & $h=32$ & $h=64$ & $h=32$ & $h=64$ \\
	\midrule
	\multirow{2}{*}{$L={1}$} & p-value &0:\textcolor{red}{\textbf{78}}:23 & 7:\textcolor{red}{\textbf{85}}:9 & 0:\textcolor{red}{\textbf{98}}:3 & 2:\textcolor{red}{\textbf{99}}:0 \\
	 & metric &7:0:\textbf{94} & \textbf{51}:5:45 & 18:2:\textbf{81} & \textbf{69}:4:28 \\
	\midrule
	\multirow{2}{*}{$L={2}$} & p-value &4:\textcolor{red}{\textbf{93}}:4 & 7:\textcolor{red}{\textbf{93}}:1 & 1:\textcolor{red}{\textbf{94}}:6 & 0:\textcolor{red}{\textbf{92}}:9 \\
	 & metric &45:5:\textbf{51} & \textbf{52}:3:46 & \textbf{54}:3:44 & 12:0:\textbf{89} \\
	\midrule
	\multirow{2}{*}{$L={3}$} & p-value &23:\textcolor{red}{\textbf{78}}:0 & 1:\textcolor{red}{\textbf{98}}:2 & 0:\textcolor{red}{\textbf{83}}:18 & 0:\textcolor{red}{\textbf{100}}:1 \\
	 & metric &\textbf{91}:0:10 & 49:2:\textbf{50} & 13:4:\textbf{84} & 18:0:\textbf{83} \\
	\bottomrule
	\end{tabular}}
	\caption{\textsc{Chem} on HGNN++}
	\label{tab:p_metric_val_HGNN++_Mutag_Chem}
        \end{subtable}

        \vspace{0.32cm} % adjust vertical space between tables
        
        \begin{subtable}
            \centering
	\resizebox{.99\textwidth}{!}{\begin{tabular}{lccccc}
	\toprule
	\multicolumn{2}{c}{\textbf{learning rate}} & \multicolumn{2}{c}{$10^{-2}$} & \multicolumn{2}{c}{$10^{-3}$} \\
	\multicolumn{2}{c}{\textbf{hidden size}} & $h=32$ & $h=64$ & $h=32$ & $h=64$ \\
	\midrule
	\multirow{2}{*}{$L={1}$} & p-value &0:\textcolor{red}{\textbf{87}}:14 & 0:\textcolor{red}{\textbf{91}}:10 & 0:\textcolor{red}{\textbf{89}}:12 & 0:\textcolor{red}{\textbf{79}}:22 \\
	 & metric &25:2:\textbf{74} & 35:6:\textbf{60} & 20:0:\textbf{81} & 5:1:\textbf{95} \\
	\midrule
	\multirow{2}{*}{$L={2}$} & p-value &5:\textcolor{red}{\textbf{96}}:0 & 35:\textcolor{red}{\textbf{66}}:0 & 0:\textcolor{red}{\textbf{99}}:2 & 0:\textcolor{red}{\textbf{81}}:20 \\
	 & metric &\textbf{74}:5:22 & \textbf{90}:0:11 & 39:5:\textbf{57} & 5:2:\textbf{94} \\
	\midrule
	\multirow{2}{*}{$L={3}$} & p-value &4:\textcolor{red}{\textbf{97}}:0 & 12:\textcolor{red}{\textbf{89}}:0 & 0:\textcolor{red}{\textbf{99}}:2 & 0:\textcolor{red}{\textbf{101}}:0 \\
	 & metric &\textbf{83}:3:15 & \textbf{80}:3:18 & 8:2:\textbf{91} & 40:7:\textbf{54} \\
	\bottomrule
	\end{tabular}}
	\caption{\textsc{H2g} on HGNN++}
	\label{tab:p_metric_val_HGNN++_Mutag_H2g}
        \end{subtable}
    \end{minipage}
    \hfill
    \begin{minipage}[b]{0.32\linewidth}
        \centering
        \begin{subtable}{}
            \centering
	\resizebox{.99\textwidth}{!}{\begin{tabular}{lccccc}
	\toprule
	\multicolumn{2}{c}{\textbf{learning rate}} & \multicolumn{2}{c}{$10^{-2}$} & \multicolumn{2}{c}{$10^{-3}$} \\
	\multicolumn{2}{c}{\textbf{hidden size}} & $h=32$ & $h=64$ & $h=32$ & $h=64$ \\
	\midrule
	\multirow{2}{*}{$L={1}$} & p-value &6:\textcolor{red}{\textbf{95}}:0 & 11:\textcolor{red}{\textbf{90}}:0 & 13:\textcolor{red}{\textbf{88}}:0 & 27:\textcolor{red}{\textbf{74}}:0 \\
	 & metric &\textbf{93}:0:8 & \textbf{76}:2:23 & \textbf{79}:4:18 & \textbf{99}:0:2 \\
	\midrule
	\multirow{2}{*}{$L={2}$} & p-value &9:\textcolor{red}{\textbf{92}}:0 & 10:\textcolor{red}{\textbf{91}}:0 & 0:\textcolor{red}{\textbf{96}}:5 & 6:\textcolor{red}{\textbf{95}}:0 \\
	 & metric &\textbf{84}:0:17 & \textbf{94}:2:5 & 38:1:\textbf{62} & \textbf{75}:2:24 \\
	\midrule
	\multirow{2}{*}{$L={3}$} & p-value &20:\textcolor{red}{\textbf{81}}:0 & 9:\textcolor{red}{\textbf{91}}:1 & 24:\textcolor{red}{\textbf{77}}:0 & 2:\textcolor{red}{\textbf{99}}:0 \\
	 & metric &\textbf{100}:1:0 & \textbf{70}:2:29 & \textbf{101}:0:0 & \textbf{61}:2:38 \\
	\bottomrule
	\end{tabular}}
	\caption{\textsc{Centroid} on HNHN}
	\label{tab:p_metric_val_HNHN_Mutag_Centroid}
        \end{subtable}
        
        \vspace{0.32cm} % adjust vertical space between tables
        
        \begin{subtable}
            \centering
	\resizebox{.99\textwidth}{!}{\begin{tabular}{lccccc}
	\toprule
	\multicolumn{2}{c}{\textbf{learning rate}} & \multicolumn{2}{c}{$10^{-2}$} & \multicolumn{2}{c}{$10^{-3}$} \\
	\multicolumn{2}{c}{\textbf{hidden size}} & $h=32$ & $h=64$ & $h=32$ & $h=64$ \\
	\midrule
	\multirow{2}{*}{$L={1}$} & p-value &19:\textcolor{red}{\textbf{82}}:0 & 39:\textcolor{red}{\textbf{62}}:0 & 9:\textcolor{red}{\textbf{92}}:0 & 2:\textcolor{red}{\textbf{99}}:0 \\
	 & metric &\textbf{91}:2:8 & \textbf{101}:0:0 & \textbf{97}:0:4 & \textbf{69}:4:28 \\
	\midrule
	\multirow{2}{*}{$L={2}$} & p-value &9:\textcolor{red}{\textbf{91}}:1 & 9:\textcolor{red}{\textbf{92}}:0 & 0:\textcolor{red}{\textbf{101}}:0 & 0:\textcolor{red}{\textbf{100}}:1 \\
	 & metric &\textbf{83}:1:17 & \textbf{99}:0:2 & \textbf{62}:6:33 & 27:0:\textbf{74} \\
	\midrule
	\multirow{2}{*}{$L={3}$} & p-value &3:\textcolor{red}{\textbf{98}}:0 & 11:\textcolor{red}{\textbf{90}}:0 & 16:\textcolor{red}{\textbf{85}}:0 & \textcolor{red}{\textbf{67}}:34:0 \\
	 & metric &\textbf{80}:3:18 & \textbf{98}:0:3 & \textbf{96}:1:4 & \textbf{101}:0:0 \\
	\bottomrule
	\end{tabular}}
	\caption{\textsc{Chem} on HNHN}
	\label{tab:p_metric_val_HNHN_Mutag_Chem}
        \end{subtable}

        \vspace{0.32cm} % adjust vertical space between tables
        
        \begin{subtable}
            \centering
	\resizebox{.99\textwidth}{!}{\begin{tabular}{lccccc}
	\toprule
	\multicolumn{2}{c}{\textbf{learning rate}} & \multicolumn{2}{c}{$10^{-2}$} & \multicolumn{2}{c}{$10^{-3}$} \\
	\multicolumn{2}{c}{\textbf{hidden size}} & $h=32$ & $h=64$ & $h=32$ & $h=64$ \\
	\midrule
	\multirow{2}{*}{$L={1}$} & p-value &13:\textcolor{red}{\textbf{88}}:0 & 16:\textcolor{red}{\textbf{85}}:0 & 17:\textcolor{red}{\textbf{84}}:0 & 2:\textcolor{red}{\textbf{98}}:1 \\
	 & metric &\textbf{93}:0:8 & \textbf{83}:3:15 & \textbf{92}:0:9 & \textbf{51}:4:46 \\
	\midrule
	\multirow{2}{*}{$L={2}$} & p-value &17:\textcolor{red}{\textbf{83}}:1 & 7:\textcolor{red}{\textbf{94}}:0 & 0:\textcolor{red}{\textbf{97}}:4 & 3:\textcolor{red}{\textbf{95}}:3 \\
	 & metric &\textbf{90}:3:8 & \textbf{89}:0:12 & 7:0:\textbf{94} & 39:3:\textbf{59} \\
	\midrule
	\multirow{2}{*}{$L={3}$} & p-value &11:\textcolor{red}{\textbf{89}}:1 & 28:\textcolor{red}{\textbf{73}}:0 & 0:\textcolor{red}{\textbf{93}}:8 & 14:\textcolor{red}{\textbf{87}}:0 \\
	 & metric &\textbf{85}:1:15 & \textbf{99}:0:2 & 32:2:\textbf{67} & \textbf{93}:0:8 \\
	\bottomrule
	\end{tabular}}
	\caption{\textsc{H2g} on HNHN}
	\label{tab:p_metric_val_HNHN_Mutag_H2g}
        \end{subtable}
    \end{minipage}
    \caption{Performance comparison on accuracy with p-value $<0.05$ and metric value on \textsc{Mutag}. For each triplet $a$:$b$:$c$, $a, b, c$ denote the number of times \textsc{GraphBPE} is \textcolor{red}{\textbf{statistically}}/\textbf{numerically} better/the same/worse compared with hypergraphs constructed by \textsc{Method} on Model (e.g., ``\textsc{Centroud} on HyperConv'' means comparing \textsc{GraphBPE} with \textsc{Centroid} on the HyperConv model).}
    \label{tab: p_metric_3hgnns_on_mutag}
\end{table}
\FloatBarrier  % Prevent floats from moving past this point
\newpage
\FloatBarrier  % Prevent floats from moving past this point
\FloatBarrier  % Prevent floats from moving past this point

%%%%%%%%%%%%%%%%%%%%%%%%%%%%%%%%%%%%%%%%%
\newpage

\subsection{\textsc{Enzymes}}\label{app: enzymes_result}
\begin{figure}[H]
\centering
\includegraphics[width=0.9\linewidth, height=0.59\textwidth]{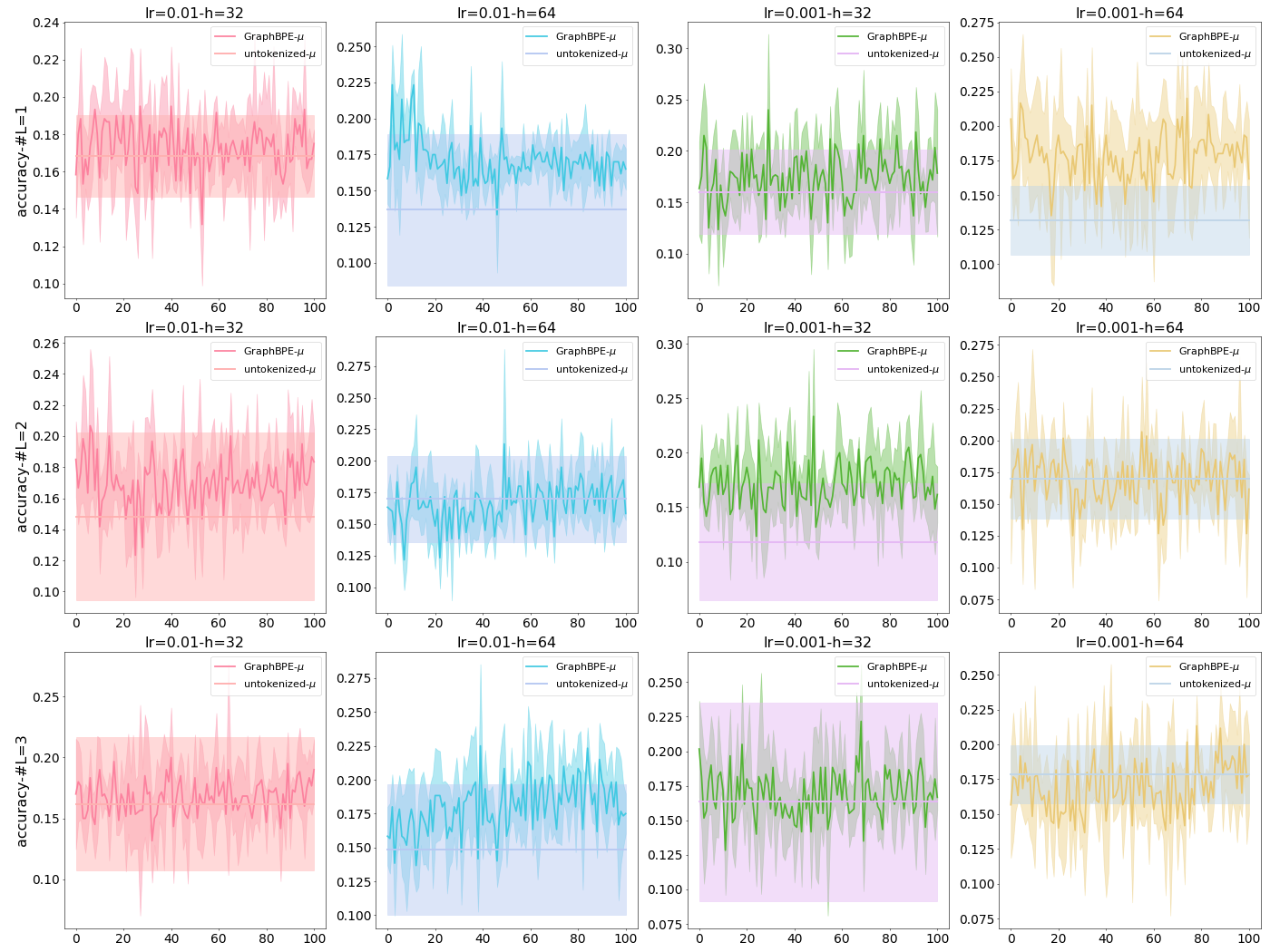}
\caption{Results of GCN on \textsc{Enzymes}, with \textbf{accuracy} the \textit{higher} the better} 
\label{fig:enzymes_gcn}
\end{figure}

\begin{figure}[H]
\centering
\includegraphics[width=0.9\linewidth, height=0.59\textwidth]{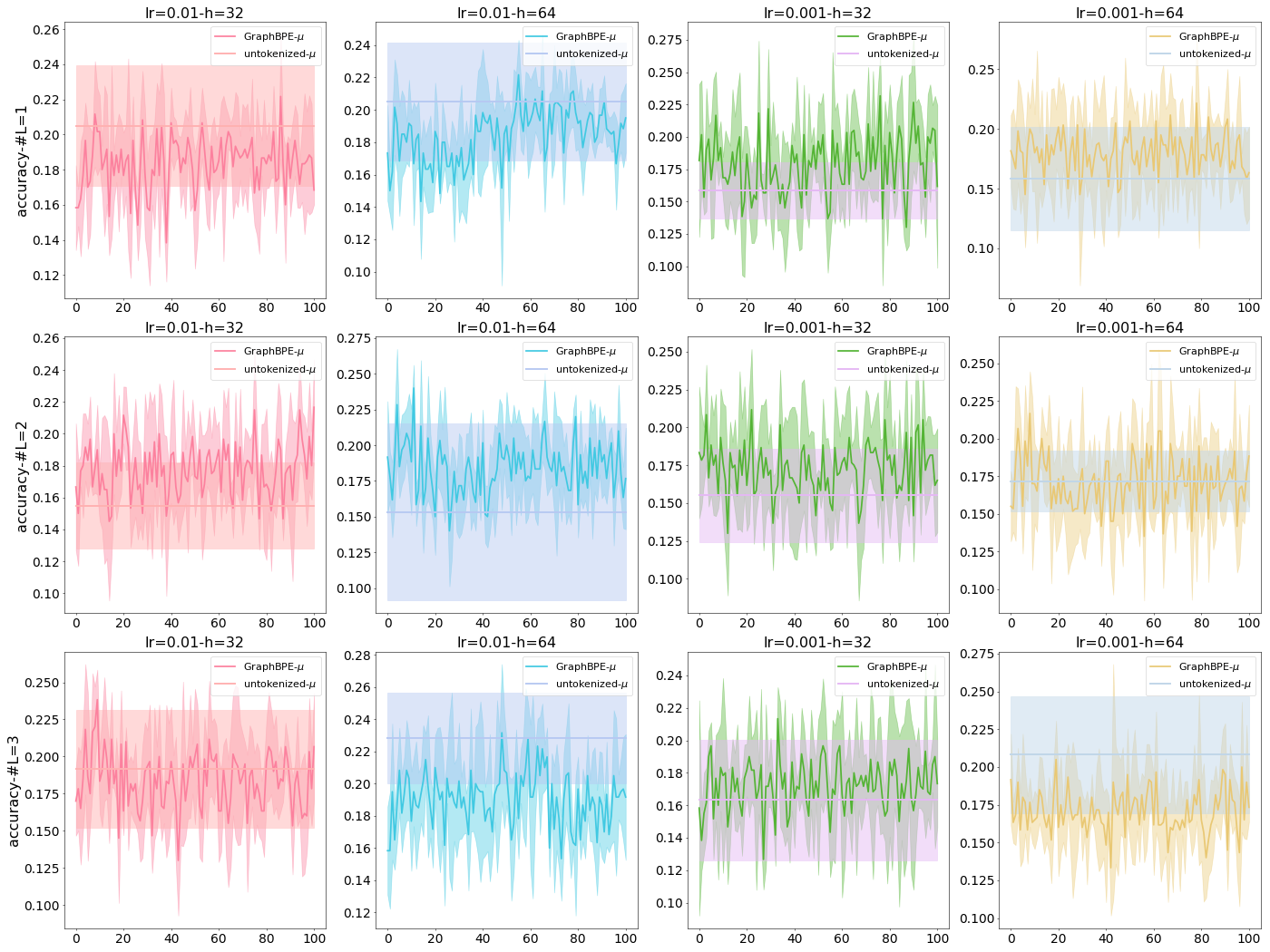}
\caption{Results of GAT on \textsc{Enzymes}, with \textbf{accuracy} the \textit{higher} the better} 
\label{fig:enzymes_gat}
\end{figure}

\begin{figure}[H]
\centering
\includegraphics[width=0.9\linewidth, height=0.59\textwidth]{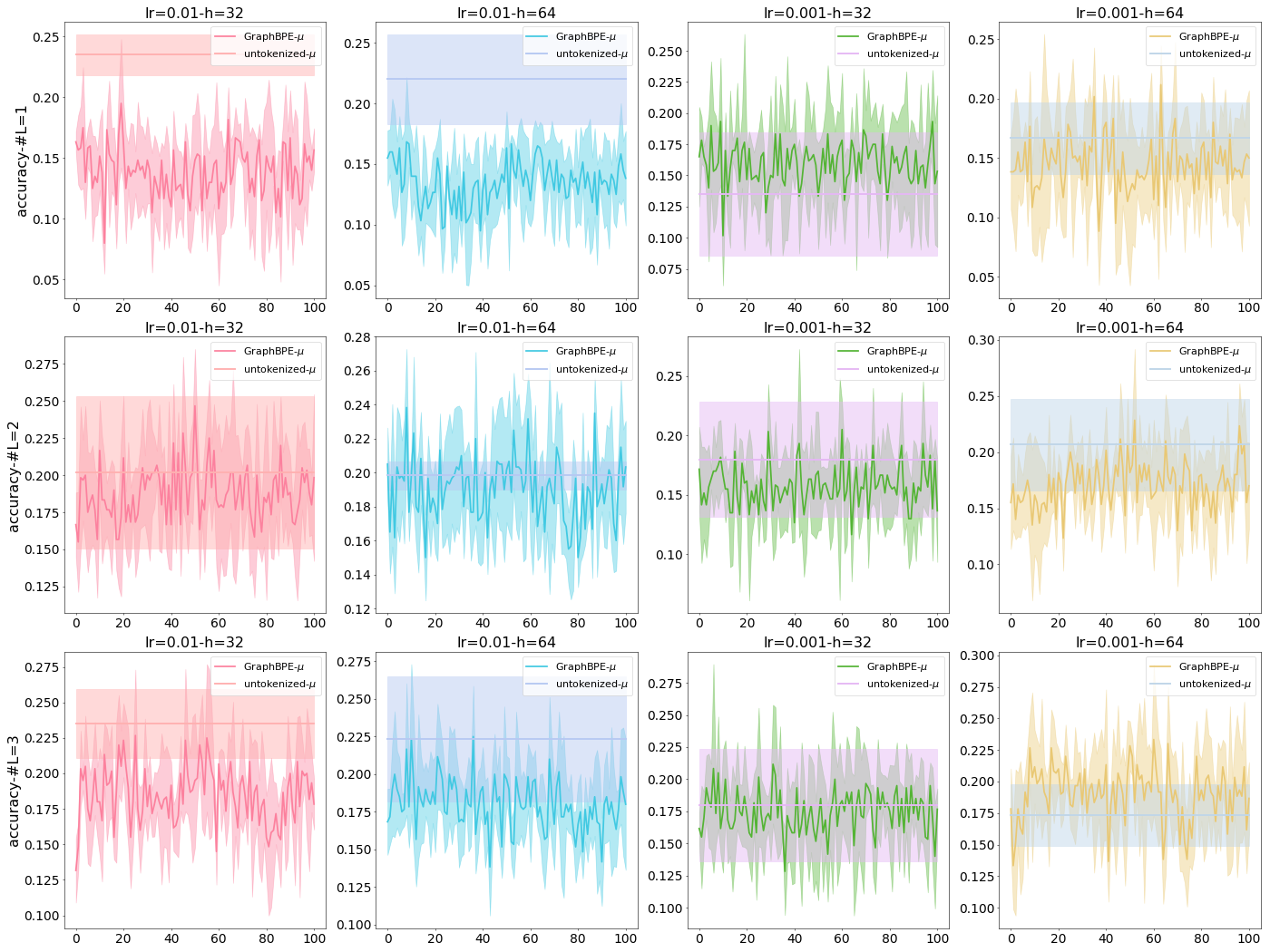}
\caption{Results of GIN on \textsc{Enzymes}, with \textbf{accuracy} the \textit{higher} the better} 
\label{fig:enzymes_gin}
\end{figure}

\begin{figure}[H]
\centering
\includegraphics[width=0.9\linewidth, height=0.59\textwidth]{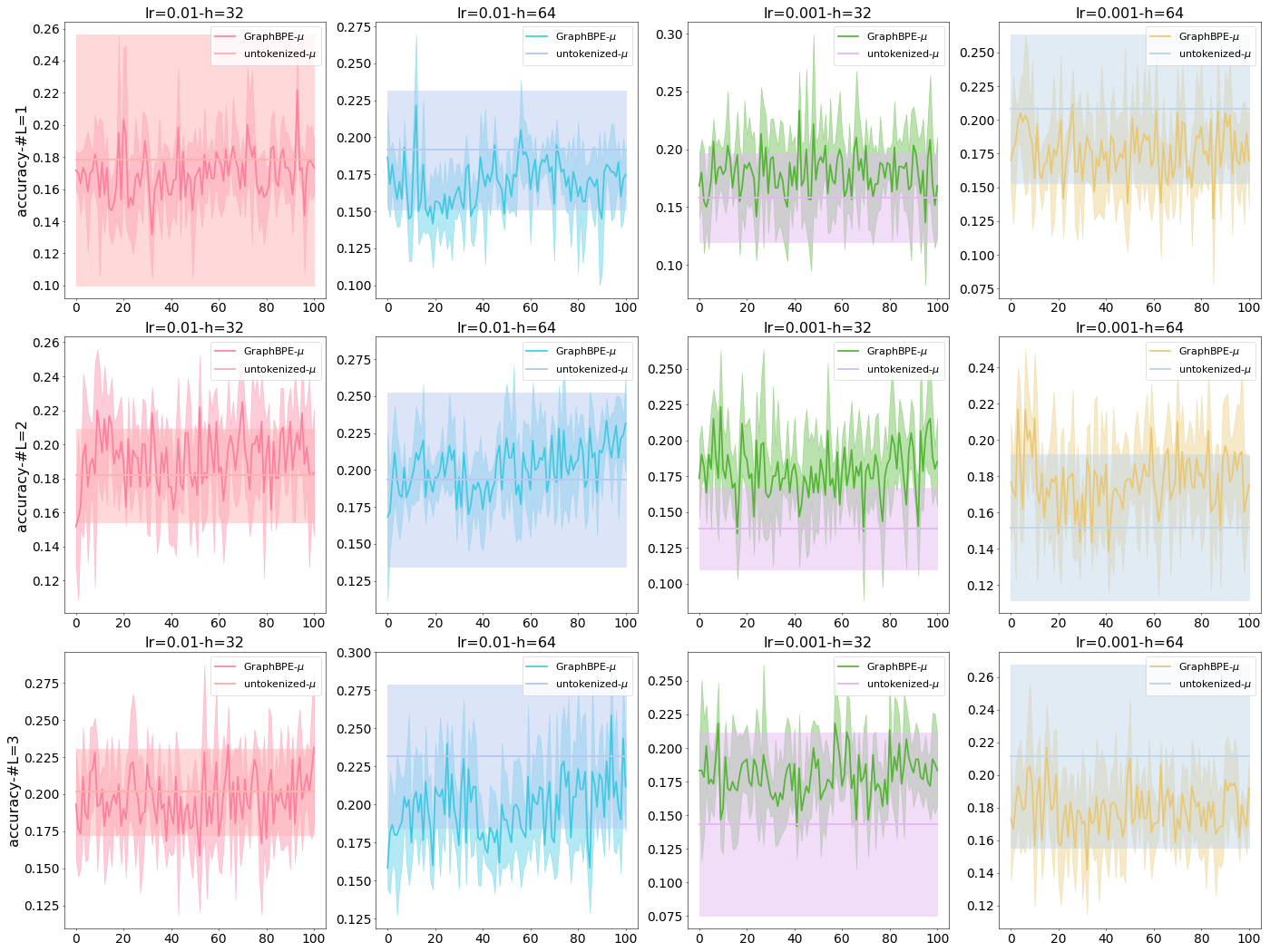}
\caption{Results of GraphSAGE on \textsc{Enzymes}, with \textbf{accuracy} the \textit{higher} the better} 
\label{fig:enzymes_graphsage}
\end{figure}

For GNNs, we include the performance comparison results in Table~\ref{tab: p_metric_4gnns_on_enzymes}, and the visualization over different tokenization steps in Figure~\ref{fig:enzymes_gcn}, ~\ref{fig:enzymes_gat}, ~\ref{fig:enzymes_gin}, and ~\ref{fig:enzymes_graphsage} for GCN, GAT, GIN, and GraphSAGE.

\begin{figure}[H]
\centering
\includegraphics[width=0.9\linewidth, height=0.59\textwidth]{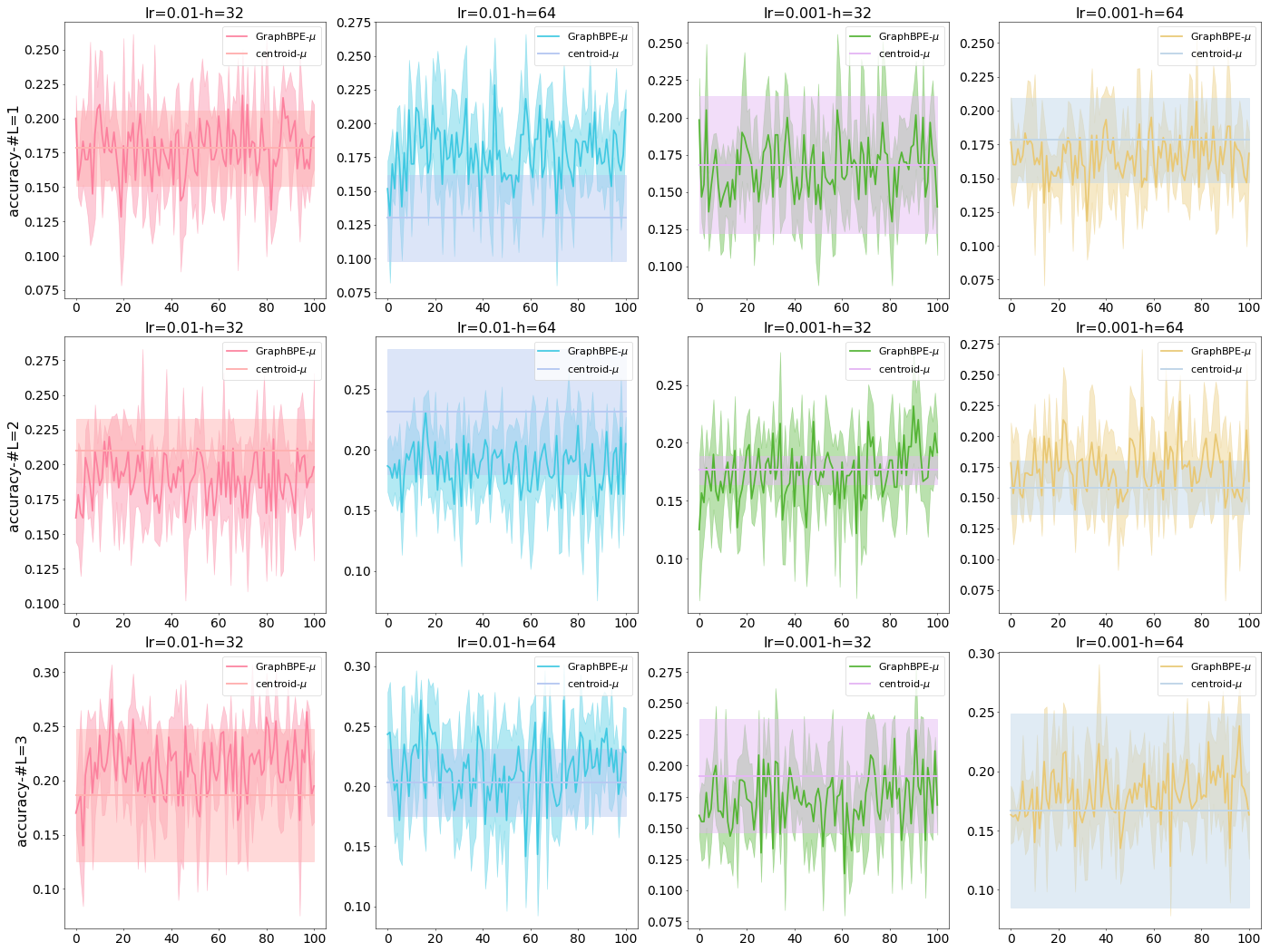}
\caption{Results of HyperConv on \textsc{Enzymes}, with \textbf{accuracy} the \textit{higher} the better} 
\label{fig:enzymes_hyperconv}
\end{figure}

\begin{figure}[H]
\centering
\includegraphics[width=0.9\linewidth, height=0.59\textwidth]{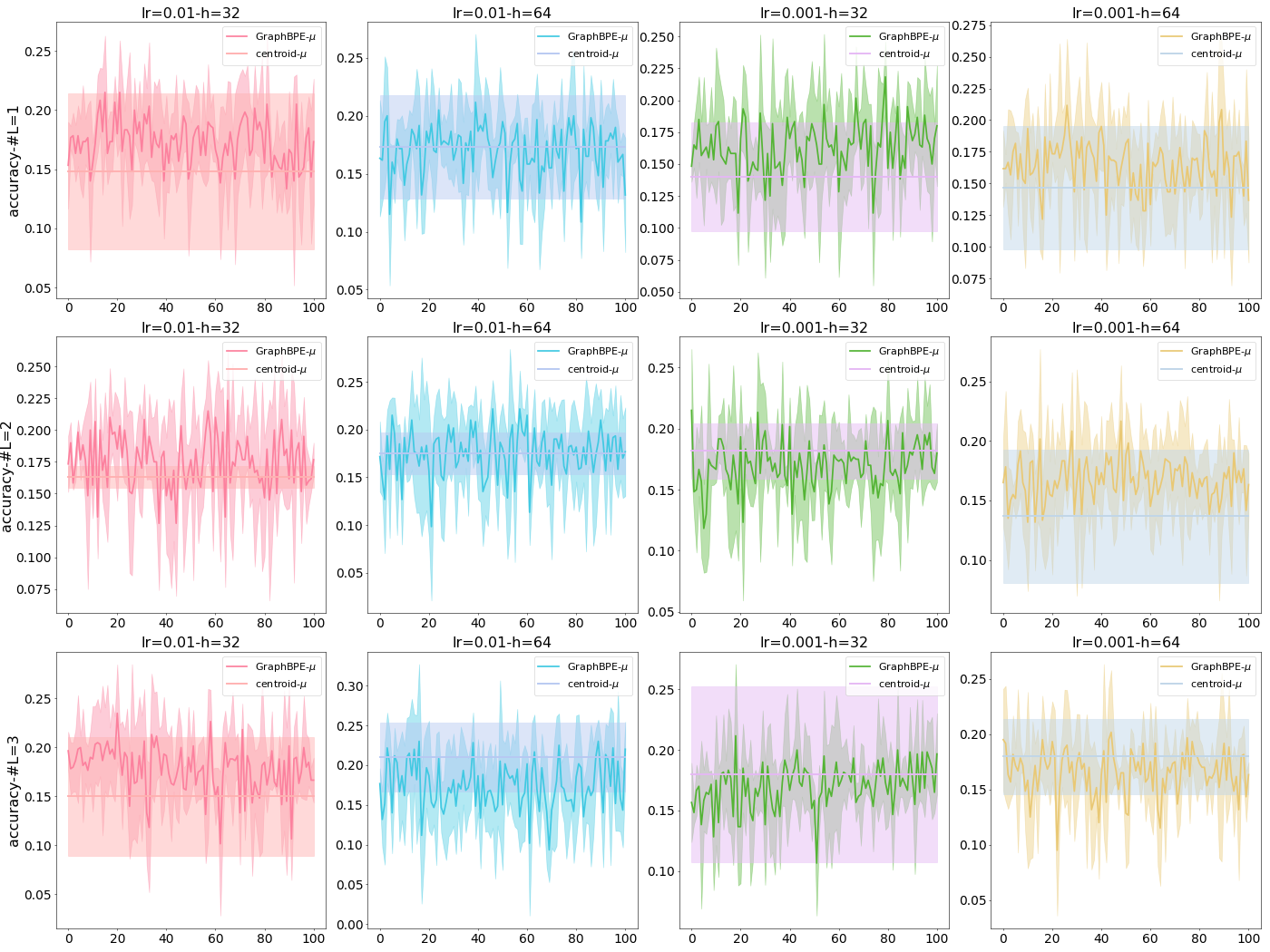}
\caption{Results of HGNN++ on \textsc{Enzymes}, with \textbf{accuracy} the \textit{higher} the better} 
\label{fig:enzymes_hgnnp}
\end{figure}

\begin{figure}[H]
\centering
\includegraphics[width=0.9\linewidth, height=0.59\textwidth]{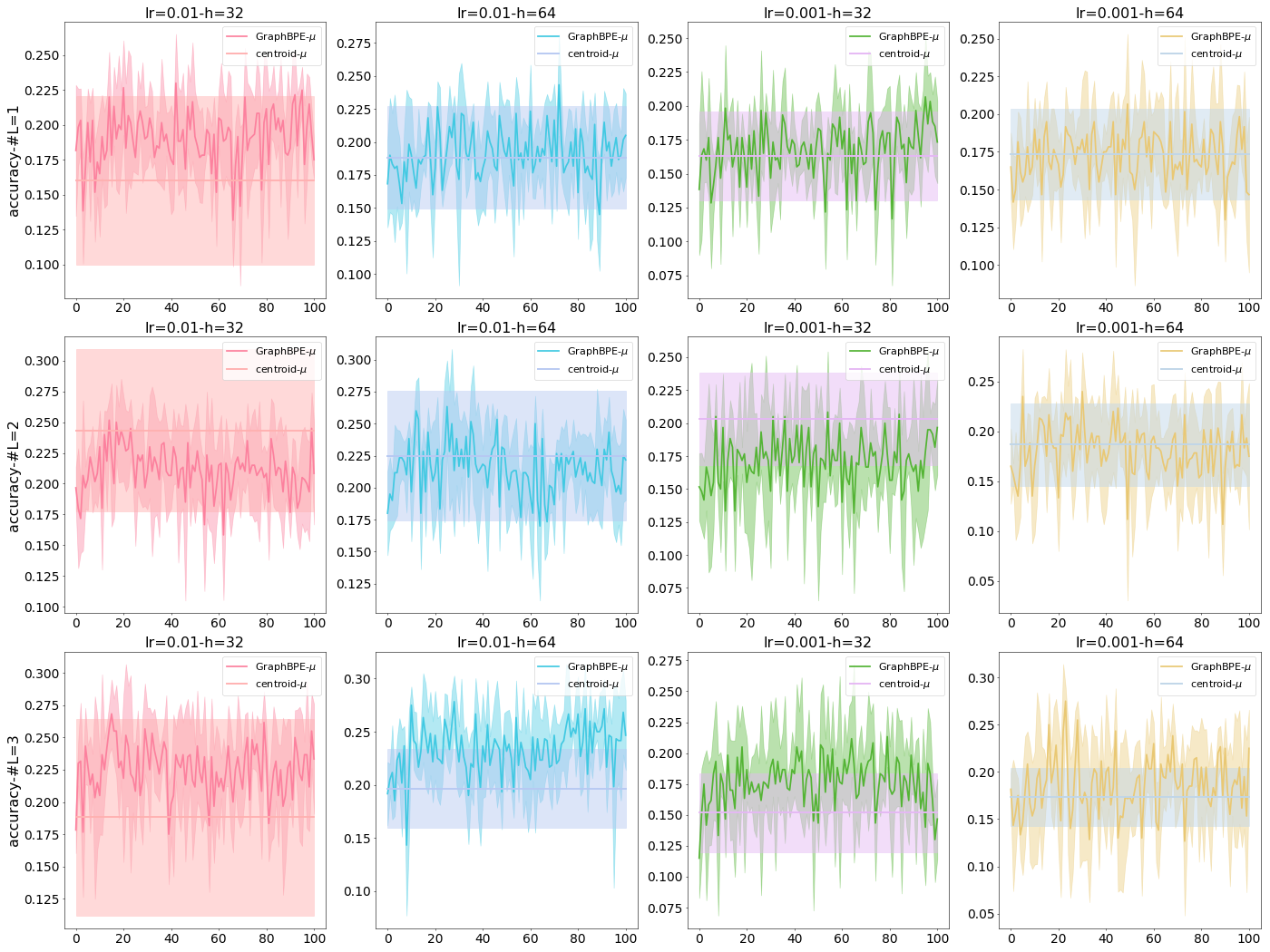}
\caption{Results of HNHN on \textsc{Enzymes}, with \textbf{accuracy} the \textit{higher} the better} 
\label{fig:enzymes_HNHN}
\end{figure}

For HyperGNNs, we include the performance comparison results in Table~\ref{tab: p_metric_3hgnns_on_enzymes}, and the visualization over different tokenization steps in Figure~\ref{fig:enzymes_hyperconv}, ~\ref{fig:enzymes_hgnnp}, and ~\ref{fig:enzymes_HNHN} for HyperConv, HGNN++, and HNHN.
\begin{table}[h]
    \centering
    \begin{minipage}[b]{0.45\linewidth}
        \centering
        \begin{subtable} % {.5\textwidth}
            \centering
	\resizebox{.99\textwidth}{!}{\begin{tabular}{lccccc}
	\toprule
	\multicolumn{2}{c}{\textbf{learning rate}} & \multicolumn{2}{c}{$10^{-2}$} & \multicolumn{2}{c}{$10^{-3}$} \\
	\multicolumn{2}{c}{\textbf{hidden size}} & $h=32$ & $h=64$ & $h=32$ & $h=64$ \\
	\midrule
	\multirow{2}{*}{$L={1}$} & p-value &2:\textcolor{red}{\textbf{99}}:0 & 2:\textcolor{red}{\textbf{99}}:0 & 1:\textcolor{red}{\textbf{100}}:0 & 41:\textcolor{red}{\textbf{60}}:0 \\
	 & metric &\textbf{66}:0:35 & \textbf{100}:0:1 & \textbf{77}:0:24 & \textbf{101}:0:0 \\
	\midrule
	\multirow{2}{*}{$L={2}$} & p-value &0:\textcolor{red}{\textbf{101}}:0 & 0:\textcolor{red}{\textbf{101}}:0 & 13:\textcolor{red}{\textbf{88}}:0 & 0:\textcolor{red}{\textbf{99}}:2 \\
	 & metric &\textbf{91}:2:8 & 39:1:\textbf{61} & \textbf{101}:0:0 & 49:0:\textbf{52} \\
	\midrule
	\multirow{2}{*}{$L={3}$} & p-value &0:\textcolor{red}{\textbf{101}}:0 & 3:\textcolor{red}{\textbf{98}}:0 & 0:\textcolor{red}{\textbf{101}}:0 & 2:\textcolor{red}{\textbf{96}}:3 \\
	 & metric &\textbf{66}:1:34 & \textbf{96}:0:5 & \textbf{63}:0:38 & 29:3:\textbf{69} \\
	\bottomrule
	\end{tabular}}
	\caption{Comparison with p-/metric value of GCN}
	\label{tab:p_metric_val_GCN_Enzymes}
        \end{subtable}
        
        \vspace{0.3cm} % adjust vertical space between tables
        
        \begin{subtable} %{.5\textwidth}
            \centering
	\resizebox{.99\textwidth}{!}{\begin{tabular}{lccccc}
	\toprule
	\multicolumn{2}{c}{\textbf{learning rate}} & \multicolumn{2}{c}{$10^{-2}$} & \multicolumn{2}{c}{$10^{-3}$} \\
	\multicolumn{2}{c}{\textbf{hidden size}} & $h=32$ & $h=64$ & $h=32$ & $h=64$ \\
	\midrule
	\multirow{2}{*}{$L={1}$} & p-value &0:6:\textcolor{red}{\textbf{95}} & 0:19:\textcolor{red}{\textbf{82}} & 1:\textcolor{red}{\textbf{100}}:0 & 0:\textcolor{red}{\textbf{97}}:4 \\
	 & metric &0:0:\textbf{101} & 0:0:\textbf{101} & \textbf{94}:0:7 & 15:0:\textbf{86} \\
	\midrule
	\multirow{2}{*}{$L={2}$} & p-value &0:\textcolor{red}{\textbf{101}}:0 & 1:\textcolor{red}{\textbf{90}}:10 & 0:\textcolor{red}{\textbf{100}}:1 & 0:\textcolor{red}{\textbf{90}}:11 \\
	 & metric &18:2:\textbf{81} & 35:0:\textbf{66} & 12:0:\textbf{89} & 4:1:\textbf{96} \\
	\midrule
	\multirow{2}{*}{$L={3}$} & p-value &0:\textcolor{red}{\textbf{54}}:47 & 0:\textcolor{red}{\textbf{86}}:15 & 0:\textcolor{red}{\textbf{101}}:0 & 7:\textcolor{red}{\textbf{94}}:0 \\
	 & metric &0:0:\textbf{101} & 2:0:\textbf{99} & 37:0:\textbf{64} & \textbf{82}:0:19 \\
	\bottomrule
	\end{tabular}}
	\caption{Comparison with p-/metric value of GIN}
	\label{tab:p_metric_val_GIN_Enzymes}
        \end{subtable}
    \end{minipage}
    \hspace{0.5cm}
    \begin{minipage}[b]{0.45\linewidth}
        \centering
        \begin{subtable} %{.5\textwidth}
            \centering
	\resizebox{.99\textwidth}{!}{\begin{tabular}{lccccc}
	\toprule
	\multicolumn{2}{c}{\textbf{learning rate}} & \multicolumn{2}{c}{$10^{-2}$} & \multicolumn{2}{c}{$10^{-3}$} \\
	\multicolumn{2}{c}{\textbf{hidden size}} & $h=32$ & $h=64$ & $h=32$ & $h=64$ \\
	\midrule
	\multirow{2}{*}{$L={1}$} & p-value &0:\textcolor{red}{\textbf{95}}:6 & 0:\textcolor{red}{\textbf{95}}:6 & 9:\textcolor{red}{\textbf{91}}:1 & 3:\textcolor{red}{\textbf{98}}:0 \\
	 & metric &6:0:\textbf{95} & 6:0:\textbf{95} & \textbf{81}:0:20 & \textbf{91}:1:9 \\
	\midrule
	\multirow{2}{*}{$L={2}$} & p-value &11:\textcolor{red}{\textbf{90}}:0 & 1:\textcolor{red}{\textbf{100}}:0 & 2:\textcolor{red}{\textbf{99}}:0 & 2:\textcolor{red}{\textbf{98}}:1 \\
	 & metric &\textbf{91}:0:10 & \textbf{97}:0:4 & \textbf{88}:1:12 & 47:2:\textbf{52} \\
	\midrule
	\multirow{2}{*}{$L={3}$} & p-value &0:\textcolor{red}{\textbf{100}}:1 & 0:\textcolor{red}{\textbf{75}}:26 & 1:\textcolor{red}{\textbf{100}}:0 & 0:\textcolor{red}{\textbf{91}}:10 \\
	 & metric &38:0:\textbf{63} & 2:0:\textbf{99} & \textbf{77}:1:23 & 0:0:\textbf{101} \\
	\bottomrule
	\end{tabular}}
	\caption{Comparison with p-/metric value of GAT}
	\label{tab:p_metric_val_GAT_Enzymes}
        \end{subtable}
        
        \vspace{0.3cm} % adjust vertical space between tables
        
        \begin{subtable} %{.5\textwidth}
            \centering
	\resizebox{.99\textwidth}{!}{\begin{tabular}{lccccc}
	\toprule
	\multicolumn{2}{c}{\textbf{learning rate}} & \multicolumn{2}{c}{$10^{-2}$} & \multicolumn{2}{c}{$10^{-3}$} \\
	\multicolumn{2}{c}{\textbf{hidden size}} & $h=32$ & $h=64$ & $h=32$ & $h=64$ \\
	\midrule
	\multirow{2}{*}{$L={1}$} & p-value &0:\textcolor{red}{\textbf{101}}:0 & 0:\textcolor{red}{\textbf{100}}:1 & 3:\textcolor{red}{\textbf{98}}:0 & 0:\textcolor{red}{\textbf{99}}:2 \\
	 & metric &20:0:\textbf{81} & 5:0:\textbf{96} & \textbf{88}:2:11 & 1:0:\textbf{100} \\
	\midrule
	\multirow{2}{*}{$L={2}$} & p-value &1:\textcolor{red}{\textbf{100}}:0 & 0:\textcolor{red}{\textbf{101}}:0 & 27:\textcolor{red}{\textbf{74}}:0 & 4:\textcolor{red}{\textbf{97}}:0 \\
	 & metric &\textbf{74}:0:27 & \textbf{65}:3:33 & \textbf{99}:0:2 & \textbf{95}:0:6 \\
	\midrule
	\multirow{2}{*}{$L={3}$} & p-value &0:\textcolor{red}{\textbf{101}}:0 & 0:\textcolor{red}{\textbf{93}}:8 & 1:\textcolor{red}{\textbf{100}}:0 & 0:\textcolor{red}{\textbf{101}}:0 \\
	 & metric &37:1:\textbf{63} & 3:0:\textbf{98} & \textbf{100}:0:1 & 1:0:\textbf{100} \\
	\bottomrule
	\end{tabular}}
	\caption{Comparison with p-/metric value of GraphSAGE}
	\label{tab:p_metric_val_GraphSAGE_Enzymes}
        \end{subtable}
    \end{minipage}
    \caption{Performance comparison on accuracy with p-value $<0.05$ and metric value on \textsc{Enzymes}. For each triplet $a$:$b$:$c$, $a, b, c$ denote the number of times \textsc{GraphBPE} is \textcolor{red}{\textbf{statistically}}/\textbf{numerically} better/the same/worse compared with (untokenized) simple graph.}
    \label{tab: p_metric_4gnns_on_enzymes}
\end{table}
\newpage
\FloatBarrier  % Prevent floats from moving past this point
%%%%%%%%%%%%%%%%%%%%%%%%%%%%%%%%%%%%%%%%%
\FloatBarrier  % Prevent floats from moving past this point
%%%%%%%%%%%%%%%%%%%%%%%%%%%%%%%%%%%%%%%%%
\FloatBarrier  % Prevent floats from moving past this point
%%%%%%%%%%%%%%%%%%%%%%%%%%%%%%%%%%%%%%%%%
\FloatBarrier  % Prevent floats from moving past this point
\newpage
\begin{table}[h]
    \centering
    \begin{minipage}[b]{0.45\linewidth}
        \centering
        \begin{subtable}
            \centering
	\resizebox{.99\textwidth}{!}{\begin{tabular}{lccccc}
	\toprule
	\multicolumn{2}{c}{\textbf{learning rate}} & \multicolumn{2}{c}{$10^{-2}$} & \multicolumn{2}{c}{$10^{-3}$} \\
	\multicolumn{2}{c}{\textbf{hidden size}} & $h=32$ & $h=64$ & $h=32$ & $h=64$ \\
	\midrule
	\multirow{2}{*}{$L={1}$} & p-value &1:\textcolor{red}{\textbf{96}}:4 & 36:\textcolor{red}{\textbf{65}}:0 & 0:\textcolor{red}{\textbf{101}}:0 & 0:\textcolor{red}{\textbf{100}}:1 \\
	 & metric &\textbf{52}:0:49 & \textbf{101}:0:0 & 45:0:\textbf{56} & 20:0:\textbf{81} \\
	\midrule
	\multirow{2}{*}{$L={2}$} & p-value &0:\textcolor{red}{\textbf{93}}:8 & 0:\textcolor{red}{\textbf{91}}:10 & 2:\textcolor{red}{\textbf{97}}:2 & 5:\textcolor{red}{\textbf{96}}:0 \\
	 & metric &12:0:\textbf{89} & 0:0:\textbf{101} & 42:0:\textbf{59} & \textbf{79}:0:22 \\
	\midrule
	\multirow{2}{*}{$L={3}$} & p-value &2:\textcolor{red}{\textbf{99}}:0 & 5:\textcolor{red}{\textbf{95}}:1 & 0:\textcolor{red}{\textbf{98}}:3 & 0:\textcolor{red}{\textbf{101}}:0 \\
	 & metric &\textbf{82}:0:19 & \textbf{72}:0:29 & 16:0:\textbf{85} & \textbf{81}:0:20 \\
	\bottomrule
	\end{tabular}}
	\caption{\textsc{Centroid} on HyperConv}
	\label{tab:p_metric_val_HyperConv_Enzymes_Centroid}
        \end{subtable}
    \end{minipage}
    \hspace{0.5cm}
    \vspace{0.3cm} % adjust vertical space between tables
    \begin{minipage}[b]{0.45\linewidth}
        \centering
        \begin{subtable}
            \centering
	\resizebox{.99\textwidth}{!}{\begin{tabular}{lccccc}
	\toprule
	\multicolumn{2}{c}{\textbf{learning rate}} & \multicolumn{2}{c}{$10^{-2}$} & \multicolumn{2}{c}{$10^{-3}$} \\
	\multicolumn{2}{c}{\textbf{hidden size}} & $h=32$ & $h=64$ & $h=32$ & $h=64$ \\
	\midrule
	\multirow{2}{*}{$L={1}$} & p-value &0:\textcolor{red}{\textbf{101}}:0 & 0:\textcolor{red}{\textbf{100}}:1 & 2:\textcolor{red}{\textbf{99}}:0 & 1:\textcolor{red}{\textbf{100}}:0 \\
	 & metric &\textbf{90}:0:11 & 49:0:\textbf{52} & \textbf{92}:0:9 & \textbf{81}:0:20 \\
	\midrule
	\multirow{2}{*}{$L={2}$} & p-value &8:\textcolor{red}{\textbf{93}}:0 & 0:\textcolor{red}{\textbf{101}}:0 & 0:\textcolor{red}{\textbf{96}}:5 & 2:\textcolor{red}{\textbf{99}}:0 \\
	 & metric &\textbf{69}:0:32 & \textbf{57}:0:44 & 22:2:\textbf{77} & \textbf{97}:0:4 \\
	\midrule
	\multirow{2}{*}{$L={3}$} & p-value &4:\textcolor{red}{\textbf{97}}:0 & 0:\textcolor{red}{\textbf{94}}:7 & 0:\textcolor{red}{\textbf{101}}:0 & 0:\textcolor{red}{\textbf{100}}:1 \\
	 & metric &\textbf{89}:0:12 & 13:0:\textbf{88} & 30:0:\textbf{71} & 21:0:\textbf{80} \\
	\bottomrule
	\end{tabular}}
	\caption{\textsc{Centroid} on HGNN++}
	\label{tab:p_metric_val_HGNN++_Enzymes_Centroid}
        \end{subtable}
    \end{minipage}
    \hspace{0.5cm}
    \vspace{0.3cm} % adjust vertical space between tables
    \centering
    \resizebox{.49\textwidth}{!}{\begin{tabular}{lccccc}
	\toprule
	\multicolumn{2}{c}{\textbf{learning rate}} & \multicolumn{2}{c}{$10^{-2}$} & \multicolumn{2}{c}{$10^{-3}$} \\
	\multicolumn{2}{c}{\textbf{hidden size}} & $h=32$ & $h=64$ & $h=32$ & $h=64$ \\
	\midrule
	\multirow{2}{*}{$L={1}$} & p-value &3:\textcolor{red}{\textbf{98}}:0 & 0:\textcolor{red}{\textbf{101}}:0 & 1:\textcolor{red}{\textbf{99}}:1 & 1:\textcolor{red}{\textbf{100}}:0 \\
	 & metric &\textbf{95}:0:6 & \textbf{50}:1:\textbf{50} & \textbf{65}:0:36 & 48:0:\textbf{53} \\
	\midrule
	\multirow{2}{*}{$L={2}$} & p-value &0:\textcolor{red}{\textbf{101}}:0 & 0:\textcolor{red}{\textbf{100}}:1 & 0:\textcolor{red}{\textbf{91}}:10 & 0:\textcolor{red}{\textbf{100}}:1 \\
	 & metric &4:0:\textbf{97} & 29:0:\textbf{72} & 5:0:\textbf{96} & 45:0:\textbf{56} \\
	\midrule
	\multirow{2}{*}{$L={3}$} & p-value &0:\textcolor{red}{\textbf{101}}:0 & 22:\textcolor{red}{\textbf{79}}:0 & 8:\textcolor{red}{\textbf{93}}:0 & 6:\textcolor{red}{\textbf{94}}:1 \\
	 & metric &\textbf{96}:0:5 & \textbf{95}:0:6 & \textbf{90}:0:11 & \textbf{68}:1:32 \\
	\bottomrule
	\end{tabular}}
    \caption{\textsc{Centroid} on HNHN}
    \label{tab:p_metric_val_HNHN_Enzymes_Centroid}
    \caption{Performance comparison on accuracy with p-value $<0.05$ and metric value on \textsc{Enzymes}. For each triplet $a$:$b$:$c$, $a, b, c$ denote the number of times \textsc{GraphBPE} is \textcolor{red}{\textbf{statistically}}/\textbf{numerically} better/the same/worse compared with hypergraphs constructed by \textsc{Method} on Model (e.g., ``\textsc{Centroud} on HyperConv'' means comparing \textsc{GraphBPE} with \textsc{Centroid} on the HyperConv model).}
    \label{tab: p_metric_3hgnns_on_enzymes}
\end{table}
\FloatBarrier  % Prevent floats from moving past this point
\FloatBarrier  % Prevent floats from moving past this point
\FloatBarrier  % Prevent floats from moving past this point

%%%%%%%%%%%%%%%%%%%%%%%%%%%%%%%%%%%%%%%%%
\newpage

\subsection{\textsc{Proteins}}\label{app: proteins_result}
\begin{figure}[H]
\centering
\includegraphics[width=0.9\linewidth, height=0.59\textwidth]{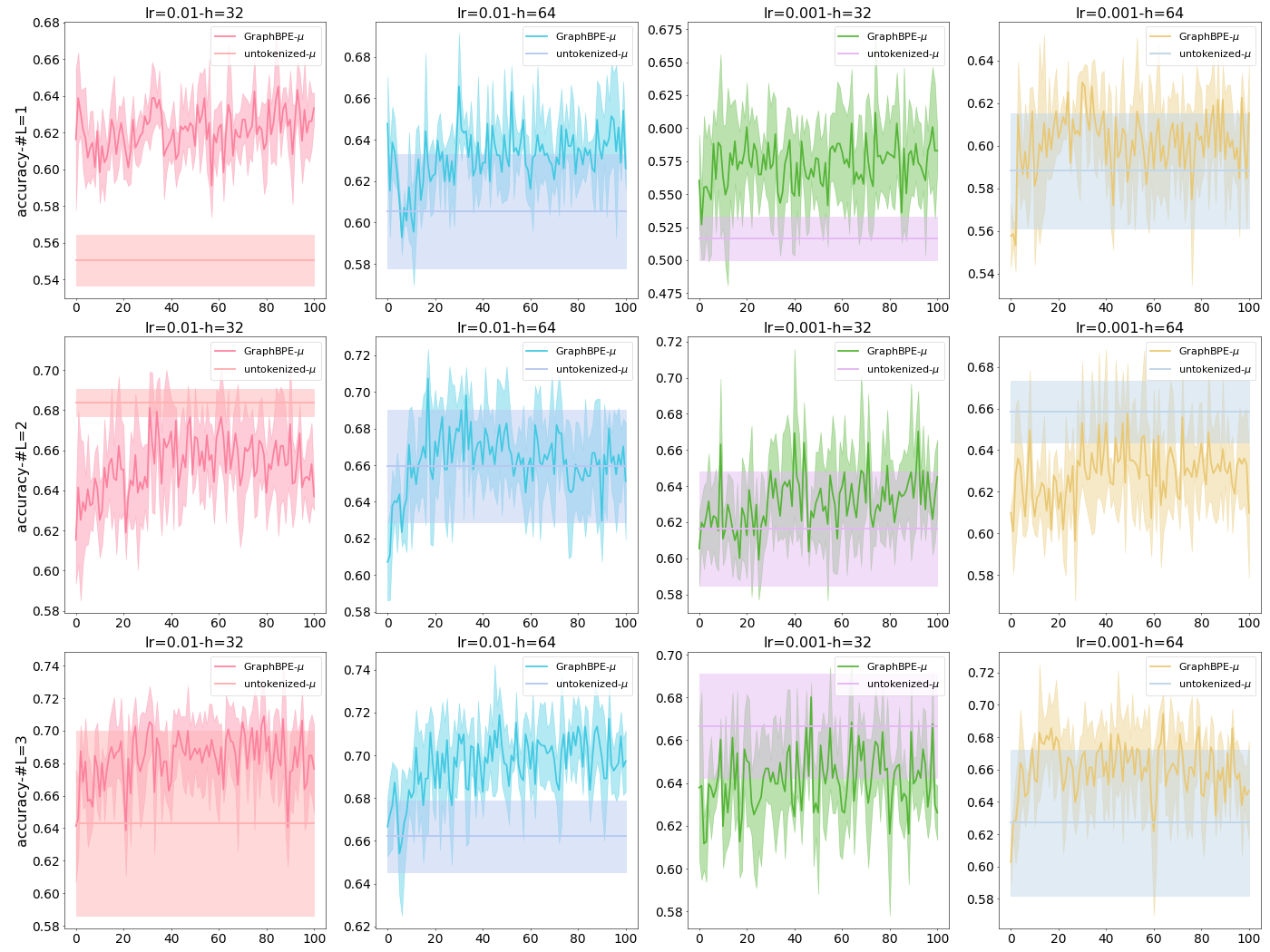}
\caption{Results of GCN on \textsc{Proteins}, with \textbf{accuracy} the \textit{higher} the better} 
\label{fig:proteins_gcn}
\end{figure}

\begin{figure}[H]
\centering
\includegraphics[width=0.9\linewidth, height=0.59\textwidth]{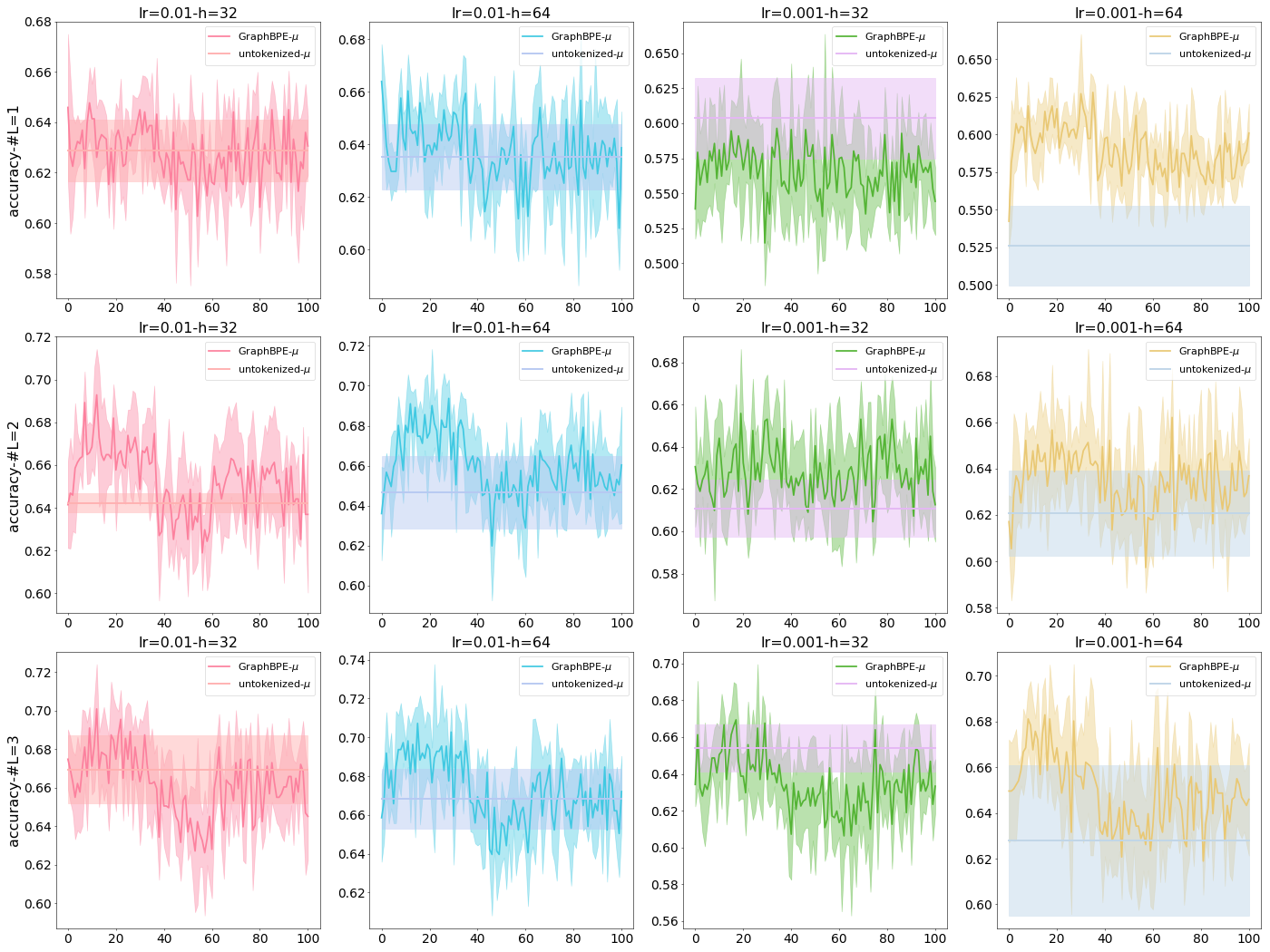}
\caption{Results of GAT on \textsc{Proteins}, with \textbf{accuracy} the \textit{higher} the better} 
\label{fig:proteins_gat}
\end{figure}

\begin{figure}[H]
\centering
\includegraphics[width=0.9\linewidth, height=0.59\textwidth]{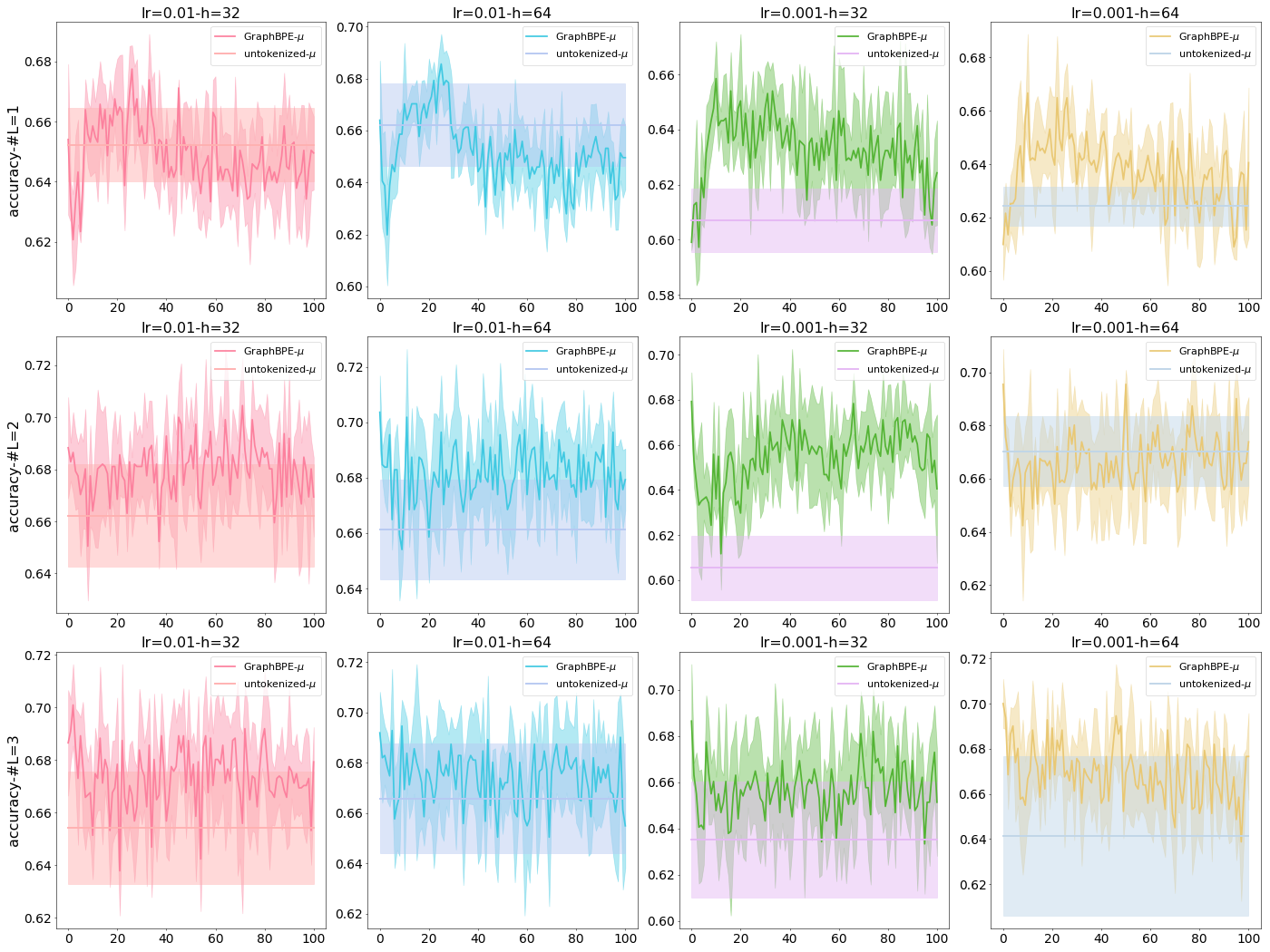}
\caption{Results of GIN on \textsc{Proteins}, with \textbf{accuracy} the \textit{higher} the better} 
\label{fig:proteins_gin}
\end{figure}

\begin{figure}[H]
\centering
\includegraphics[width=0.9\linewidth, height=0.59\textwidth]{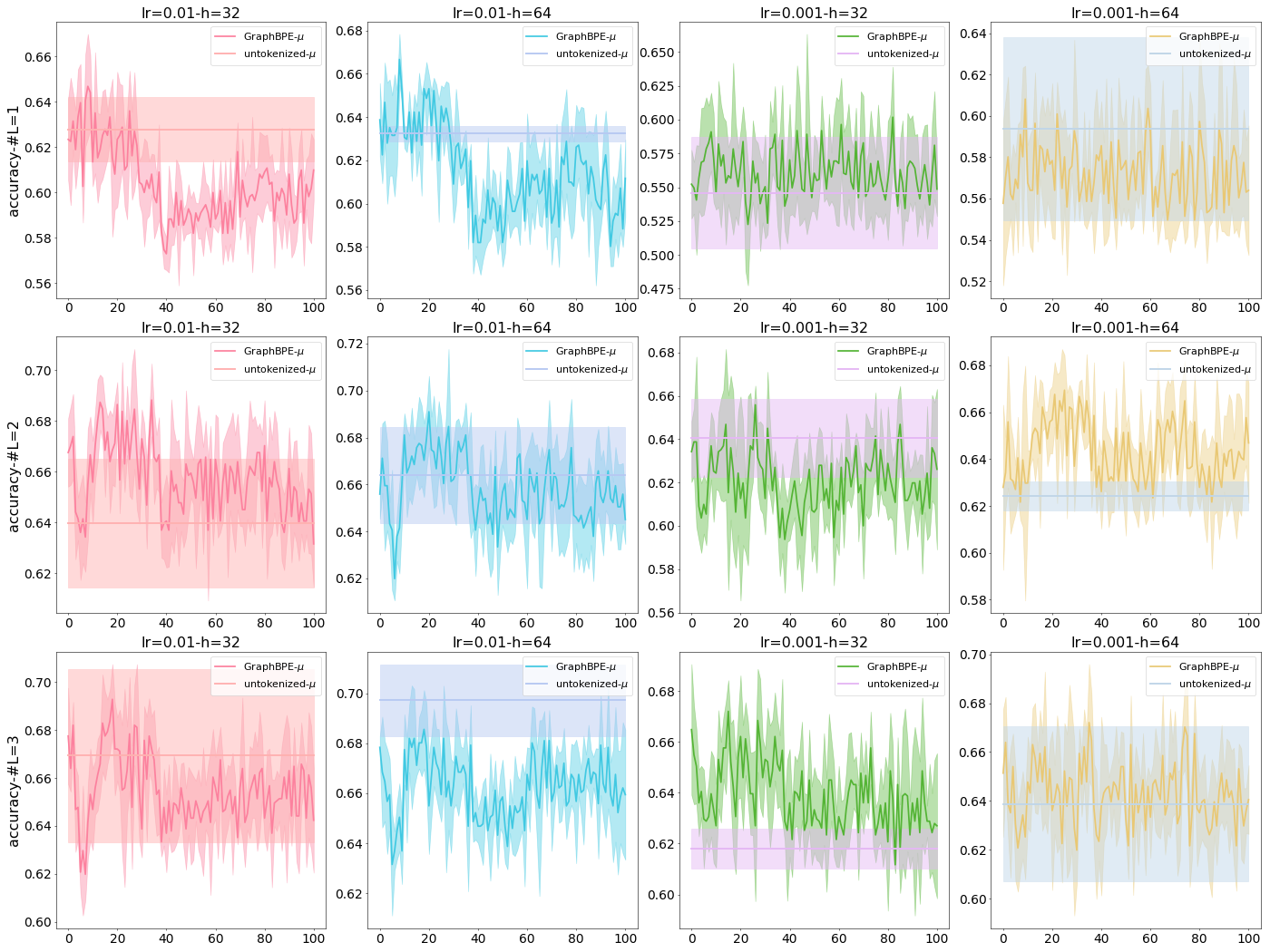}
\caption{Results of GraphSAGE on \textsc{Proteins}, with \textbf{accuracy} the \textit{higher} the better} 
\label{fig:proteins_graphsage}
\end{figure}

For GNNs, we include the performance comparison results in Table~\ref{tab: p_metric_4gnns_on_proteins}, and the visualization over different tokenization steps in Figure~\ref{fig:proteins_gcn}, ~\ref{fig:proteins_gat}, ~\ref{fig:proteins_gin}, and ~\ref{fig:proteins_graphsage} for GCN, GAT, GIN, and GraphSAGE.

\begin{figure}[H]
\centering
\includegraphics[width=0.9\linewidth, height=0.59\textwidth]{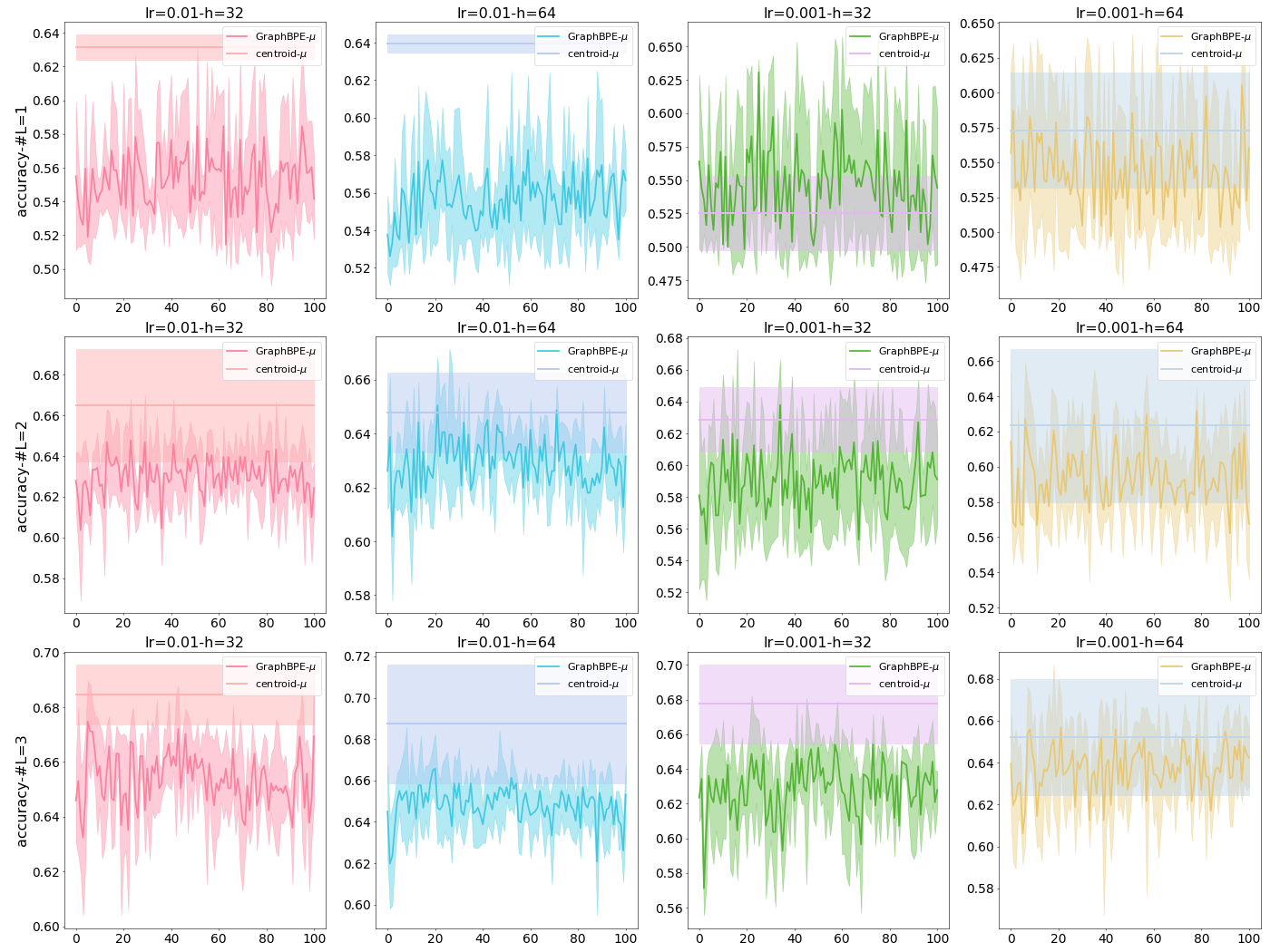}
\caption{Results of HyperConv on \textsc{Proteins}, with \textbf{accuracy} the \textit{higher} the better} 
\label{fig:proteins_hyperconv}
\end{figure}

\begin{figure}[H]
\centering
\includegraphics[width=0.9\linewidth, height=0.59\textwidth]{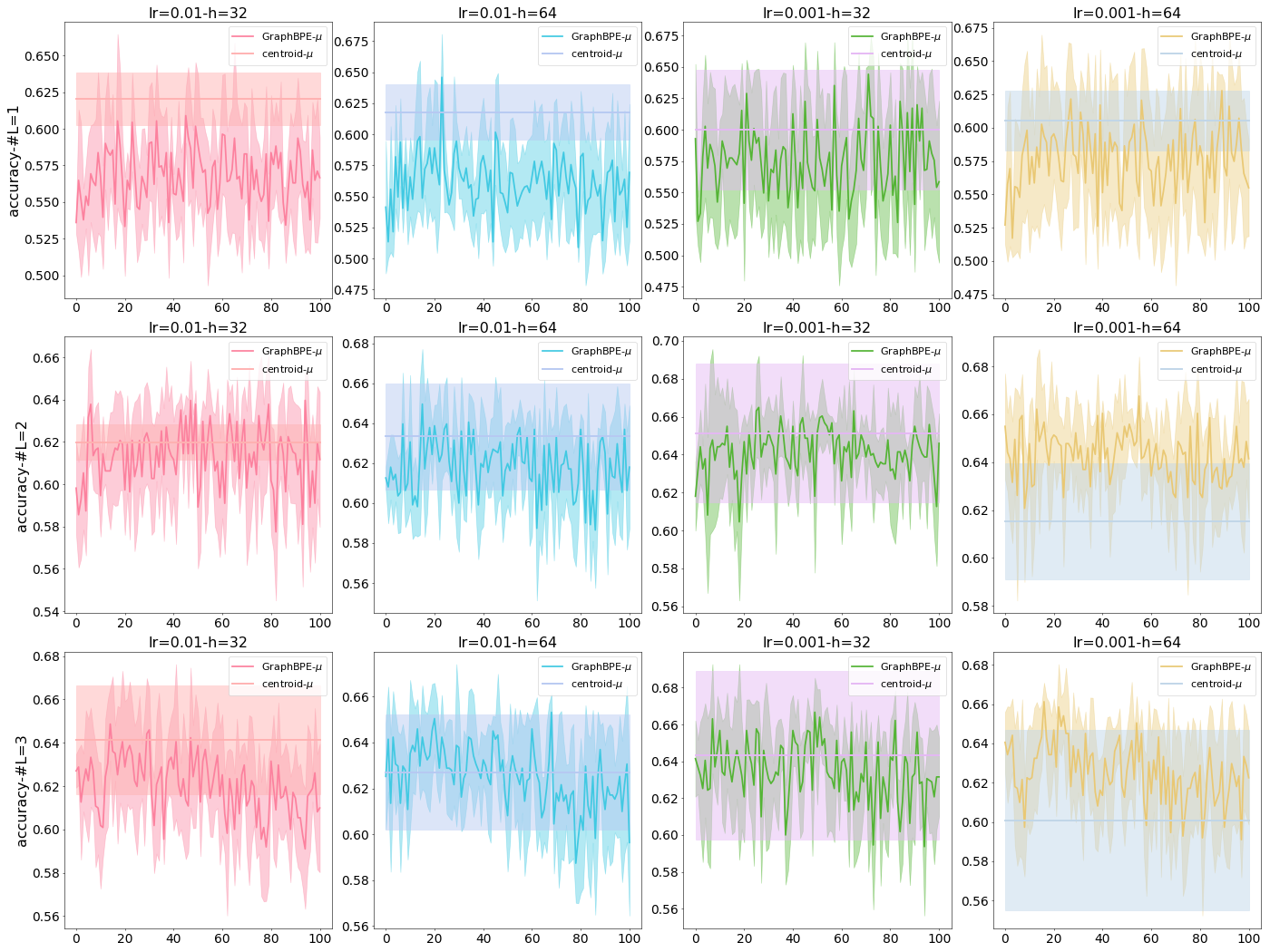}
\caption{Results of HGNN++ on \textsc{Proteins}, with \textbf{accuracy} the \textit{higher} the better} 
\label{fig:proteins_HGNNP}
\end{figure}

\begin{figure}[H]
\centering
\includegraphics[width=0.9\linewidth, height=0.59\textwidth]{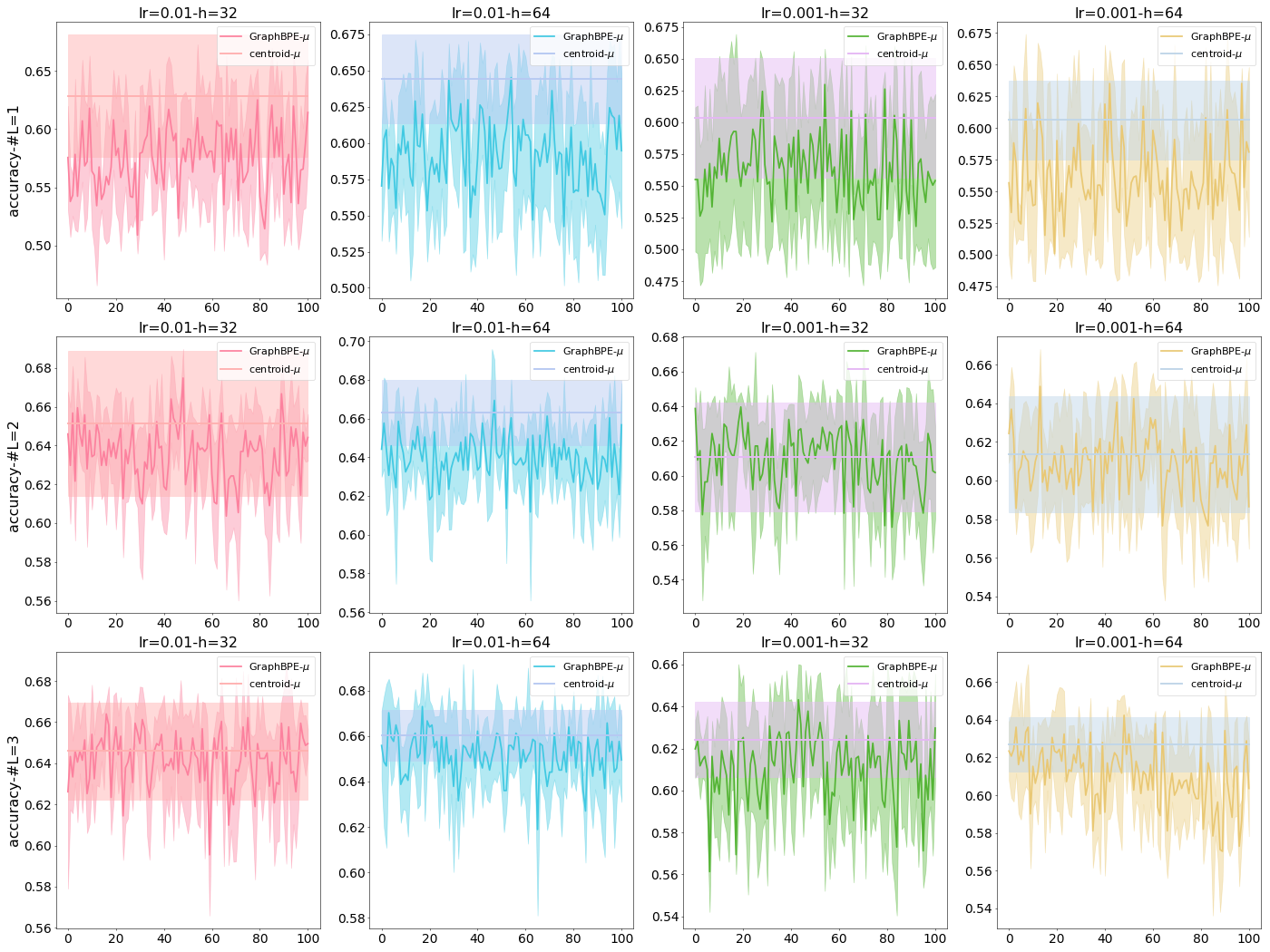}
\caption{Results of HNHN on \textsc{Proteins}, with \textbf{accuracy} the \textit{higher} the better} 
\label{fig:proteins_HNHN}
\end{figure}

For HyperGNNs, we include the performance comparison results in Table~\ref{tab: p_metric_3hgnns_on_proteins}, and the visualization over different tokenization steps in Figure~\ref{fig:proteins_hyperconv}, ~\ref{fig:proteins_HGNNP}, and ~\ref{fig:proteins_HNHN} for HyperConv, HGNN++, and HNHN.
\begin{table}[h]
    \centering
    \begin{minipage}[b]{0.45\linewidth}
        \centering
        \begin{subtable} % {.5\textwidth}
            \centering
	\resizebox{.99\textwidth}{!}{\begin{tabular}{lccccc}
	\toprule
	\multicolumn{2}{c}{\textbf{learning rate}} & \multicolumn{2}{c}{$10^{-2}$} & \multicolumn{2}{c}{$10^{-3}$} \\
	\multicolumn{2}{c}{\textbf{hidden size}} & $h=32$ & $h=64$ & $h=32$ & $h=64$ \\
	\midrule
	\multirow{2}{*}{$L={1}$} & p-value &\textcolor{red}{\textbf{101}}:0:0 & 9:\textcolor{red}{\textbf{92}}:0 & \textcolor{red}{\textbf{57}}:44:0 & 3:\textcolor{red}{\textbf{98}}:0 \\
	 & metric &\textbf{101}:0:0 & \textbf{97}:0:4 & \textbf{101}:0:0 & \textbf{88}:0:13 \\
	\midrule
	\multirow{2}{*}{$L={2}$} & p-value &0:48:\textcolor{red}{\textbf{53}} & 3:\textcolor{red}{\textbf{97}}:1 & 2:\textcolor{red}{\textbf{99}}:0 & 0:\textcolor{red}{\textbf{61}}:40 \\
	 & metric &0:0:\textbf{101} & \textbf{64}:0:37 & \textbf{89}:1:11 & 0:0:\textbf{101} \\
	\midrule
	\multirow{2}{*}{$L={3}$} & p-value &1:\textcolor{red}{\textbf{100}}:0 & \textcolor{red}{\textbf{53}}:48:0 & 0:\textcolor{red}{\textbf{83}}:18 & 5:\textcolor{red}{\textbf{96}}:0 \\
	 & metric &\textbf{98}:0:3 & \textbf{99}:0:2 & 3:1:\textbf{97} & \textbf{99}:0:2 \\
	\bottomrule
	\end{tabular}}
	\caption{Comparison with p-/metric value of GCN}
	\label{tab:p_metric_val_GCN_Proteins}
        \end{subtable}
        
        \vspace{0.3cm} % adjust vertical space between tables
        
        \begin{subtable} %{.5\textwidth}
            \centering
	\resizebox{.99\textwidth}{!}{\begin{tabular}{lccccc}
	\toprule
	\multicolumn{2}{c}{\textbf{learning rate}} & \multicolumn{2}{c}{$10^{-2}$} & \multicolumn{2}{c}{$10^{-3}$} \\
	\multicolumn{2}{c}{\textbf{hidden size}} & $h=32$ & $h=64$ & $h=32$ & $h=64$ \\
	\midrule
	\multirow{2}{*}{$L={1}$} & p-value &3:\textcolor{red}{\textbf{93}}:5 & 0:\textcolor{red}{\textbf{90}}:11 & \textcolor{red}{\textbf{53}}:48:0 & 29:\textcolor{red}{\textbf{70}}:2 \\
	 & metric &43:0:\textbf{58} & 20:0:\textbf{81} & \textbf{98}:0:3 & \textbf{87}:1:13 \\
	\midrule
	\multirow{2}{*}{$L={2}$} & p-value &6:\textcolor{red}{\textbf{95}}:0 & 15:\textcolor{red}{\textbf{86}}:0 & \textcolor{red}{\textbf{88}}:13:0 & 2:\textcolor{red}{\textbf{97}}:2 \\
	 & metric &\textbf{98}:0:3 & \textbf{98}:0:3 & \textbf{101}:0:0 & 28:4:\textbf{69} \\
	\midrule
	\multirow{2}{*}{$L={3}$} & p-value &16:\textcolor{red}{\textbf{84}}:1 & 3:\textcolor{red}{\textbf{98}}:0 & 11:\textcolor{red}{\textbf{90}}:0 & 8:\textcolor{red}{\textbf{93}}:0 \\
	 & metric &\textbf{95}:0:6 & \textbf{78}:1:22 & \textbf{98}:1:2 & \textbf{100}:0:1 \\
	\bottomrule
	\end{tabular}}
	\caption{Comparison with p-/metric value of GIN}
	\label{tab:p_metric_val_GIN_Proteins}
        \end{subtable}
    \end{minipage}
    \hspace{0.5cm}
    \begin{minipage}[b]{0.45\linewidth}
        \centering
        \begin{subtable} %{.5\textwidth}
            \centering
	\resizebox{.99\textwidth}{!}{\begin{tabular}{lccccc}
	\toprule
	\multicolumn{2}{c}{\textbf{learning rate}} & \multicolumn{2}{c}{$10^{-2}$} & \multicolumn{2}{c}{$10^{-3}$} \\
	\multicolumn{2}{c}{\textbf{hidden size}} & $h=32$ & $h=64$ & $h=32$ & $h=64$ \\
	\midrule
	\multirow{2}{*}{$L={1}$} & p-value &3:\textcolor{red}{\textbf{98}}:0 & 2:\textcolor{red}{\textbf{96}}:3 & 0:\textcolor{red}{\textbf{81}}:20 & \textcolor{red}{\textbf{76}}:25:0 \\
	 & metric &47:1:\textbf{53} & \textbf{54}:2:45 & 0:0:\textbf{101} & \textbf{101}:0:0 \\
	\midrule
	\multirow{2}{*}{$L={2}$} & p-value &19:\textcolor{red}{\textbf{78}}:4 & 13:\textcolor{red}{\textbf{88}}:0 & 18:\textcolor{red}{\textbf{83}}:0 & 9:\textcolor{red}{\textbf{92}}:0 \\
	 & metric &\textbf{77}:0:24 & \textbf{84}:0:17 & \textbf{96}:0:5 & \textbf{88}:0:13 \\
	\midrule
	\multirow{2}{*}{$L={3}$} & p-value &1:\textcolor{red}{\textbf{94}}:6 & 7:\textcolor{red}{\textbf{92}}:2 & 0:\textcolor{red}{\textbf{81}}:20 & 6:\textcolor{red}{\textbf{95}}:0 \\
	 & metric &32:1:\textbf{68} & \textbf{54}:1:46 & 9:1:\textbf{91} & \textbf{93}:2:6 \\
	\bottomrule
	\end{tabular}}
	\caption{Comparison with p-/metric value of GAT}
	\label{tab:p_metric_val_GAT_Proteins}
        \end{subtable}
        
        \vspace{0.3cm} % adjust vertical space between tables
        
        \begin{subtable} %{.5\textwidth}
            \centering
	\resizebox{.99\textwidth}{!}{\begin{tabular}{lccccc}
	\toprule
	\multicolumn{2}{c}{\textbf{learning rate}} & \multicolumn{2}{c}{$10^{-2}$} & \multicolumn{2}{c}{$10^{-3}$} \\
	\multicolumn{2}{c}{\textbf{hidden size}} & $h=32$ & $h=64$ & $h=32$ & $h=64$ \\
	\midrule
	\multirow{2}{*}{$L={1}$} & p-value &0:\textcolor{red}{\textbf{58}}:43 & 5:\textcolor{red}{\textbf{60}}:36 & 1:\textcolor{red}{\textbf{100}}:0 & 0:\textcolor{red}{\textbf{100}}:1 \\
	 & metric &10:1:\textbf{90} & 19:0:\textbf{82} & \textbf{84}:0:17 & 6:0:\textbf{95} \\
	\midrule
	\multirow{2}{*}{$L={2}$} & p-value &10:\textcolor{red}{\textbf{91}}:0 & 0:\textcolor{red}{\textbf{97}}:4 & 0:\textcolor{red}{\textbf{82}}:19 & 34:\textcolor{red}{\textbf{67}}:0 \\
	 & metric &\textbf{91}:0:10 & 35:1:\textbf{65} & 5:0:\textbf{96} & \textbf{97}:1:3 \\
	\midrule
	\multirow{2}{*}{$L={3}$} & p-value &0:\textcolor{red}{\textbf{101}}:0 & 0:28:\textcolor{red}{\textbf{73}} & 35:\textcolor{red}{\textbf{66}}:0 & 0:\textcolor{red}{\textbf{101}}:0 \\
	 & metric &16:0:\textbf{85} & 0:0:\textbf{101} & \textbf{99}:1:1 & \textbf{61}:1:39 \\
	\bottomrule
	\end{tabular}}
	\caption{Comparison with p-/metric value of GraphSAGE}
	\label{tab:p_metric_val_GraphSAGE_Proteins}
        \end{subtable}
    \end{minipage}
    \caption{Performance comparison on accuracy with p-value $<0.05$ and metric value on \textsc{Proteins}. For each triplet $a$:$b$:$c$, $a, b, c$ denote the number of times \textsc{GraphBPE} is \textcolor{red}{\textbf{statistically}}/\textbf{numerically} better/the same/worse compared with (untokenized) simple graph.}
    \label{tab: p_metric_4gnns_on_proteins}
\end{table}
\newpage
\FloatBarrier  % Prevent floats from moving past this point
%%%%%%%%%%%%%%%%%%%%%%%%%%%%%%%%%%%%%%%%%
\FloatBarrier  % Prevent floats from moving past this point
%%%%%%%%%%%%%%%%%%%%%%%%%%%%%%%%%%%%%%%%%
\FloatBarrier  % Prevent floats from moving past this point
%%%%%%%%%%%%%%%%%%%%%%%%%%%%%%%%%%%%%%%%%
\FloatBarrier  % Prevent floats from moving past this point
\newpage
\begin{table}[h]
    \centering
    \begin{minipage}[b]{0.45\linewidth}
        \centering
        \begin{subtable}
            \centering
	\resizebox{.99\textwidth}{!}{\begin{tabular}{lccccc}
	\toprule
	\multicolumn{2}{c}{\textbf{learning rate}} & \multicolumn{2}{c}{$10^{-2}$} & \multicolumn{2}{c}{$10^{-3}$} \\
	\multicolumn{2}{c}{\textbf{hidden size}} & $h=32$ & $h=64$ & $h=32$ & $h=64$ \\
	\midrule
	\multirow{2}{*}{$L={1}$} & p-value &0:10:\textcolor{red}{\textbf{91}} & 0:2:\textcolor{red}{\textbf{99}} & 2:\textcolor{red}{\textbf{97}}:2 & 1:\textcolor{red}{\textbf{90}}:10 \\
	 & metric &0:0:\textbf{101} & 0:0:\textbf{101} & \textbf{81}:1:19 & 9:0:\textbf{92} \\
	\midrule
	\multirow{2}{*}{$L={2}$} & p-value &0:\textcolor{red}{\textbf{71}}:30 & 0:\textcolor{red}{\textbf{74}}:27 & 0:\textcolor{red}{\textbf{76}}:25 & 0:\textcolor{red}{\textbf{100}}:1 \\
	 & metric &0:0:\textbf{101} & 2:0:\textbf{99} & 1:0:\textbf{100} & 4:0:\textbf{97} \\
	\midrule
	\multirow{2}{*}{$L={3}$} & p-value &0:28:\textcolor{red}{\textbf{73}} & 0:\textcolor{red}{\textbf{69}}:32 & 0:43:\textcolor{red}{\textbf{58}} & 0:\textcolor{red}{\textbf{97}}:4 \\
	 & metric &0:0:\textbf{101} & 0:0:\textbf{101} & 0:0:\textbf{101} & 7:0:\textbf{94} \\
	\bottomrule
	\end{tabular}}
	\caption{\textsc{Centroid} on HyperConv}
	\label{tab:p_metric_val_HyperConv_Proteins_Centroid}
        \end{subtable}
    \end{minipage}
    \vspace{0.3cm} % adjust vertical space between tables
    \hspace{0.5cm}
    \begin{minipage}[b]{0.45\linewidth}
        \centering
        \begin{subtable}
            \centering
	\resizebox{.99\textwidth}{!}{\begin{tabular}{lccccc}
	\toprule
	\multicolumn{2}{c}{\textbf{learning rate}} & \multicolumn{2}{c}{$10^{-2}$} & \multicolumn{2}{c}{$10^{-3}$} \\
	\multicolumn{2}{c}{\textbf{hidden size}} & $h=32$ & $h=64$ & $h=32$ & $h=64$ \\
	\midrule
	\multirow{2}{*}{$L={1}$} & p-value &0:\textcolor{red}{\textbf{53}}:48 & 0:\textcolor{red}{\textbf{57}}:44 & 0:\textcolor{red}{\textbf{97}}:4 & 0:\textcolor{red}{\textbf{92}}:9 \\
	 & metric &0:0:\textbf{101} & 1:0:\textbf{100} & 19:0:\textbf{82} & 9:2:\textbf{90} \\
	\midrule
	\multirow{2}{*}{$L={2}$} & p-value &2:\textcolor{red}{\textbf{94}}:5 & 0:\textcolor{red}{\textbf{97}}:4 & 0:\textcolor{red}{\textbf{99}}:2 & 16:\textcolor{red}{\textbf{85}}:0 \\
	 & metric &30:2:\textbf{69} & 14:1:\textbf{86} & 16:2:\textbf{83} & \textbf{101}:0:0 \\
	\midrule
	\multirow{2}{*}{$L={3}$} & p-value &0:\textcolor{red}{\textbf{96}}:5 & 0:\textcolor{red}{\textbf{98}}:3 & 0:\textcolor{red}{\textbf{99}}:2 & 0:\textcolor{red}{\textbf{101}}:0 \\
	 & metric &4:0:\textbf{97} & 48:1:\textbf{52} & 29:0:\textbf{72} & \textbf{94}:0:7 \\
	\bottomrule
	\end{tabular}}
	\caption{\textsc{Centroid} on HGNN++}
	\label{tab:p_metric_val_HGNN++_Proteins_Centroid}
        \end{subtable}
    \end{minipage}
    \vspace{0.3cm} % adjust vertical space between tables
    \centering
    \resizebox{.49\textwidth}{!}{\begin{tabular}{lccccc}
	\toprule
	\multicolumn{2}{c}{\textbf{learning rate}} & \multicolumn{2}{c}{$10^{-2}$} & \multicolumn{2}{c}{$10^{-3}$} \\
	\multicolumn{2}{c}{\textbf{hidden size}} & $h=32$ & $h=64$ & $h=32$ & $h=64$ \\
	\midrule
	\multirow{2}{*}{$L={1}$} & p-value &0:\textcolor{red}{\textbf{80}}:21 & 0:\textcolor{red}{\textbf{71}}:30 & 0:\textcolor{red}{\textbf{91}}:10 & 1:\textcolor{red}{\textbf{86}}:14 \\
	 & metric &0:0:\textbf{101} & 1:0:\textbf{100} & 7:0:\textbf{94} & 8:1:\textbf{92} \\
	\midrule
	\multirow{2}{*}{$L={2}$} & p-value &0:\textcolor{red}{\textbf{100}}:1 & 0:\textcolor{red}{\textbf{76}}:25 & 0:\textcolor{red}{\textbf{101}}:0 & 0:\textcolor{red}{\textbf{101}}:0 \\
	 & metric &11:4:\textbf{86} & 1:0:\textbf{100} & \textbf{55}:0:46 & 34:0:\textbf{67} \\
	\midrule
	\multirow{2}{*}{$L={3}$} & p-value &0:\textcolor{red}{\textbf{101}}:0 & 0:\textcolor{red}{\textbf{94}}:7 & 0:\textcolor{red}{\textbf{95}}:6 & 0:\textcolor{red}{\textbf{86}}:15 \\
	 & metric &38:2:\textbf{61} & 13:2:\textbf{86} & 17:3:\textbf{81} & 14:0:\textbf{87} \\
	\bottomrule
	\end{tabular}}
    \caption{\textsc{Centroid} on HNHN}
    \label{tab:p_metric_val_HNHN_Proteins_Centroid}

    \caption{Performance comparison on accuracy with p-value $<0.05$ and metric value on \textsc{Proteins}. For each triplet $a$:$b$:$c$, $a, b, c$ denote the number of times \textsc{GraphBPE} is \textcolor{red}{\textbf{statistically}}/\textbf{numerically} better/the same/worse compared with hypergraphs constructed by \textsc{Method} on Model (e.g., ``\textsc{Centroud} on HyperConv'' means comparing \textsc{GraphBPE} with \textsc{Centroid} on the HyperConv model).}
    \label{tab: p_metric_3hgnns_on_proteins}
\end{table}
\FloatBarrier  % Prevent floats from moving past this point
\FloatBarrier  % Prevent floats from moving past this point
\FloatBarrier  % Prevent floats from moving past this point
%%%%%%%%%%%%%%%%%%%%%%%%%%regression datasets%%%%%%%%%%%%%%%%%%%%
\newpage

\subsection{\textsc{Freesolv}}\label{app: freesolv_result}
\begin{figure}[H]
\centering
\includegraphics[width=0.9\linewidth, height=0.59\textwidth]{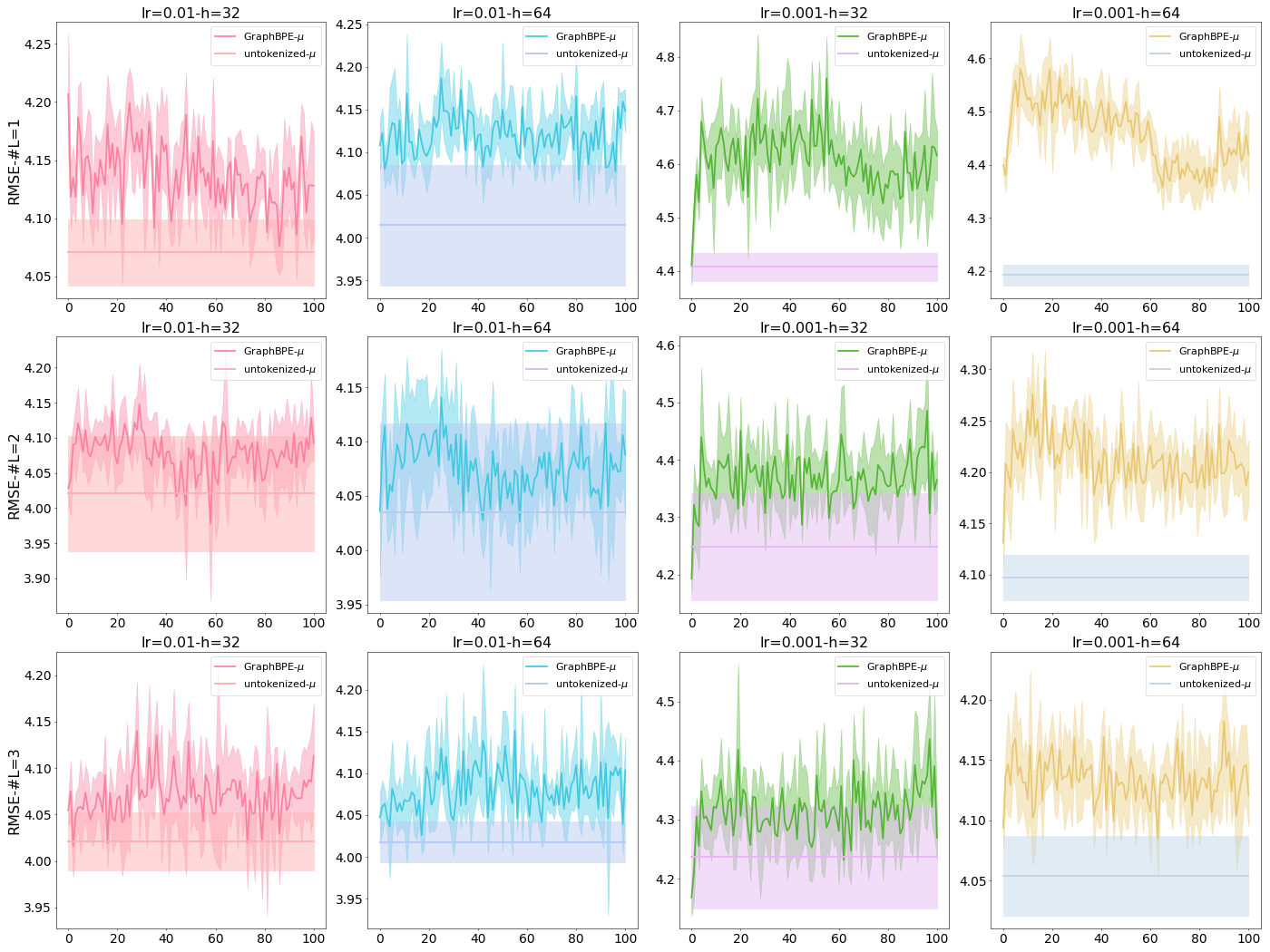}
\caption{Results of GCN on \textsc{Freesolv}, with \textbf{RMSE} the \textit{lower} the better} 
\label{fig:freesolv_gcn}
\end{figure}

\begin{figure}[H]
\centering
\includegraphics[width=0.9\linewidth, height=0.59\textwidth]{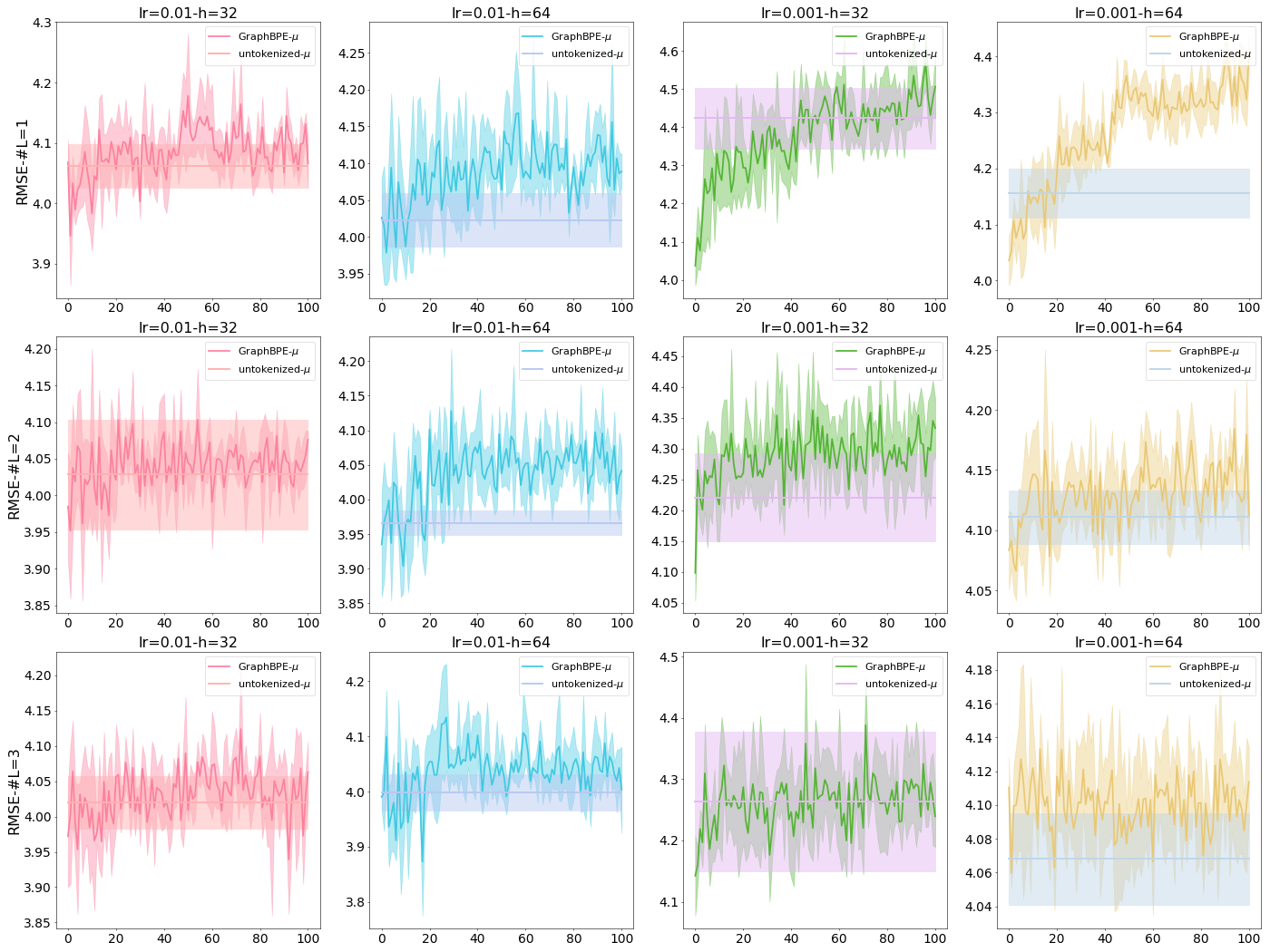}
\caption{Results of GAT on \textsc{Freesolv}, with \textbf{RMSE} the \textit{lower} the better} 
\label{fig:freesolv_gat}
\end{figure}

\begin{figure}[H]
\centering
\includegraphics[width=0.9\linewidth, height=0.59\textwidth]{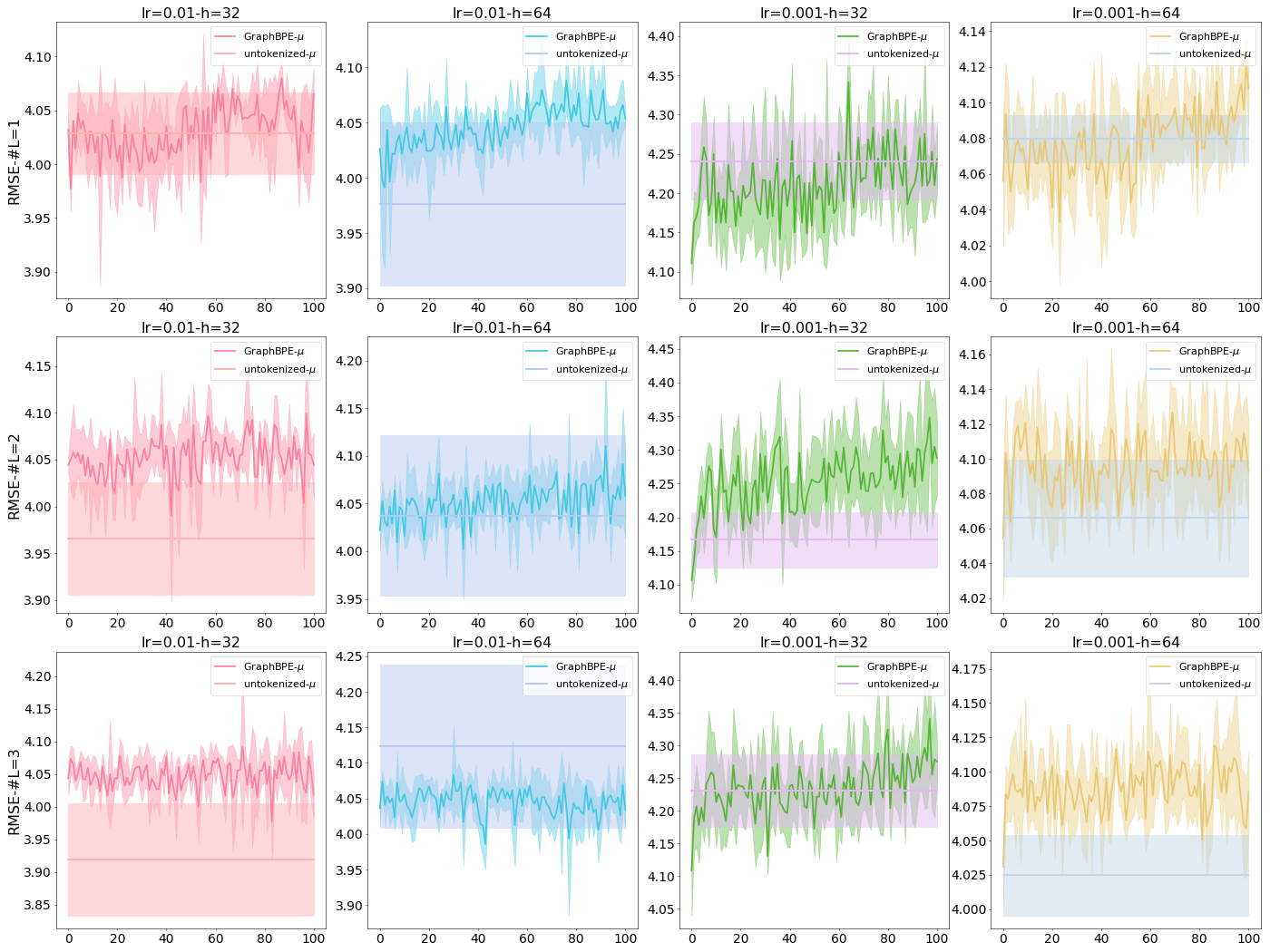}
\caption{Results of GIN on \textsc{Freesolv}, with \textbf{RMSE} the \textit{lower} the better} 
\label{fig:freesolv_gin}
\end{figure}

\begin{figure}[H]
\centering
\includegraphics[width=0.9\linewidth, height=0.59\textwidth]{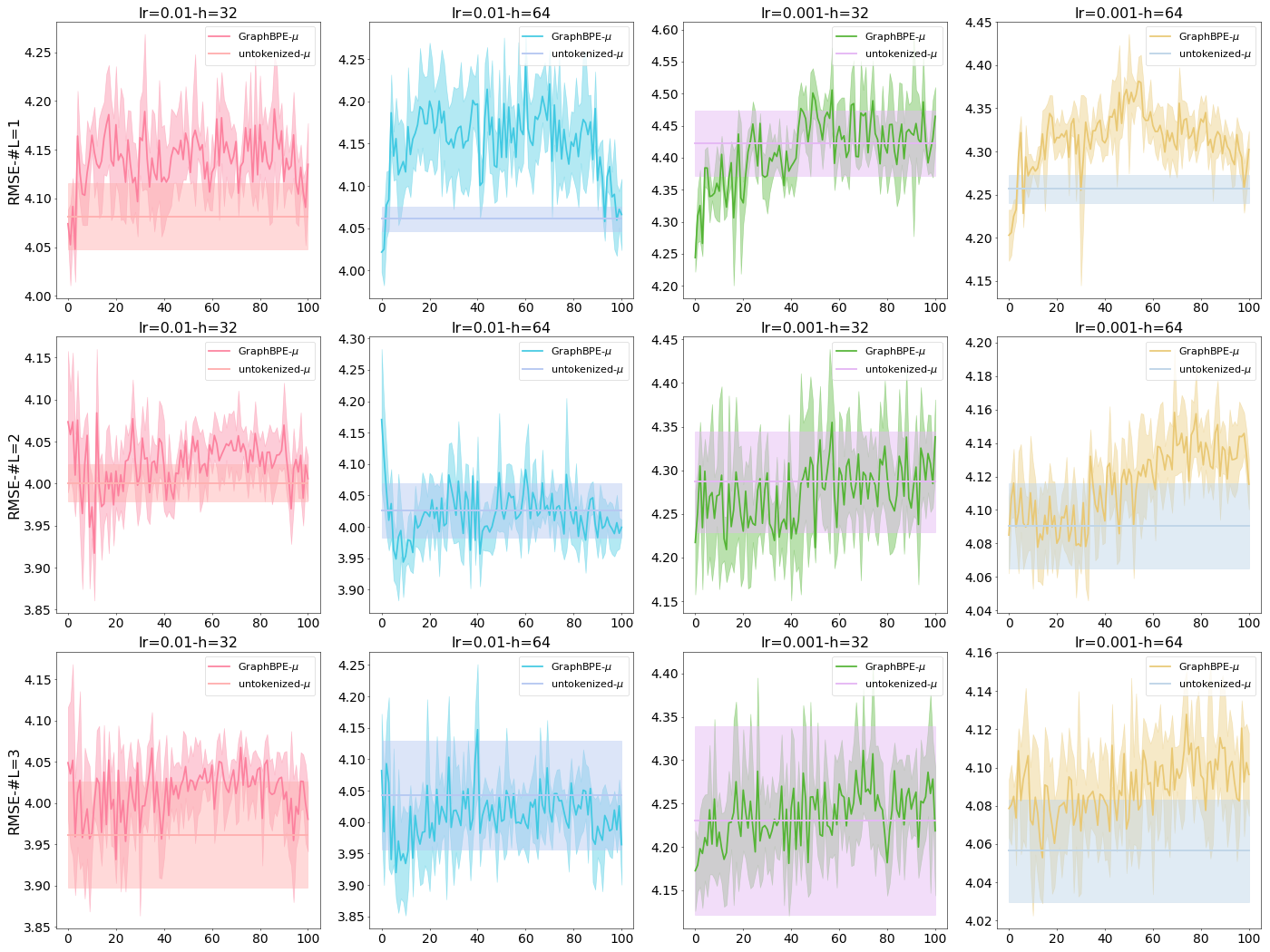}
\caption{Results of GraphSAGE on \textsc{Freesolv}, with \textbf{RMSE} the \textit{lower} the better} 
\label{fig:freesolv_graphsage}
\end{figure}

For GNNs, we include the performance comparison results in Table~\ref{tab: p_metric_4gnns_on_freesolv}, and the visualization over different tokenization steps in Figure~\ref{fig:freesolv_gcn}, ~\ref{fig:freesolv_gat}, ~\ref{fig:freesolv_gin}, and ~\ref{fig:freesolv_graphsage} for GCN, GAT, GIN, and GraphSAGE.

\begin{figure}[H]
\centering
\includegraphics[width=0.9\linewidth, height=0.52\textwidth]{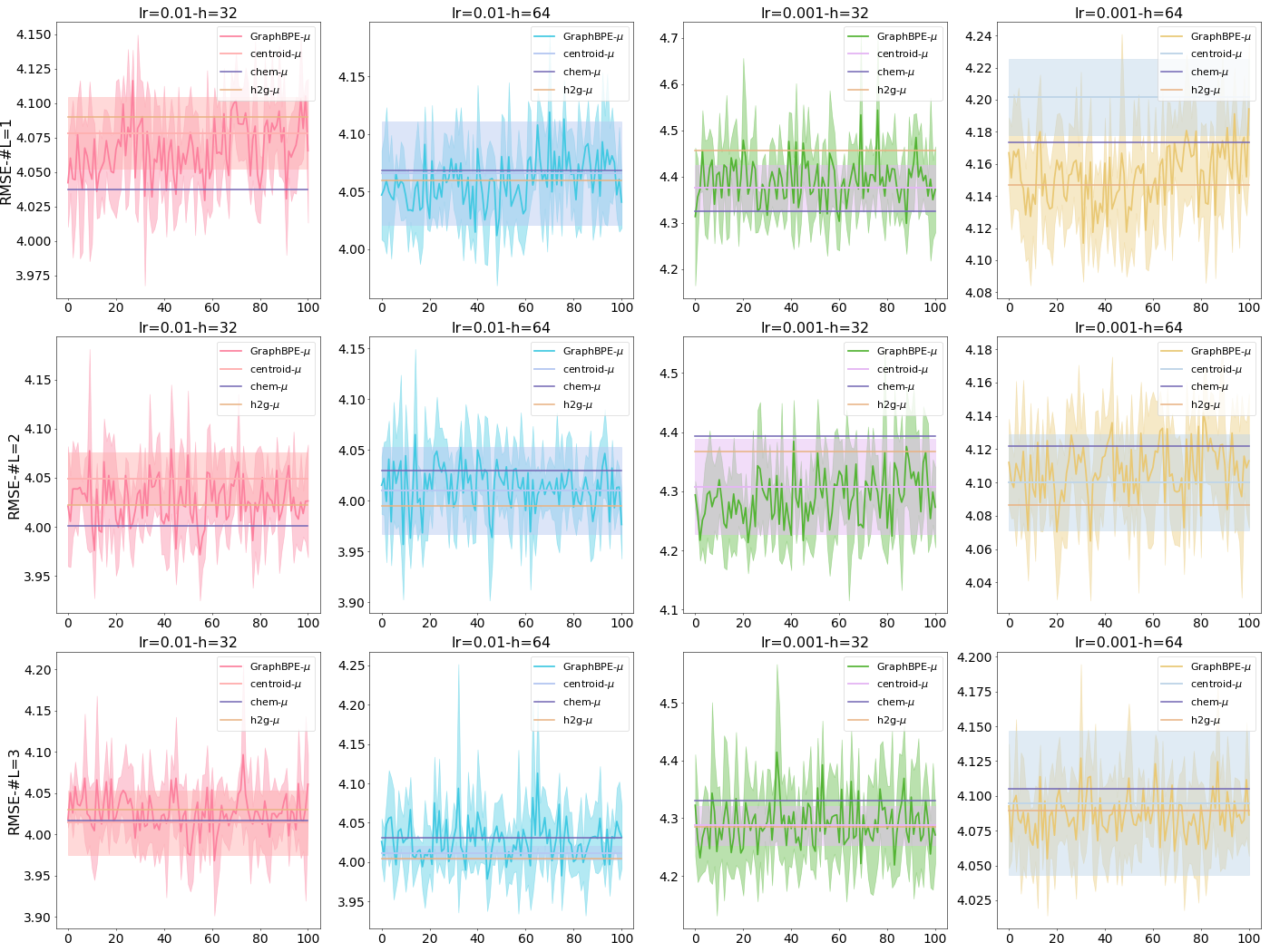}
\caption{Results of HyperConv on \textsc{Freesolv}, with \textbf{RMSE} the \textit{lower} the better} 
\label{fig:freesolv_hyperconv}
\end{figure}

\begin{figure}[H]
\centering
\includegraphics[width=0.9\linewidth, height=0.59\textwidth]{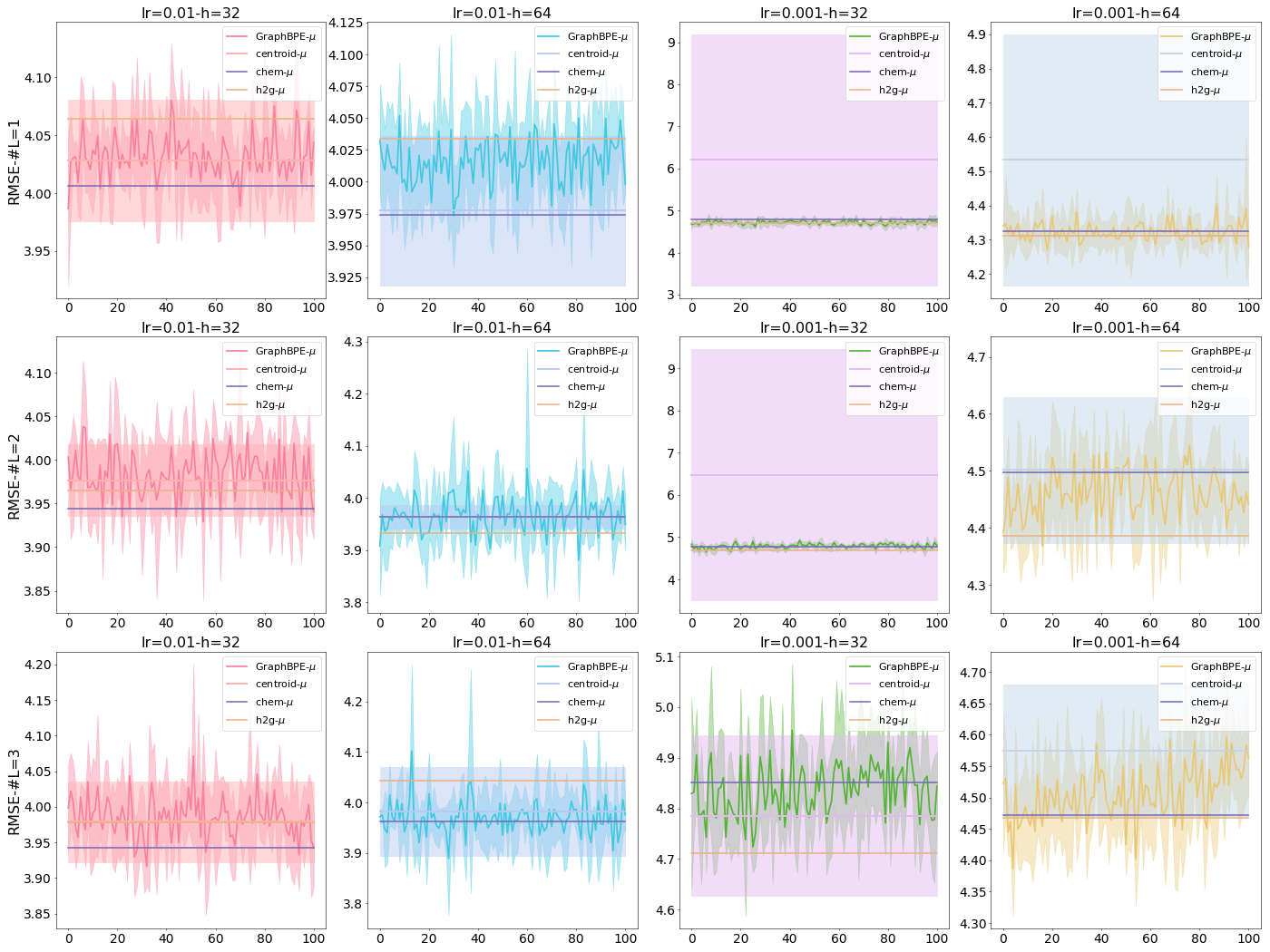}
\caption{Results of HGNN++ on \textsc{Freesolv}, with \textbf{RMSE} the \textit{lower} the better} 
\label{fig:freesolv_HGNNP}
\end{figure}

\begin{figure}[H]
\centering
\includegraphics[width=0.9\linewidth, height=0.59\textwidth]{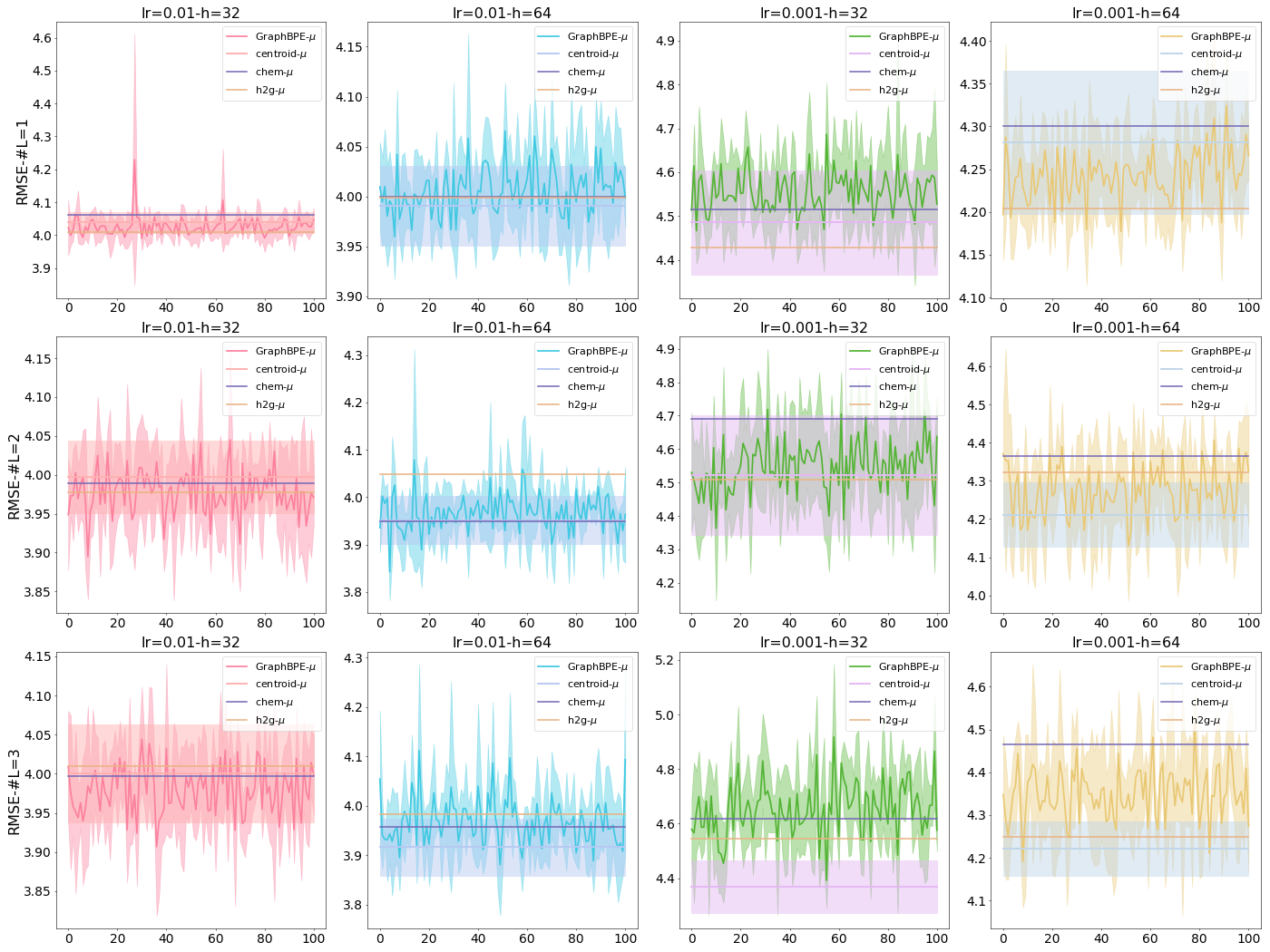}
\caption{Results of HNHN on \textsc{Freesolv}, with \textbf{RMSE} the \textit{lower} the better} 
\label{fig:freesolv_HNHN}
\end{figure}

For HyperGNNs, we include the performance comparison results in Table~\ref{tab: p_metric_3hgnns_on_freesolv}, and the visualization over different tokenization steps in Figure~\ref{fig:freesolv_hyperconv}, ~\ref{fig:freesolv_HGNNP}, and ~\ref{fig:freesolv_HNHN} for HyperConv, HGNN++, and HNHN.
\begin{table}[h]
    \centering
    \begin{minipage}[b]{0.45\linewidth}
        \centering
        \begin{subtable} % {.5\textwidth}
            \centering
	\resizebox{.99\textwidth}{!}{\begin{tabular}{lccccc}
	\toprule
	\multicolumn{2}{c}{\textbf{learning rate}} & \multicolumn{2}{c}{$10^{-2}$} & \multicolumn{2}{c}{$10^{-3}$} \\
	\multicolumn{2}{c}{\textbf{hidden size}} & $h=32$ & $h=64$ & $h=32$ & $h=64$ \\
	\midrule
	\multirow{2}{*}{$L={1}$} & p-value &0:46:\textcolor{red}{\textbf{55}} & 0:\textcolor{red}{\textbf{56}}:45 & 0:3:\textcolor{red}{\textbf{98}} & 0:0:\textcolor{red}{\textbf{101}} \\
	 & metric &0:0:\textbf{101} & 0:0:\textbf{101} & 0:0:\textbf{101} & 0:0:\textbf{101} \\
	\midrule
	\multirow{2}{*}{$L={2}$} & p-value &0:\textcolor{red}{\textbf{97}}:4 & 0:\textcolor{red}{\textbf{101}}:0 & 0:\textcolor{red}{\textbf{82}}:19 & 0:1:\textcolor{red}{\textbf{100}} \\
	 & metric &3:0:\textbf{98} & 3:0:\textbf{98} & 1:0:\textbf{100} & 0:0:\textbf{101} \\
	\midrule
	\multirow{2}{*}{$L={3}$} & p-value &0:\textcolor{red}{\textbf{71}}:30 & 0:44:\textcolor{red}{\textbf{57}} & 0:\textcolor{red}{\textbf{94}}:7 & 0:24:\textcolor{red}{\textbf{77}} \\
	 & metric &3:0:\textbf{98} & 0:0:\textbf{101} & 3:0:\textbf{98} & 0:0:\textbf{101} \\
	\bottomrule
	\end{tabular}}
	\caption{Comparison with p-/metric value of GCN}
	\label{tab:p_metric_val_GCN_Freesolv}
        \end{subtable}
        
        \vspace{0.3cm} % adjust vertical space between tables
        
        \begin{subtable} %{.5\textwidth}
            \centering
	\resizebox{.99\textwidth}{!}{\begin{tabular}{lccccc}
	\toprule
	\multicolumn{2}{c}{\textbf{learning rate}} & \multicolumn{2}{c}{$10^{-2}$} & \multicolumn{2}{c}{$10^{-3}$} \\
	\multicolumn{2}{c}{\textbf{hidden size}} & $h=32$ & $h=64$ & $h=32$ & $h=64$ \\
	\midrule
	\multirow{2}{*}{$L={1}$} & p-value &1:\textcolor{red}{\textbf{99}}:1 & 0:\textcolor{red}{\textbf{94}}:7 & 11:\textcolor{red}{\textbf{89}}:1 & 4:\textcolor{red}{\textbf{94}}:3 \\
	 & metric &46:0:\textbf{55} & 0:0:\textbf{101} & \textbf{77}:0:24 & \textbf{52}:0:49 \\
	\midrule
	\multirow{2}{*}{$L={2}$} & p-value &0:\textcolor{red}{\textbf{57}}:44 & 0:\textcolor{red}{\textbf{101}}:0 & 1:\textcolor{red}{\textbf{54}}:46 & 0:\textcolor{red}{\textbf{91}}:10 \\
	 & metric &0:0:\textbf{101} & 21:0:\textbf{80} & 2:0:\textbf{99} & 3:0:\textbf{98} \\
	\midrule
	\multirow{2}{*}{$L={3}$} & p-value &0:43:\textcolor{red}{\textbf{58}} & 1:\textcolor{red}{\textbf{100}}:0 & 0:\textcolor{red}{\textbf{99}}:2 & 0:41:\textcolor{red}{\textbf{60}} \\
	 & metric &0:0:\textbf{101} & \textbf{101}:0:0 & 46:0:\textbf{55} & 0:0:\textbf{101} \\
	\bottomrule
	\end{tabular}}
	\caption{Comparison with p-/metric value of GIN}
	\label{tab:p_metric_val_GIN_Freesolv}
        \end{subtable}
    \end{minipage}
    \hspace{0.5cm}
    \begin{minipage}[b]{0.45\linewidth}
        \centering
        \begin{subtable} %{.5\textwidth}
            \centering
	\resizebox{.99\textwidth}{!}{\begin{tabular}{lccccc}
	\toprule
	\multicolumn{2}{c}{\textbf{learning rate}} & \multicolumn{2}{c}{$10^{-2}$} & \multicolumn{2}{c}{$10^{-3}$} \\
	\multicolumn{2}{c}{\textbf{hidden size}} & $h=32$ & $h=64$ & $h=32$ & $h=64$ \\
	\midrule
	\multirow{2}{*}{$L={1}$} & p-value &2:\textcolor{red}{\textbf{93}}:6 & 0:\textcolor{red}{\textbf{67}}:34 & 13:\textcolor{red}{\textbf{87}}:1 & 3:30:\textcolor{red}{\textbf{68}} \\
	 & metric &24:0:\textbf{77} & 6:0:\textbf{95} & \textbf{62}:0:39 & 15:0:\textbf{86} \\
	\midrule
	\multirow{2}{*}{$L={2}$} & p-value &0:\textcolor{red}{\textbf{101}}:0 & 1:41:\textcolor{red}{\textbf{59}} & 1:\textcolor{red}{\textbf{88}}:12 & 1:\textcolor{red}{\textbf{89}}:11 \\
	 & metric &32:0:\textbf{69} & 8:0:\textbf{93} & 5:0:\textbf{96} & 17:0:\textbf{84} \\
	\midrule
	\multirow{2}{*}{$L={3}$} & p-value &0:\textcolor{red}{\textbf{96}}:5 & 1:\textcolor{red}{\textbf{83}}:17 & 0:\textcolor{red}{\textbf{101}}:0 & 0:\textcolor{red}{\textbf{93}}:8 \\
	 & metric &34:0:\textbf{67} & 14:0:\textbf{87} & 50:0:\textbf{51} & 3:0:\textbf{98} \\
	\bottomrule
	\end{tabular}}
	\caption{Comparison with p-/metric value of GAT}
	\label{tab:p_metric_val_GAT_Freesolv}
        \end{subtable}
        
        \vspace{0.3cm} % adjust vertical space between tables
        
        \begin{subtable} %{.5\textwidth}
            \centering
	\resizebox{.99\textwidth}{!}{\begin{tabular}{lccccc}
	\toprule
	\multicolumn{2}{c}{\textbf{learning rate}} & \multicolumn{2}{c}{$10^{-2}$} & \multicolumn{2}{c}{$10^{-3}$} \\
	\multicolumn{2}{c}{\textbf{hidden size}} & $h=32$ & $h=64$ & $h=32$ & $h=64$ \\
	\midrule
	\multirow{2}{*}{$L={1}$} & p-value &0:\textcolor{red}{\textbf{62}}:39 & 1:29:\textcolor{red}{\textbf{71}} & 9:\textcolor{red}{\textbf{88}}:4 & 2:28:\textcolor{red}{\textbf{71}} \\
	 & metric &3:0:\textbf{98} & 4:0:\textbf{97} & \textbf{53}:0:48 & 6:0:\textbf{95} \\
	\midrule
	\multirow{2}{*}{$L={2}$} & p-value &1:\textcolor{red}{\textbf{86}}:14 & 2:\textcolor{red}{\textbf{97}}:2 & 2:\textcolor{red}{\textbf{98}}:1 & 0:\textcolor{red}{\textbf{73}}:28 \\
	 & metric &19:0:\textbf{82} & \textbf{62}:0:39 & \textbf{62}:0:39 & 18:0:\textbf{83} \\
	\midrule
	\multirow{2}{*}{$L={3}$} & p-value &0:\textcolor{red}{\textbf{89}}:12 & 2:\textcolor{red}{\textbf{99}}:0 & 0:\textcolor{red}{\textbf{101}}:0 & 0:\textcolor{red}{\textbf{81}}:20 \\
	 & metric &7:0:\textbf{94} & \textbf{81}:0:20 & 44:0:\textbf{57} & 1:0:\textbf{100} \\
	\bottomrule
	\end{tabular}}
	\caption{Comparison with p-/metric value of GraphSAGE}
	\label{tab:p_metric_val_GraphSAGE_Freesolv}
        \end{subtable}
    \end{minipage}
    \caption{Performance comparison on accuracy with p-value $<0.05$ and metric value on \textsc{Freesolv}. For each triplet $a$:$b$:$c$, $a, b, c$ denote the number of times \textsc{GraphBPE} is \textcolor{red}{\textbf{statistically}}/\textbf{numerically} better/the same/worse compared with (untokenized) simple graph.}
    \label{tab: p_metric_4gnns_on_freesolv}
\end{table}
\newpage
\FloatBarrier  % Prevent floats from moving past this point
%%%%%%%%%%%%%%%%%%%%%%%%%%%%%%%%%%%%%%%%%
\FloatBarrier  % Prevent floats from moving past this point
%%%%%%%%%%%%%%%%%%%%%%%%%%%%%%%%%%%%%%%%%
\FloatBarrier  % Prevent floats from moving past this point
%%%%%%%%%%%%%%%%%%%%%%%%%%%%%%%%%%%%%%%%%
\FloatBarrier  % Prevent floats from moving past this point
\newpage
\begin{table}[h]
    \centering
    \begin{minipage}[b]{0.32\linewidth}
        \centering
        \begin{subtable}
            \centering
	\resizebox{.99\textwidth}{!}{\begin{tabular}{lccccc}
	\toprule
	\multicolumn{2}{c}{\textbf{learning rate}} & \multicolumn{2}{c}{$10^{-2}$} & \multicolumn{2}{c}{$10^{-3}$} \\
	\multicolumn{2}{c}{\textbf{hidden size}} & $h=32$ & $h=64$ & $h=32$ & $h=64$ \\
	\midrule
	\multirow{2}{*}{$L={1}$} & p-value &4:\textcolor{red}{\textbf{96}}:1 & 1:\textcolor{red}{\textbf{99}}:1 & 0:\textcolor{red}{\textbf{100}}:1 & 41:\textcolor{red}{\textbf{60}}:0 \\
	 & metric &\textbf{71}:0:30 & \textbf{66}:0:35 & 36:0:\textbf{65} & \textbf{101}:0:0 \\
	\midrule
	\multirow{2}{*}{$L={2}$} & p-value &14:\textcolor{red}{\textbf{87}}:0 & 0:\textcolor{red}{\textbf{101}}:0 & 0:\textcolor{red}{\textbf{101}}:0 & 0:\textcolor{red}{\textbf{100}}:1 \\
	 & metric &\textbf{84}:0:17 & 38:0:\textbf{63} & \textbf{71}:0:30 & 31:0:\textbf{70} \\
	\midrule
	\multirow{2}{*}{$L={3}$} & p-value &0:\textcolor{red}{\textbf{101}}:0 & 1:\textcolor{red}{\textbf{98}}:2 & 1:\textcolor{red}{\textbf{99}}:1 & 0:\textcolor{red}{\textbf{101}}:0 \\
	 & metric &26:0:\textbf{75} & 26:0:\textbf{75} & \textbf{53}:0:48 & \textbf{77}:0:24 \\
	\bottomrule
	\end{tabular}}
	\caption{\textsc{Centroid} on HyperConv}
	\label{tab:p_metric_val_HyperConv_Freesolv_Centroid}
        \end{subtable}
        
        \vspace{0.25cm} % adjust vertical space between tables
        
        \begin{subtable}
            \centering
	\resizebox{.99\textwidth}{!}{\begin{tabular}{lccccc}
	\toprule
	\multicolumn{2}{c}{\textbf{learning rate}} & \multicolumn{2}{c}{$10^{-2}$} & \multicolumn{2}{c}{$10^{-3}$} \\
	\multicolumn{2}{c}{\textbf{hidden size}} & $h=32$ & $h=64$ & $h=32$ & $h=64$ \\
	\midrule
	\multirow{2}{*}{$L={1}$} & p-value &0:\textcolor{red}{\textbf{93}}:8 & 0:\textcolor{red}{\textbf{101}}:0 & 0:\textcolor{red}{\textbf{92}}:9 & 1:\textcolor{red}{\textbf{100}}:0 \\
	 & metric &7:0:\textbf{94} & \textbf{70}:0:31 & 6:0:\textbf{95} & \textbf{92}:0:9 \\
	\midrule
	\multirow{2}{*}{$L={2}$} & p-value &0:\textcolor{red}{\textbf{88}}:13 & 6:\textcolor{red}{\textbf{95}}:0 & \textcolor{red}{\textbf{61}}:40:0 & 8:\textcolor{red}{\textbf{93}}:0 \\
	 & metric &12:0:\textbf{89} & \textbf{79}:0:22 & \textbf{101}:0:0 & \textbf{81}:0:20 \\
	\midrule
	\multirow{2}{*}{$L={3}$} & p-value &0:\textcolor{red}{\textbf{99}}:2 & 1:\textcolor{red}{\textbf{100}}:0 & 6:\textcolor{red}{\textbf{95}}:0 & 0:\textcolor{red}{\textbf{101}}:0 \\
	 & metric &28:0:\textbf{73} & \textbf{62}:0:39 & \textbf{89}:0:12 & \textbf{89}:0:12 \\
	\bottomrule
	\end{tabular}}
	\caption{\textsc{Chem} on HyperConv}
	\label{tab:p_metric_val_HyperConv_Freesolv_Chem}
        \end{subtable}

        \vspace{0.25cm} % adjust vertical space between tables
        
        \begin{subtable}
            \centering
	\resizebox{.99\textwidth}{!}{\begin{tabular}{lccccc}
	\toprule
	\multicolumn{2}{c}{\textbf{learning rate}} & \multicolumn{2}{c}{$10^{-2}$} & \multicolumn{2}{c}{$10^{-3}$} \\
	\multicolumn{2}{c}{\textbf{hidden size}} & $h=32$ & $h=64$ & $h=32$ & $h=64$ \\
	\midrule
	\multirow{2}{*}{$L={1}$} & p-value &8:\textcolor{red}{\textbf{93}}:0 & 1:\textcolor{red}{\textbf{100}}:0 & 0:\textcolor{red}{\textbf{101}}:0 & 0:\textcolor{red}{\textbf{99}}:2 \\
	 & metric &\textbf{89}:0:12 & \textbf{55}:0:46 & \textbf{94}:0:7 & 38:0:\textbf{63} \\
	\midrule
	\multirow{2}{*}{$L={2}$} & p-value &1:\textcolor{red}{\textbf{100}}:0 & 0:\textcolor{red}{\textbf{97}}:4 & 6:\textcolor{red}{\textbf{95}}:0 & 0:\textcolor{red}{\textbf{99}}:2 \\
	 & metric &43:0:\textbf{58} & 18:0:\textbf{83} & \textbf{99}:0:2 & 9:0:\textbf{92} \\
	\midrule
	\multirow{2}{*}{$L={3}$} & p-value &1:\textcolor{red}{\textbf{100}}:0 & 0:\textcolor{red}{\textbf{95}}:6 & 0:\textcolor{red}{\textbf{100}}:1 & 0:\textcolor{red}{\textbf{101}}:0 \\
	 & metric &\textbf{60}:0:41 & 17:0:\textbf{84} & 50:0:\textbf{51} & \textbf{61}:0:40 \\
	\bottomrule
	\end{tabular}}
	\caption{\textsc{H2g} on HyperConv}
	\label{tab:p_metric_val_HyperConv_Freesolv_H2g}
        \end{subtable}
    \end{minipage}
    \hfill
    \begin{minipage}[b]{0.32\linewidth}
        \centering
        \begin{subtable}
            \centering
	\resizebox{.99\textwidth}{!}{\begin{tabular}{lccccc}
	\toprule
	\multicolumn{2}{c}{\textbf{learning rate}} & \multicolumn{2}{c}{$10^{-2}$} & \multicolumn{2}{c}{$10^{-3}$} \\
	\multicolumn{2}{c}{\textbf{hidden size}} & $h=32$ & $h=64$ & $h=32$ & $h=64$ \\
	\midrule
	\multirow{2}{*}{$L={1}$} & p-value &0:\textcolor{red}{\textbf{101}}:0 & 0:\textcolor{red}{\textbf{94}}:7 & 0:\textcolor{red}{\textbf{101}}:0 & 0:\textcolor{red}{\textbf{101}}:0 \\
	 & metric &43:0:\textbf{58} & 1:0:\textbf{100} & \textbf{101}:0:0 & \textbf{101}:0:0 \\
	\midrule
	\multirow{2}{*}{$L={2}$} & p-value &1:\textcolor{red}{\textbf{98}}:2 & 1:\textcolor{red}{\textbf{97}}:3 & 0:\textcolor{red}{\textbf{101}}:0 & 1:\textcolor{red}{\textbf{100}}:0 \\
	 & metric &41:0:\textbf{60} & 43:0:\textbf{58} & \textbf{101}:0:0 & \textbf{83}:0:18 \\
	\midrule
	\multirow{2}{*}{$L={3}$} & p-value &0:\textcolor{red}{\textbf{101}}:0 & 0:\textcolor{red}{\textbf{101}}:0 & 0:\textcolor{red}{\textbf{99}}:2 & 2:\textcolor{red}{\textbf{99}}:0 \\
	 & metric &38:0:\textbf{63} & \textbf{73}:0:28 & 19:0:\textbf{82} & \textbf{95}:0:6 \\
	\bottomrule
	\end{tabular}}
	\caption{\textsc{Centroid} on HGNN++}
	\label{tab:p_metric_val_HGNN++_Freesolv_Centroid}
        \end{subtable}
        
        \vspace{0.32cm} % adjust vertical space between tables
        
        \begin{subtable}
            \centering
	\resizebox{.99\textwidth}{!}{\begin{tabular}{lccccc}
	\toprule
	\multicolumn{2}{c}{\textbf{learning rate}} & \multicolumn{2}{c}{$10^{-2}$} & \multicolumn{2}{c}{$10^{-3}$} \\
	\multicolumn{2}{c}{\textbf{hidden size}} & $h=32$ & $h=64$ & $h=32$ & $h=64$ \\
	\midrule
	\multirow{2}{*}{$L={1}$} & p-value &0:\textcolor{red}{\textbf{87}}:14 & 0:\textcolor{red}{\textbf{73}}:28 & 4:\textcolor{red}{\textbf{97}}:0 & 0:\textcolor{red}{\textbf{101}}:0 \\
	 & metric &6:0:\textbf{95} & 1:0:\textbf{100} & \textbf{96}:0:5 & \textbf{56}:0:45 \\
	\midrule
	\multirow{2}{*}{$L={2}$} & p-value &0:\textcolor{red}{\textbf{91}}:10 & 0:\textcolor{red}{\textbf{101}}:0 & 0:\textcolor{red}{\textbf{100}}:1 & 11:\textcolor{red}{\textbf{90}}:0 \\
	 & metric &7:0:\textbf{94} & 43:0:\textbf{58} & 42:0:\textbf{59} & \textbf{81}:0:20 \\
	\midrule
	\multirow{2}{*}{$L={3}$} & p-value &0:\textcolor{red}{\textbf{97}}:4 & 0:\textcolor{red}{\textbf{99}}:2 & 1:\textcolor{red}{\textbf{100}}:0 & 1:\textcolor{red}{\textbf{98}}:2 \\
	 & metric &5:0:\textbf{96} & 42:0:\textbf{59} & \textbf{59}:0:42 & 22:0:\textbf{79} \\
	\bottomrule
	\end{tabular}}
	\caption{\textsc{Chem} on HGNN++}
	\label{tab:p_metric_val_HGNN++_Freesolv_Chem}
        \end{subtable}

        \vspace{0.32cm} % adjust vertical space between tables
        
        \begin{subtable}
            \centering
	\resizebox{.99\textwidth}{!}{\begin{tabular}{lccccc}
	\toprule
	\multicolumn{2}{c}{\textbf{learning rate}} & \multicolumn{2}{c}{$10^{-2}$} & \multicolumn{2}{c}{$10^{-3}$} \\
	\multicolumn{2}{c}{\textbf{hidden size}} & $h=32$ & $h=64$ & $h=32$ & $h=64$ \\
	\midrule
	\multirow{2}{*}{$L={1}$} & p-value &10:\textcolor{red}{\textbf{91}}:0 & 1:\textcolor{red}{\textbf{100}}:0 & 0:\textcolor{red}{\textbf{101}}:0 & 0:\textcolor{red}{\textbf{99}}:2 \\
	 & metric &\textbf{95}:0:6 & \textbf{84}:0:17 & 38:0:\textbf{63} & 33:0:\textbf{68} \\
	\midrule
	\multirow{2}{*}{$L={2}$} & p-value &0:\textcolor{red}{\textbf{100}}:1 & 0:\textcolor{red}{\textbf{97}}:4 & 0:\textcolor{red}{\textbf{79}}:22 & 0:\textcolor{red}{\textbf{82}}:19 \\
	 & metric &24:0:\textbf{77} & 13:0:\textbf{88} & 5:0:\textbf{96} & 2:0:\textbf{99} \\
	\midrule
	\multirow{2}{*}{$L={3}$} & p-value &0:\textcolor{red}{\textbf{99}}:2 & 13:\textcolor{red}{\textbf{88}}:0 & 0:\textcolor{red}{\textbf{89}}:12 & 0:\textcolor{red}{\textbf{95}}:6 \\
	 & metric &35:0:\textbf{66} & \textbf{100}:0:1 & 0:0:\textbf{101} & 19:0:\textbf{82} \\
	\bottomrule
	\end{tabular}}
	\caption{\textsc{H2g} on HGNN++}
	\label{tab:p_metric_val_HGNN++_Freesolv_H2g}
        \end{subtable}
    \end{minipage}
    \hfill
    \begin{minipage}[b]{0.32\linewidth}
        \centering
        \begin{subtable}{}
            \centering
	\resizebox{.99\textwidth}{!}{\begin{tabular}{lccccc}
	\toprule
	\multicolumn{2}{c}{\textbf{learning rate}} & \multicolumn{2}{c}{$10^{-2}$} & \multicolumn{2}{c}{$10^{-3}$} \\
	\multicolumn{2}{c}{\textbf{hidden size}} & $h=32$ & $h=64$ & $h=32$ & $h=64$ \\
	\midrule
	\multirow{2}{*}{$L={1}$} & p-value &7:\textcolor{red}{\textbf{94}}:0 & 0:\textcolor{red}{\textbf{99}}:2 & 0:\textcolor{red}{\textbf{97}}:4 & 0:\textcolor{red}{\textbf{101}}:0 \\
	 & metric &\textbf{79}:0:22 & 27:0:\textbf{74} & 5:0:\textbf{96} & \textbf{92}:0:9 \\
	\midrule
	\multirow{2}{*}{$L={2}$} & p-value &2:\textcolor{red}{\textbf{99}}:0 & 2:\textcolor{red}{\textbf{99}}:0 & 0:\textcolor{red}{\textbf{101}}:0 & 0:\textcolor{red}{\textbf{95}}:6 \\
	 & metric &\textbf{81}:0:20 & 34:0:\textbf{67} & 35:0:\textbf{66} & 21:0:\textbf{80} \\
	\midrule
	\multirow{2}{*}{$L={3}$} & p-value &1:\textcolor{red}{\textbf{100}}:0 & 0:\textcolor{red}{\textbf{92}}:9 & 0:39:\textcolor{red}{\textbf{62}} & 0:\textcolor{red}{\textbf{65}}:36 \\
	 & metric &\textbf{75}:0:26 & 9:0:\textbf{92} & 0:0:\textbf{101} & 2:0:\textbf{99} \\
	\bottomrule
	\end{tabular}}
	\caption{\textsc{Centroid} on HNHN}
	\label{tab:p_metric_val_HNHN_Freesolv_Centroid}
        \end{subtable}
        
        \vspace{0.32cm} % adjust vertical space between tables
        
        \begin{subtable}
            \centering
	\resizebox{.99\textwidth}{!}{\begin{tabular}{lccccc}
	\toprule
	\multicolumn{2}{c}{\textbf{learning rate}} & \multicolumn{2}{c}{$10^{-2}$} & \multicolumn{2}{c}{$10^{-3}$} \\
	\multicolumn{2}{c}{\textbf{hidden size}} & $h=32$ & $h=64$ & $h=32$ & $h=64$ \\
	\midrule
	\multirow{2}{*}{$L={1}$} & p-value &26:\textcolor{red}{\textbf{75}}:0 & 0:\textcolor{red}{\textbf{99}}:2 & 0:\textcolor{red}{\textbf{100}}:1 & 4:\textcolor{red}{\textbf{97}}:0 \\
	 & metric &\textbf{98}:0:3 & 39:0:\textbf{62} & 19:0:\textbf{82} & \textbf{99}:0:2 \\
	\midrule
	\multirow{2}{*}{$L={2}$} & p-value &0:\textcolor{red}{\textbf{101}}:0 & 0:\textcolor{red}{\textbf{97}}:4 & 20:\textcolor{red}{\textbf{81}}:0 & 3:\textcolor{red}{\textbf{98}}:0 \\
	 & metric &\textbf{71}:0:30 & 30:0:\textbf{71} & \textbf{97}:0:4 & \textbf{91}:0:10 \\
	\midrule
	\multirow{2}{*}{$L={3}$} & p-value &0:\textcolor{red}{\textbf{100}}:1 & 1:\textcolor{red}{\textbf{99}}:1 & 1:\textcolor{red}{\textbf{99}}:1 & 14:\textcolor{red}{\textbf{87}}:0 \\
	 & metric &\textbf{71}:0:30 & 41:0:\textbf{60} & 38:0:\textbf{63} & \textbf{99}:0:2 \\
	\bottomrule
	\end{tabular}}
	\caption{\textsc{Chem} on HNHN}
	\label{tab:p_metric_val_HNHN_Freesolv_Chem}
        \end{subtable}

        \vspace{0.32cm} % adjust vertical space between tables
        
        \begin{subtable}
            \centering
	\resizebox{.99\textwidth}{!}{\begin{tabular}{lccccc}
	\toprule
	\multicolumn{2}{c}{\textbf{learning rate}} & \multicolumn{2}{c}{$10^{-2}$} & \multicolumn{2}{c}{$10^{-3}$} \\
	\multicolumn{2}{c}{\textbf{hidden size}} & $h=32$ & $h=64$ & $h=32$ & $h=64$ \\
	\midrule
	\multirow{2}{*}{$L={1}$} & p-value &0:\textcolor{red}{\textbf{94}}:7 & 0:\textcolor{red}{\textbf{101}}:0 & 0:\textcolor{red}{\textbf{67}}:34 & 0:\textcolor{red}{\textbf{96}}:5 \\
	 & metric &24:0:\textbf{77} & 38:0:\textbf{63} & 0:0:\textbf{101} & 8:0:\textbf{93} \\
	\midrule
	\multirow{2}{*}{$L={2}$} & p-value &0:\textcolor{red}{\textbf{101}}:0 & 35:\textcolor{red}{\textbf{66}}:0 & 0:\textcolor{red}{\textbf{100}}:1 & 6:\textcolor{red}{\textbf{95}}:0 \\
	 & metric &49:0:\textbf{52} & \textbf{99}:0:2 & 26:0:\textbf{75} & \textbf{76}:0:25 \\
	\midrule
	\multirow{2}{*}{$L={3}$} & p-value &0:\textcolor{red}{\textbf{101}}:0 & 2:\textcolor{red}{\textbf{99}}:0 & 0:\textcolor{red}{\textbf{93}}:8 & 0:\textcolor{red}{\textbf{90}}:11 \\
	 & metric &\textbf{83}:0:18 & \textbf{69}:0:32 & 12:0:\textbf{89} & 3:0:\textbf{98} \\
	\bottomrule
	\end{tabular}}
	\caption{\textsc{H2g} on HNHN}
	\label{tab:p_metric_val_HNHN_Freesolv_H2g}
        \end{subtable}
    \end{minipage}
    \caption{Performance comparison on accuracy with p-value $<0.05$ and metric value on \textsc{Freesolv}. For each triplet $a$:$b$:$c$, $a, b, c$ denote the number of times \textsc{GraphBPE} is \textcolor{red}{\textbf{statistically}}/\textbf{numerically} better/the same/worse compared with hypergraphs constructed by \textsc{Method} on Model (e.g., ``\textsc{Centroud} on HyperConv'' means comparing \textsc{GraphBPE} with \textsc{Centroid} on the HyperConv model).}
    \label{tab: p_metric_3hgnns_on_freesolv}
\end{table}
\FloatBarrier  % Prevent floats from moving past this point
\FloatBarrier  % Prevent floats from moving past this point
\FloatBarrier  % Prevent floats from moving past this point
%%%%%%%%%%%%%%%%%%%%%%%%%%%%%%%%%%%%%%%%%
\newpage

\subsection{\textsc{Esol}}\label{app: esol_result}
\begin{figure}[H]
\centering
\includegraphics[width=0.9\linewidth, height=0.59\textwidth]{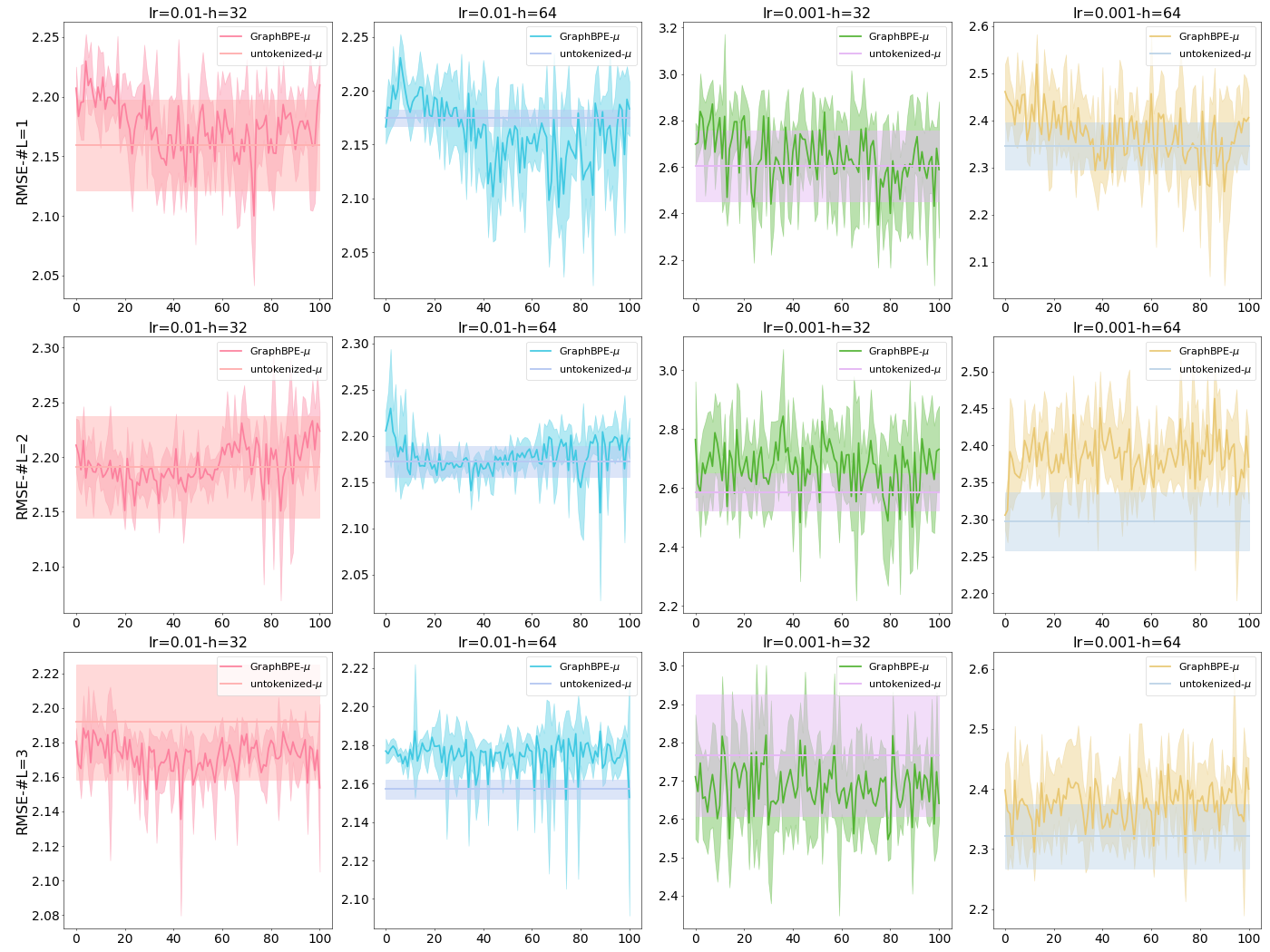}
\caption{Results of GCN on \textsc{Esol}, with \textbf{RMSE} the \textit{lower} the better} 
\label{fig:esol_gcn}
\end{figure}

\begin{figure}[H]
\centering
\includegraphics[width=0.9\linewidth, height=0.59\textwidth]{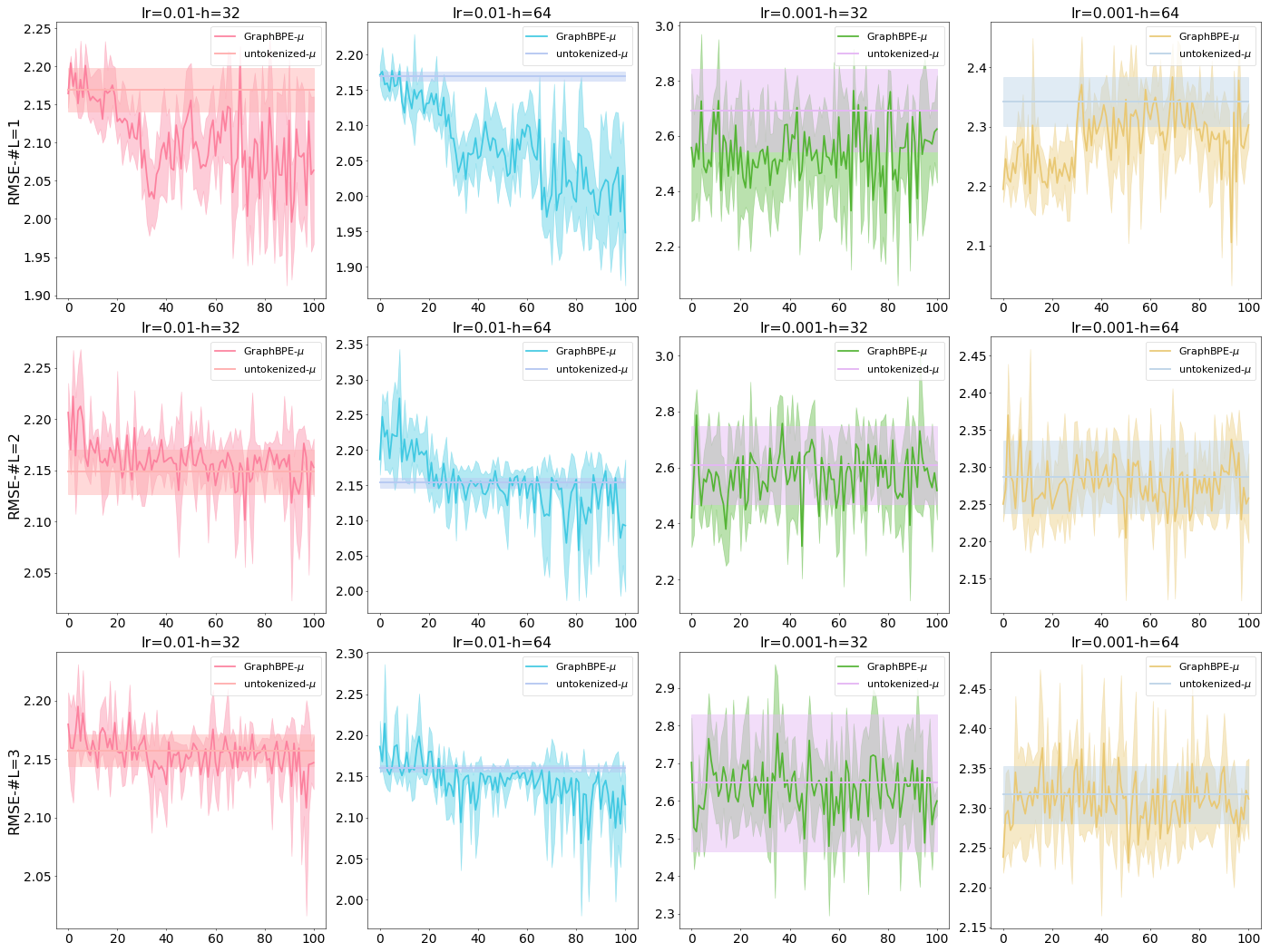}
\caption{Results of GAT on \textsc{Esol}, with \textbf{RMSE} the \textit{lower} the better} 
\label{fig:esol_gat}
\end{figure}

\begin{figure}[H]
\centering
\includegraphics[width=0.9\linewidth, height=0.59\textwidth]{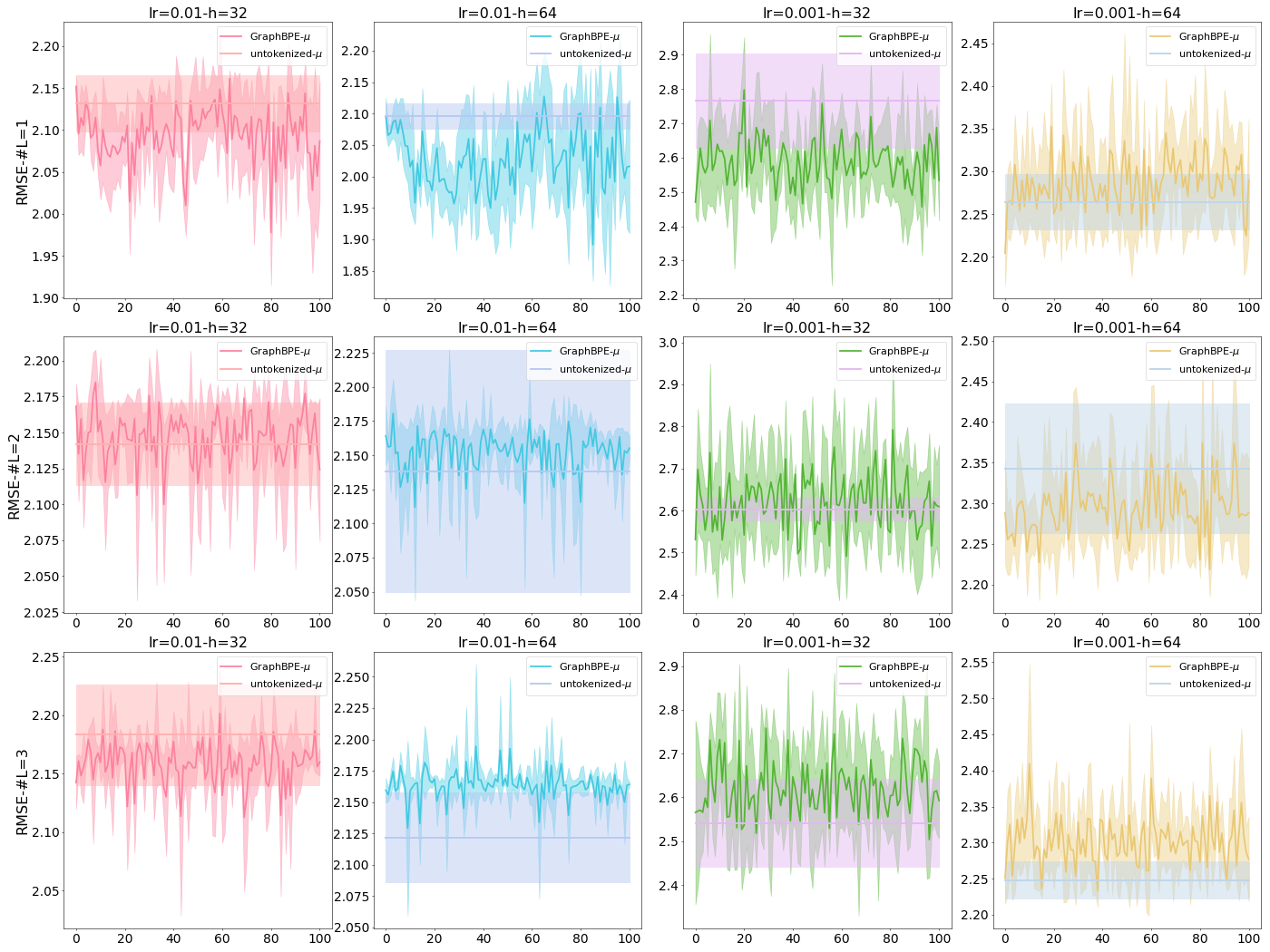}
\caption{Results of GIN on \textsc{Esol}, with \textbf{RMSE} the \textit{lower} the better} 
\label{fig:esol_gin}
\end{figure}

\begin{figure}[H]
\centering
\includegraphics[width=0.9\linewidth, height=0.59\textwidth]{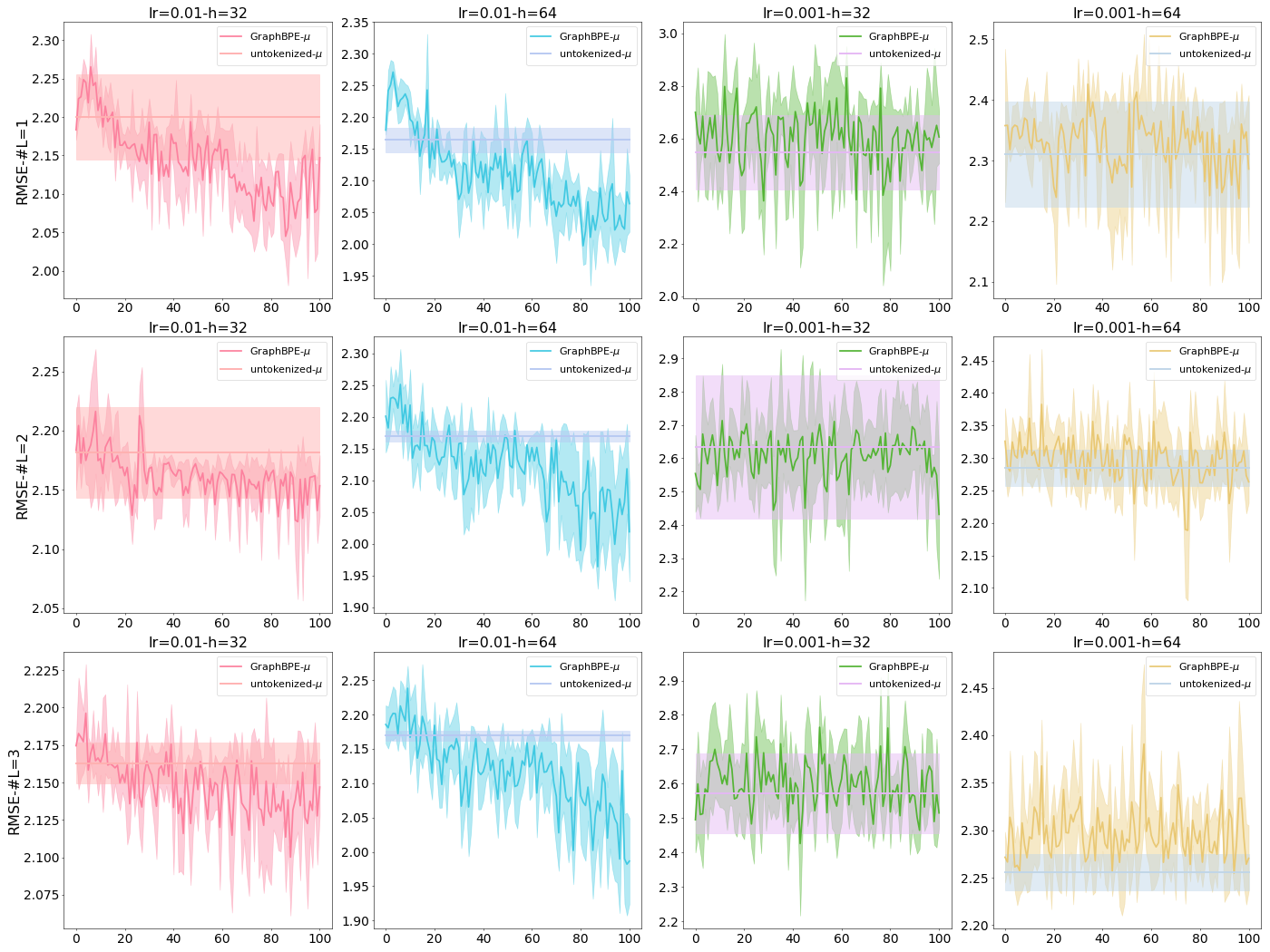}
\caption{Results of GraphSAGE on \textsc{Esol}, with \textbf{RMSE} the \textit{lower} the better} 
\label{fig:esol_graphsage}
\end{figure}

For GNNs, we include the performance comparison results in Table~\ref{tab: p_metric_4gnns_on_esol}, and the visualization over different tokenization steps in Figure~\ref{fig:esol_gcn}, ~\ref{fig:esol_gat}, ~\ref{fig:esol_gin}, and ~\ref{fig:esol_graphsage} for GCN, GAT, GIN, and GraphSAGE.

\begin{figure}[H]
\centering
\includegraphics[width=0.9\linewidth, height=0.52\textwidth]{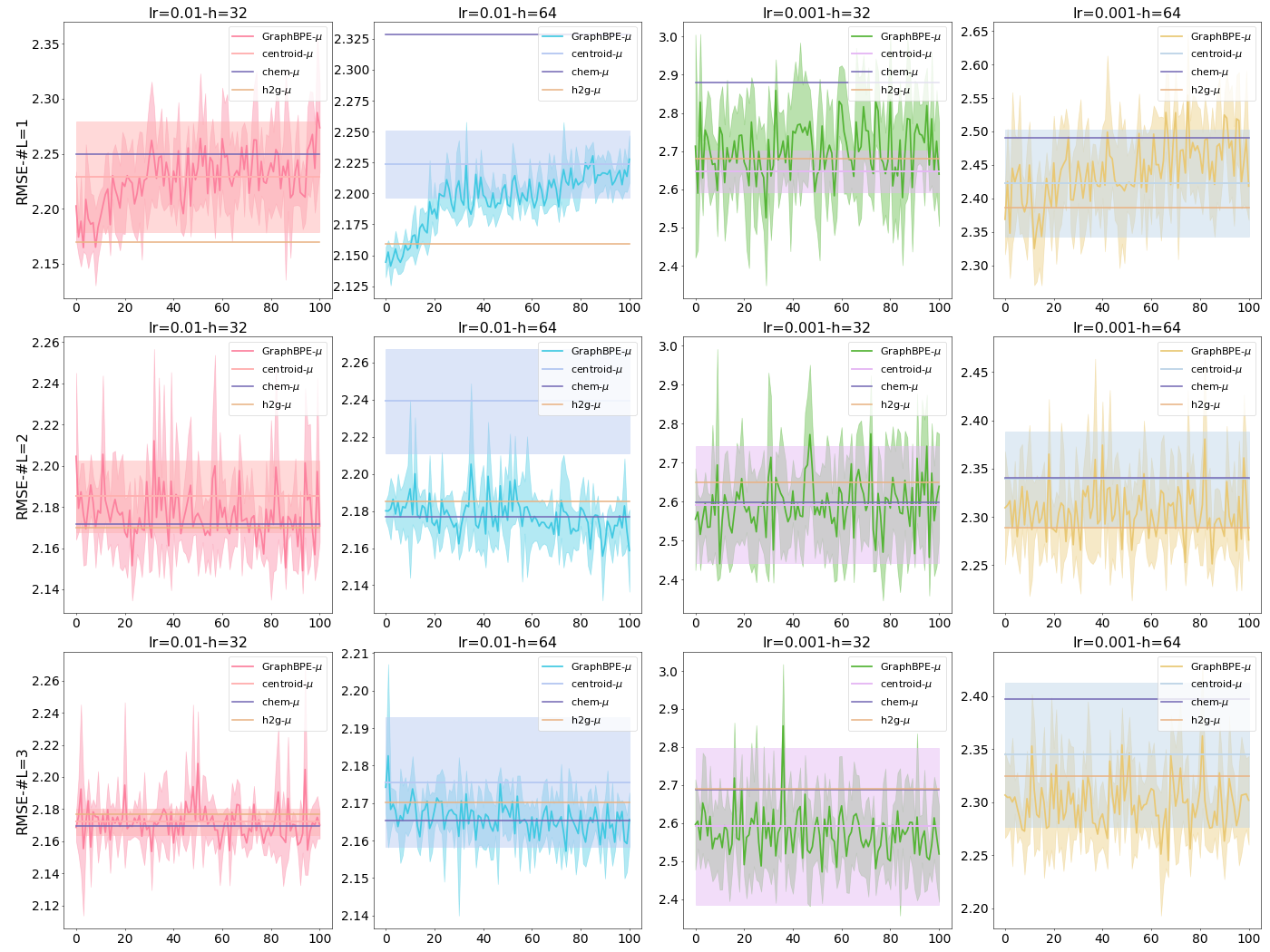}
\caption{Results of HyperConv on \textsc{Esol}, with \textbf{RMSE} the \textit{lower} the better} 
\label{fig:esol_hyperconv}
\end{figure}

\begin{figure}[H]
\centering
\includegraphics[width=0.9\linewidth, height=0.59\textwidth]{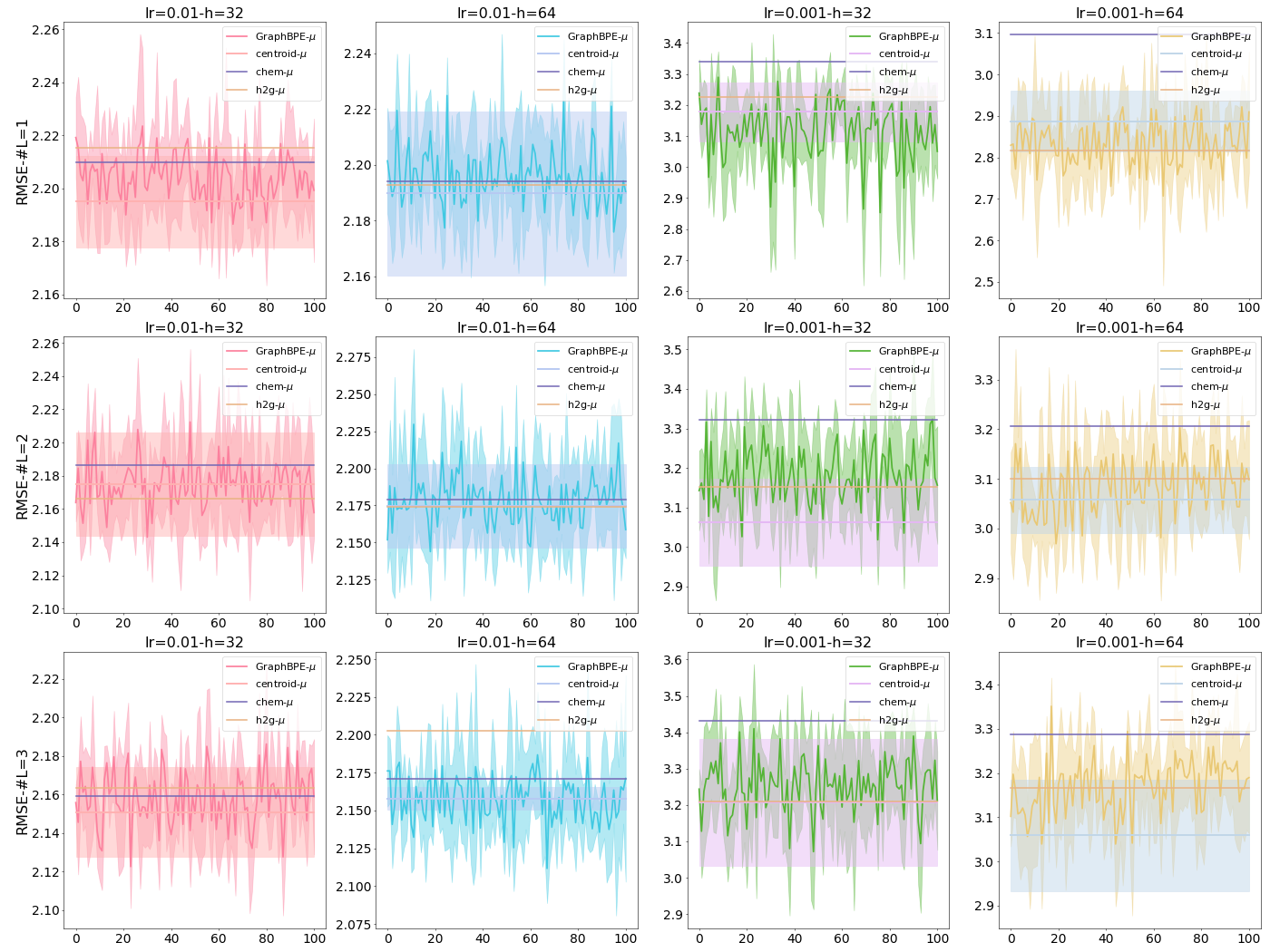}
\caption{Results of HGNN++ on \textsc{Esol}, with \textbf{RMSE} the \textit{lower} the better} 
\label{fig:esol_HGNNP}
\end{figure}

\begin{figure}[H]
\centering
\includegraphics[width=0.9\linewidth, height=0.59\textwidth]{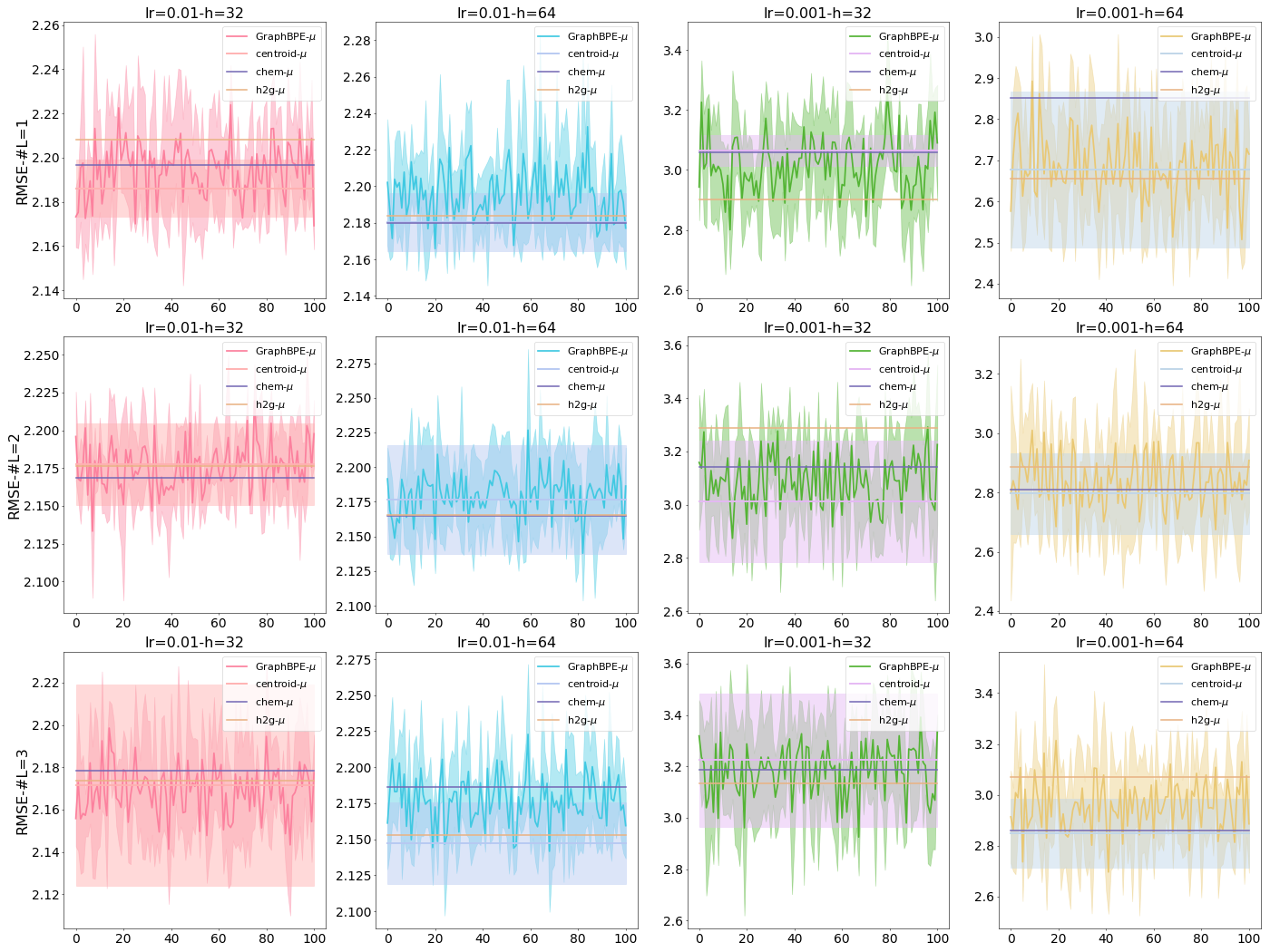}
\caption{Results of HNHN on \textsc{Esol}, with \textbf{RMSE} the \textit{lower} the better} 
\label{fig:esol_HNHN}
\end{figure}

For HyperGNNs, we include the performance comparison results in Table~\ref{tab: p_metric_3hgnns_on_esol}, and the visualization over different tokenization steps in Figure~\ref{fig:esol_hyperconv}, ~\ref{fig:esol_HGNNP}, and ~\ref{fig:esol_HNHN} for HyperConv, HGNN++, and HNHN.
\begin{table}[h]
    \centering
    \begin{minipage}[b]{0.45\linewidth}
        \centering
        \begin{subtable} % {.5\textwidth}
            \centering
	\resizebox{.99\textwidth}{!}{\begin{tabular}{lccccc}
	\toprule
	\multicolumn{2}{c}{\textbf{learning rate}} & \multicolumn{2}{c}{$10^{-2}$} & \multicolumn{2}{c}{$10^{-3}$} \\
	\multicolumn{2}{c}{\textbf{hidden size}} & $h=32$ & $h=64$ & $h=32$ & $h=64$ \\
	\midrule
	\multirow{2}{*}{$L={1}$} & p-value &0:\textcolor{red}{\textbf{97}}:4 & 12:\textcolor{red}{\textbf{85}}:4 & 0:\textcolor{red}{\textbf{94}}:7 & 0:\textcolor{red}{\textbf{94}}:7 \\
	 & metric &24:0:\textbf{77} & \textbf{64}:0:37 & 38:0:\textbf{63} & 30:0:\textbf{71} \\
	\midrule
	\multirow{2}{*}{$L={2}$} & p-value &0:\textcolor{red}{\textbf{101}}:0 & 0:\textcolor{red}{\textbf{98}}:3 & 0:\textcolor{red}{\textbf{94}}:7 & 0:48:\textcolor{red}{\textbf{53}} \\
	 & metric &\textbf{55}:0:46 & 43:0:\textbf{58} & 17:0:\textbf{84} & 0:0:\textbf{101} \\
	\midrule
	\multirow{2}{*}{$L={3}$} & p-value &1:\textcolor{red}{\textbf{100}}:0 & 0:27:\textcolor{red}{\textbf{74}} & 3:\textcolor{red}{\textbf{98}}:0 & 0:\textcolor{red}{\textbf{92}}:9 \\
	 & metric &\textbf{101}:0:0 & 3:0:\textbf{98} & \textbf{92}:0:9 & 5:0:\textbf{96} \\
	\bottomrule
	\end{tabular}}
	\caption{Comparison with p-/metric value of GCN}
	\label{tab:p_metric_val_GCN_Esol}
        \end{subtable}
        
        \vspace{0.3cm} % adjust vertical space between tables
        
        \begin{subtable} %{.5\textwidth}
            \centering
	\resizebox{.99\textwidth}{!}{\begin{tabular}{lccccc}
	\toprule
	\multicolumn{2}{c}{\textbf{learning rate}} & \multicolumn{2}{c}{$10^{-2}$} & \multicolumn{2}{c}{$10^{-3}$} \\
	\multicolumn{2}{c}{\textbf{hidden size}} & $h=32$ & $h=64$ & $h=32$ & $h=64$ \\
	\midrule
	\multirow{2}{*}{$L={1}$} & p-value &9:\textcolor{red}{\textbf{92}}:0 & 27:\textcolor{red}{\textbf{74}}:0 & 27:\textcolor{red}{\textbf{74}}:0 & 1:\textcolor{red}{\textbf{95}}:5 \\
	 & metric &\textbf{90}:0:11 & \textbf{95}:0:6 & \textbf{100}:0:1 & 20:0:\textbf{81} \\
	\midrule
	\multirow{2}{*}{$L={2}$} & p-value &0:\textcolor{red}{\textbf{100}}:1 & 0:\textcolor{red}{\textbf{101}}:0 & 4:\textcolor{red}{\textbf{95}}:2 & 7:\textcolor{red}{\textbf{94}}:0 \\
	 & metric &34:0:\textbf{67} & 18:0:\textbf{83} & 40:0:\textbf{61} & \textbf{92}:0:9 \\
	\midrule
	\multirow{2}{*}{$L={3}$} & p-value &2:\textcolor{red}{\textbf{99}}:0 & 0:\textcolor{red}{\textbf{88}}:13 & 0:\textcolor{red}{\textbf{92}}:9 & 0:\textcolor{red}{\textbf{81}}:20 \\
	 & metric &\textbf{95}:0:6 & 0:0:\textbf{101} & 7:0:\textbf{94} & 3:0:\textbf{98} \\
	\bottomrule
	\end{tabular}}
	\caption{Comparison with p-/metric value of GIN}
	\label{tab:p_metric_val_GIN_Esol}
        \end{subtable}
    \end{minipage}
    \hspace{0.5cm}
    \begin{minipage}[b]{0.45\linewidth}
        \centering
        \begin{subtable} %{.5\textwidth}
            \centering
	\resizebox{.99\textwidth}{!}{\begin{tabular}{lccccc}
	\toprule
	\multicolumn{2}{c}{\textbf{learning rate}} & \multicolumn{2}{c}{$10^{-2}$} & \multicolumn{2}{c}{$10^{-3}$} \\
	\multicolumn{2}{c}{\textbf{hidden size}} & $h=32$ & $h=64$ & $h=32$ & $h=64$ \\
	\midrule
	\multirow{2}{*}{$L={1}$} & p-value &31:\textcolor{red}{\textbf{70}}:0 & \textcolor{red}{\textbf{75}}:26:0 & 14:\textcolor{red}{\textbf{87}}:0 & 29:\textcolor{red}{\textbf{72}}:0 \\
	 & metric &\textbf{93}:0:8 & \textbf{97}:0:4 & \textbf{93}:0:8 & \textbf{92}:0:9 \\
	\midrule
	\multirow{2}{*}{$L={2}$} & p-value &0:\textcolor{red}{\textbf{95}}:6 & 1:\textcolor{red}{\textbf{90}}:10 & 3:\textcolor{red}{\textbf{98}}:0 & 3:\textcolor{red}{\textbf{98}}:0 \\
	 & metric &18:0:\textbf{83} & \textbf{55}:0:46 & \textbf{67}:0:34 & \textbf{65}:0:36 \\
	\midrule
	\multirow{2}{*}{$L={3}$} & p-value &0:\textcolor{red}{\textbf{101}}:0 & 7:\textcolor{red}{\textbf{94}}:0 & 0:\textcolor{red}{\textbf{101}}:0 & 9:\textcolor{red}{\textbf{92}}:0 \\
	 & metric &\textbf{61}:0:40 & \textbf{83}:0:18 & \textbf{57}:0:44 & \textbf{68}:0:33 \\
	\bottomrule
	\end{tabular}}
	\caption{Comparison with p-/metric value of GAT}
	\label{tab:p_metric_val_GAT_Esol}
        \end{subtable}
        
        \vspace{0.3cm} % adjust vertical space between tables
        
        \begin{subtable} %{.5\textwidth}
            \centering
	\resizebox{.99\textwidth}{!}{\begin{tabular}{lccccc}
	\toprule
	\multicolumn{2}{c}{\textbf{learning rate}} & \multicolumn{2}{c}{$10^{-2}$} & \multicolumn{2}{c}{$10^{-3}$} \\
	\multicolumn{2}{c}{\textbf{hidden size}} & $h=32$ & $h=64$ & $h=32$ & $h=64$ \\
	\midrule
	\multirow{2}{*}{$L={1}$} & p-value &24:\textcolor{red}{\textbf{77}}:0 & 43:\textcolor{red}{\textbf{49}}:9 & 0:\textcolor{red}{\textbf{96}}:5 & 0:\textcolor{red}{\textbf{101}}:0 \\
	 & metric &\textbf{89}:0:12 & \textbf{84}:0:17 & 27:0:\textbf{74} & 32:0:\textbf{69} \\
	\midrule
	\multirow{2}{*}{$L={2}$} & p-value &3:\textcolor{red}{\textbf{98}}:0 & 15:\textcolor{red}{\textbf{85}}:1 & 0:\textcolor{red}{\textbf{101}}:0 & 0:\textcolor{red}{\textbf{97}}:4 \\
	 & metric &\textbf{89}:0:12 & \textbf{86}:0:15 & \textbf{64}:0:37 & 34:0:\textbf{67} \\
	\midrule
	\multirow{2}{*}{$L={3}$} & p-value &9:\textcolor{red}{\textbf{92}}:0 & 27:\textcolor{red}{\textbf{73}}:1 & 0:\textcolor{red}{\textbf{98}}:3 & 0:\textcolor{red}{\textbf{86}}:15 \\
	 & metric &\textbf{82}:0:19 & \textbf{86}:0:15 & 33:0:\textbf{68} & 1:0:\textbf{100} \\
	\bottomrule
	\end{tabular}}
	\caption{Comparison with p-/metric value of GraphSAGE}
	\label{tab:p_metric_val_GraphSAGE_Esol}
        \end{subtable}
    \end{minipage}
    \caption{Performance comparison on accuracy with p-value $<0.05$ and metric value on \textsc{Esol}. For each triplet $a$:$b$:$c$, $a, b, c$ denote the number of times \textsc{GraphBPE} is \textcolor{red}{\textbf{statistically}}/\textbf{numerically} better/the same/worse compared with (untokenized) simple graph.}
    \label{tab: p_metric_4gnns_on_esol}
\end{table}
\newpage
\FloatBarrier  % Prevent floats from moving past this point
%%%%%%%%%%%%%%%%%%%%%%%%%%%%%%%%%%%%%%%%%
\FloatBarrier  % Prevent floats from moving past this point
%%%%%%%%%%%%%%%%%%%%%%%%%%%%%%%%%%%%%%%%%
\FloatBarrier  % Prevent floats from moving past this point
%%%%%%%%%%%%%%%%%%%%%%%%%%%%%%%%%%%%%%%%%
\FloatBarrier  % Prevent floats from moving past this point
\newpage
\begin{table}[h]
    \centering
    \begin{minipage}[b]{0.32\linewidth}
        \centering
        \begin{subtable}
            \centering
	\resizebox{.99\textwidth}{!}{\begin{tabular}{lccccc}
	\toprule
	\multicolumn{2}{c}{\textbf{learning rate}} & \multicolumn{2}{c}{$10^{-2}$} & \multicolumn{2}{c}{$10^{-3}$} \\
	\multicolumn{2}{c}{\textbf{hidden size}} & $h=32$ & $h=64$ & $h=32$ & $h=64$ \\
	\midrule
	\multirow{2}{*}{$L={1}$} & p-value &1:\textcolor{red}{\textbf{100}}:0 & 17:\textcolor{red}{\textbf{84}}:0 & 0:\textcolor{red}{\textbf{92}}:9 & 0:\textcolor{red}{\textbf{99}}:2 \\
	 & metric &\textbf{51}:0:50 & \textbf{97}:0:4 & 21:0:\textbf{80} & 34:0:\textbf{67} \\
	\midrule
	\multirow{2}{*}{$L={2}$} & p-value &6:\textcolor{red}{\textbf{95}}:0 & \textcolor{red}{\textbf{87}}:14:0 & 0:\textcolor{red}{\textbf{100}}:1 & 4:\textcolor{red}{\textbf{97}}:0 \\
	 & metric &\textbf{83}:0:18 & \textbf{101}:0:0 & \textbf{53}:0:48 & \textbf{94}:0:7 \\
	\midrule
	\multirow{2}{*}{$L={3}$} & p-value &2:\textcolor{red}{\textbf{99}}:0 & 0:\textcolor{red}{\textbf{101}}:0 & 0:\textcolor{red}{\textbf{101}}:0 & 4:\textcolor{red}{\textbf{97}}:0 \\
	 & metric &\textbf{57}:0:44 & \textbf{100}:0:1 & \textbf{66}:0:35 & \textbf{96}:0:5 \\
	\bottomrule
	\end{tabular}}
	\caption{\textsc{Centroid} on HyperConv}
	\label{tab:p_metric_val_HyperConv_Esol_Centroid}
        \end{subtable}
        
        \vspace{0.25cm} % adjust vertical space between tables
        
        \begin{subtable}
            \centering
	\resizebox{.99\textwidth}{!}{\begin{tabular}{lccccc}
	\toprule
	\multicolumn{2}{c}{\textbf{learning rate}} & \multicolumn{2}{c}{$10^{-2}$} & \multicolumn{2}{c}{$10^{-3}$} \\
	\multicolumn{2}{c}{\textbf{hidden size}} & $h=32$ & $h=64$ & $h=32$ & $h=64$ \\
	\midrule
	\multirow{2}{*}{$L={1}$} & p-value &2:\textcolor{red}{\textbf{99}}:0 & \textcolor{red}{\textbf{101}}:0:0 & 35:\textcolor{red}{\textbf{66}}:0 & 3:\textcolor{red}{\textbf{98}}:0 \\
	 & metric &\textbf{85}:0:16 & \textbf{101}:0:0 & \textbf{101}:0:0 & \textbf{90}:0:11 \\
	\midrule
	\multirow{2}{*}{$L={2}$} & p-value &0:\textcolor{red}{\textbf{101}}:0 & 0:\textcolor{red}{\textbf{99}}:2 & 5:\textcolor{red}{\textbf{96}}:0 & 9:\textcolor{red}{\textbf{92}}:0 \\
	 & metric &36:0:\textbf{65} & \textbf{51}:0:50 & \textbf{57}:0:44 & \textbf{94}:0:7 \\
	\midrule
	\multirow{2}{*}{$L={3}$} & p-value &1:\textcolor{red}{\textbf{99}}:1 & 1:\textcolor{red}{\textbf{99}}:1 & 20:\textcolor{red}{\textbf{80}}:1 & 10:\textcolor{red}{\textbf{91}}:0 \\
	 & metric &43:0:\textbf{58} & 45:0:\textbf{56} & \textbf{98}:0:3 & \textbf{101}:0:0 \\
	\bottomrule
	\end{tabular}}
	\caption{\textsc{Chem} on HyperConv}
	\label{tab:p_metric_val_HyperConv_Esol_Chem}
        \end{subtable}

        \vspace{0.25cm} % adjust vertical space between tables
        
        \begin{subtable}
            \centering
	\resizebox{.99\textwidth}{!}{\begin{tabular}{lccccc}
	\toprule
	\multicolumn{2}{c}{\textbf{learning rate}} & \multicolumn{2}{c}{$10^{-2}$} & \multicolumn{2}{c}{$10^{-3}$} \\
	\multicolumn{2}{c}{\textbf{hidden size}} & $h=32$ & $h=64$ & $h=32$ & $h=64$ \\
	\midrule
	\multirow{2}{*}{$L={1}$} & p-value &0:\textcolor{red}{\textbf{57}}:44 & 1:20:\textcolor{red}{\textbf{80}} & 0:\textcolor{red}{\textbf{100}}:1 & 0:\textcolor{red}{\textbf{99}}:2 \\
	 & metric &2:0:\textbf{99} & 12:0:\textbf{89} & 41:0:\textbf{60} & 8:0:\textbf{93} \\
	\midrule
	\multirow{2}{*}{$L={2}$} & p-value &2:\textcolor{red}{\textbf{98}}:1 & 8:\textcolor{red}{\textbf{93}}:0 & 0:\textcolor{red}{\textbf{101}}:0 & 0:\textcolor{red}{\textbf{99}}:2 \\
	 & metric &31:0:\textbf{70} & \textbf{87}:0:14 & \textbf{82}:0:19 & 26:0:\textbf{75} \\
	\midrule
	\multirow{2}{*}{$L={3}$} & p-value &11:\textcolor{red}{\textbf{90}}:0 & 9:\textcolor{red}{\textbf{92}}:0 & 7:\textcolor{red}{\textbf{94}}:0 & 3:\textcolor{red}{\textbf{98}}:0 \\
	 & metric &\textbf{78}:0:23 & \textbf{88}:0:13 & \textbf{98}:0:3 & \textbf{91}:0:10 \\
	\bottomrule
	\end{tabular}}
	\caption{\textsc{H2g} on HyperConv}
	\label{tab:p_metric_val_HyperConv_Esol_H2g}
        \end{subtable}
    \end{minipage}
    \hfill
    \begin{minipage}[b]{0.32\linewidth}
        \centering
        \begin{subtable}
            \centering
	\resizebox{.99\textwidth}{!}{\begin{tabular}{lccccc}
	\toprule
	\multicolumn{2}{c}{\textbf{learning rate}} & \multicolumn{2}{c}{$10^{-2}$} & \multicolumn{2}{c}{$10^{-3}$} \\
	\multicolumn{2}{c}{\textbf{hidden size}} & $h=32$ & $h=64$ & $h=32$ & $h=64$ \\
	\midrule
	\multirow{2}{*}{$L={1}$} & p-value &0:\textcolor{red}{\textbf{98}}:3 & 0:\textcolor{red}{\textbf{101}}:0 & 3:\textcolor{red}{\textbf{98}}:0 & 7:\textcolor{red}{\textbf{94}}:0 \\
	 & metric &14:0:\textbf{87} & 27:0:\textbf{74} & \textbf{76}:0:25 & \textbf{85}:0:16 \\
	\midrule
	\multirow{2}{*}{$L={2}$} & p-value &0:\textcolor{red}{\textbf{101}}:0 & 0:\textcolor{red}{\textbf{101}}:0 & 0:\textcolor{red}{\textbf{81}}:20 & 1:\textcolor{red}{\textbf{97}}:3 \\
	 & metric &\textbf{53}:0:48 & 42:0:\textbf{59} & 3:0:\textbf{98} & 36:0:\textbf{65} \\
	\midrule
	\multirow{2}{*}{$L={3}$} & p-value &1:\textcolor{red}{\textbf{99}}:1 & 3:\textcolor{red}{\textbf{96}}:2 & 0:\textcolor{red}{\textbf{100}}:1 & 0:\textcolor{red}{\textbf{90}}:11 \\
	 & metric &29:0:\textbf{72} & 48:0:\textbf{53} & 21:0:\textbf{80} & 4:0:\textbf{97} \\
	\bottomrule
	\end{tabular}}
	\caption{\textsc{Centroid} on HGNN++}
	\label{tab:p_metric_val_HGNN++_Esol_Centroid}
        \end{subtable}
        
        \vspace{0.32cm} % adjust vertical space between tables
        
        \begin{subtable}
            \centering
	\resizebox{.99\textwidth}{!}{\begin{tabular}{lccccc}
	\toprule
	\multicolumn{2}{c}{\textbf{learning rate}} & \multicolumn{2}{c}{$10^{-2}$} & \multicolumn{2}{c}{$10^{-3}$} \\
	\multicolumn{2}{c}{\textbf{hidden size}} & $h=32$ & $h=64$ & $h=32$ & $h=64$ \\
	\midrule
	\multirow{2}{*}{$L={1}$} & p-value &2:\textcolor{red}{\textbf{99}}:0 & 0:\textcolor{red}{\textbf{100}}:1 & 29:\textcolor{red}{\textbf{72}}:0 & 38:\textcolor{red}{\textbf{63}}:0 \\
	 & metric &\textbf{77}:0:24 & 47:0:\textbf{54} & \textbf{101}:0:0 & \textbf{101}:0:0 \\
	\midrule
	\multirow{2}{*}{$L={2}$} & p-value &0:\textcolor{red}{\textbf{101}}:0 & 1:\textcolor{red}{\textbf{100}}:0 & 12:\textcolor{red}{\textbf{89}}:0 & 43:\textcolor{red}{\textbf{58}}:0 \\
	 & metric &\textbf{87}:0:14 & 48:0:\textbf{53} & \textbf{100}:0:1 & \textbf{101}:0:0 \\
	\midrule
	\multirow{2}{*}{$L={3}$} & p-value &3:\textcolor{red}{\textbf{97}}:1 & 1:\textcolor{red}{\textbf{100}}:0 & 32:\textcolor{red}{\textbf{69}}:0 & 17:\textcolor{red}{\textbf{84}}:0 \\
	 & metric &\textbf{58}:0:43 & \textbf{81}:0:20 & \textbf{101}:0:0 & \textbf{99}:0:2 \\
	\bottomrule
	\end{tabular}}
	\caption{\textsc{Chem} on HGNN++}
	\label{tab:p_metric_val_HGNN++_Esol_Chem}
        \end{subtable}

        \vspace{0.32cm} % adjust vertical space between tables
        
        \begin{subtable}
            \centering
	\resizebox{.99\textwidth}{!}{\begin{tabular}{lccccc}
	\toprule
	\multicolumn{2}{c}{\textbf{learning rate}} & \multicolumn{2}{c}{$10^{-2}$} & \multicolumn{2}{c}{$10^{-3}$} \\
	\multicolumn{2}{c}{\textbf{hidden size}} & $h=32$ & $h=64$ & $h=32$ & $h=64$ \\
	\midrule
	\multirow{2}{*}{$L={1}$} & p-value &0:\textcolor{red}{\textbf{101}}:0 & 0:\textcolor{red}{\textbf{101}}:0 & 2:\textcolor{red}{\textbf{99}}:0 & 2:\textcolor{red}{\textbf{95}}:4 \\
	 & metric &\textbf{91}:0:10 & 43:0:\textbf{58} & \textbf{93}:0:8 & 37:0:\textbf{64} \\
	\midrule
	\multirow{2}{*}{$L={2}$} & p-value &0:\textcolor{red}{\textbf{101}}:0 & 0:\textcolor{red}{\textbf{101}}:0 & 2:\textcolor{red}{\textbf{97}}:2 & 0:\textcolor{red}{\textbf{101}}:0 \\
	 & metric &25:0:\textbf{76} & 41:0:\textbf{60} & 32:0:\textbf{69} & \textbf{61}:0:40 \\
	\midrule
	\multirow{2}{*}{$L={3}$} & p-value &6:\textcolor{red}{\textbf{94}}:1 & 13:\textcolor{red}{\textbf{88}}:0 & 0:\textcolor{red}{\textbf{101}}:0 & 0:\textcolor{red}{\textbf{98}}:3 \\
	 & metric &\textbf{65}:0:36 & \textbf{101}:0:0 & 21:0:\textbf{80} & 47:0:\textbf{54} \\
	\bottomrule
	\end{tabular}}
	\caption{\textsc{H2g} on HGNN++}
	\label{tab:p_metric_val_HGNN++_Esol_H2g}
        \end{subtable}
    \end{minipage}
    \hfill
    \begin{minipage}[b]{0.32\linewidth}
        \centering
        \begin{subtable}{}
            \centering
	\resizebox{.99\textwidth}{!}{\begin{tabular}{lccccc}
	\toprule
	\multicolumn{2}{c}{\textbf{learning rate}} & \multicolumn{2}{c}{$10^{-2}$} & \multicolumn{2}{c}{$10^{-3}$} \\
	\multicolumn{2}{c}{\textbf{hidden size}} & $h=32$ & $h=64$ & $h=32$ & $h=64$ \\
	\midrule
	\multirow{2}{*}{$L={1}$} & p-value &0:\textcolor{red}{\textbf{96}}:5 & 0:\textcolor{red}{\textbf{93}}:8 & 4:\textcolor{red}{\textbf{95}}:2 & 1:\textcolor{red}{\textbf{100}}:0 \\
	 & metric &24:0:\textbf{77} & 11:0:\textbf{90} & \textbf{74}:0:27 & 48:0:\textbf{53} \\
	\midrule
	\multirow{2}{*}{$L={2}$} & p-value &0:\textcolor{red}{\textbf{101}}:0 & 0:\textcolor{red}{\textbf{101}}:0 & 0:\textcolor{red}{\textbf{96}}:5 & 0:\textcolor{red}{\textbf{99}}:2 \\
	 & metric &\textbf{58}:0:43 & 37:0:\textbf{64} & 19:0:\textbf{82} & 36:0:\textbf{65} \\
	\midrule
	\multirow{2}{*}{$L={3}$} & p-value &0:\textcolor{red}{\textbf{101}}:0 & 0:\textcolor{red}{\textbf{90}}:11 & 2:\textcolor{red}{\textbf{99}}:0 & 1:\textcolor{red}{\textbf{91}}:9 \\
	 & metric &\textbf{58}:0:43 & 5:0:\textbf{96} & \textbf{64}:0:37 & 13:0:\textbf{88} \\
	\bottomrule
	\end{tabular}}
	\caption{\textsc{Centroid} on HNHN}
	\label{tab:p_metric_val_HNHN_Esol_Centroid}
        \end{subtable}
        
        \vspace{0.32cm} % adjust vertical space between tables
        
        \begin{subtable}
            \centering
	\resizebox{.99\textwidth}{!}{\begin{tabular}{lccccc}
	\toprule
	\multicolumn{2}{c}{\textbf{learning rate}} & \multicolumn{2}{c}{$10^{-2}$} & \multicolumn{2}{c}{$10^{-3}$} \\
	\multicolumn{2}{c}{\textbf{hidden size}} & $h=32$ & $h=64$ & $h=32$ & $h=64$ \\
	\midrule
	\multirow{2}{*}{$L={1}$} & p-value &0:\textcolor{red}{\textbf{101}}:0 & 0:\textcolor{red}{\textbf{97}}:4 & 2:\textcolor{red}{\textbf{98}}:1 & 11:\textcolor{red}{\textbf{90}}:0 \\
	 & metric &\textbf{52}:0:49 & 11:0:\textbf{90} & \textbf{71}:0:30 & \textbf{99}:0:2 \\
	\midrule
	\multirow{2}{*}{$L={2}$} & p-value &0:\textcolor{red}{\textbf{100}}:1 & 0:\textcolor{red}{\textbf{96}}:5 & 0:\textcolor{red}{\textbf{100}}:1 & 0:\textcolor{red}{\textbf{99}}:2 \\
	 & metric &37:0:\textbf{64} & 15:0:\textbf{86} & \textbf{69}:0:32 & 42:0:\textbf{59} \\
	\midrule
	\multirow{2}{*}{$L={3}$} & p-value &0:\textcolor{red}{\textbf{101}}:0 & 5:\textcolor{red}{\textbf{96}}:0 & 1:\textcolor{red}{\textbf{99}}:1 & 0:\textcolor{red}{\textbf{97}}:4 \\
	 & metric &\textbf{84}:0:17 & \textbf{71}:0:30 & 45:0:\textbf{56} & 14:0:\textbf{87} \\
	\bottomrule
	\end{tabular}}
	\caption{\textsc{Chem} on HNHN}
	\label{tab:p_metric_val_HNHN_Esol_Chem}
        \end{subtable}

        \vspace{0.32cm} % adjust vertical space between tables
        
        \begin{subtable}
            \centering
	\resizebox{.99\textwidth}{!}{\begin{tabular}{lccccc}
	\toprule
	\multicolumn{2}{c}{\textbf{learning rate}} & \multicolumn{2}{c}{$10^{-2}$} & \multicolumn{2}{c}{$10^{-3}$} \\
	\multicolumn{2}{c}{\textbf{hidden size}} & $h=32$ & $h=64$ & $h=32$ & $h=64$ \\
	\midrule
	\multirow{2}{*}{$L={1}$} & p-value &1:\textcolor{red}{\textbf{100}}:0 & 0:\textcolor{red}{\textbf{99}}:2 & 0:\textcolor{red}{\textbf{93}}:8 & 0:\textcolor{red}{\textbf{99}}:2 \\
	 & metric &\textbf{90}:0:11 & 20:0:\textbf{81} & 12:0:\textbf{89} & 36:0:\textbf{65} \\
	\midrule
	\multirow{2}{*}{$L={2}$} & p-value &1:\textcolor{red}{\textbf{99}}:1 & 0:\textcolor{red}{\textbf{97}}:4 & 7:\textcolor{red}{\textbf{94}}:0 & 1:\textcolor{red}{\textbf{100}}:0 \\
	 & metric &\textbf{55}:0:46 & 15:0:\textbf{86} & \textbf{100}:0:1 & \textbf{74}:0:27 \\
	\midrule
	\multirow{2}{*}{$L={3}$} & p-value &0:\textcolor{red}{\textbf{101}}:0 & 0:\textcolor{red}{\textbf{88}}:13 & 0:\textcolor{red}{\textbf{100}}:1 & 10:\textcolor{red}{\textbf{91}}:0 \\
	 & metric &\textbf{65}:0:36 & 8:0:\textbf{93} & 31:0:\textbf{70} & \textbf{87}:0:14 \\
	\bottomrule
	\end{tabular}}
	\caption{\textsc{H2g} on HNHN}
	\label{tab:p_metric_val_HNHN_Esol_H2g}
        \end{subtable}
    \end{minipage}
    \caption{Performance comparison on accuracy with p-value $<0.05$ and metric value on \textsc{Esol}. For each triplet $a$:$b$:$c$, $a, b, c$ denote the number of times \textsc{GraphBPE} is \textcolor{red}{\textbf{statistically}}/\textbf{numerically} better/the same/worse compared with hypergraphs constructed by \textsc{Method} on Model (e.g., ``\textsc{Centroud} on HyperConv'' means comparing \textsc{GraphBPE} with \textsc{Centroid} on the HyperConv model).}
    \label{tab: p_metric_3hgnns_on_esol}
\end{table}
\FloatBarrier  % Prevent floats from moving past this point
%%%%%%%%%%%%%%%%%%%%%%%%%%%%%%%%%%%%%%%%%
\FloatBarrier  % Prevent floats from moving past this point
\FloatBarrier  % Prevent floats from moving past this point
\newpage

\subsection{\textsc{Lipophilicity}}\label{app: lipo_result}
\begin{figure}[H]
\centering
\includegraphics[width=0.9\linewidth, height=0.59\textwidth]{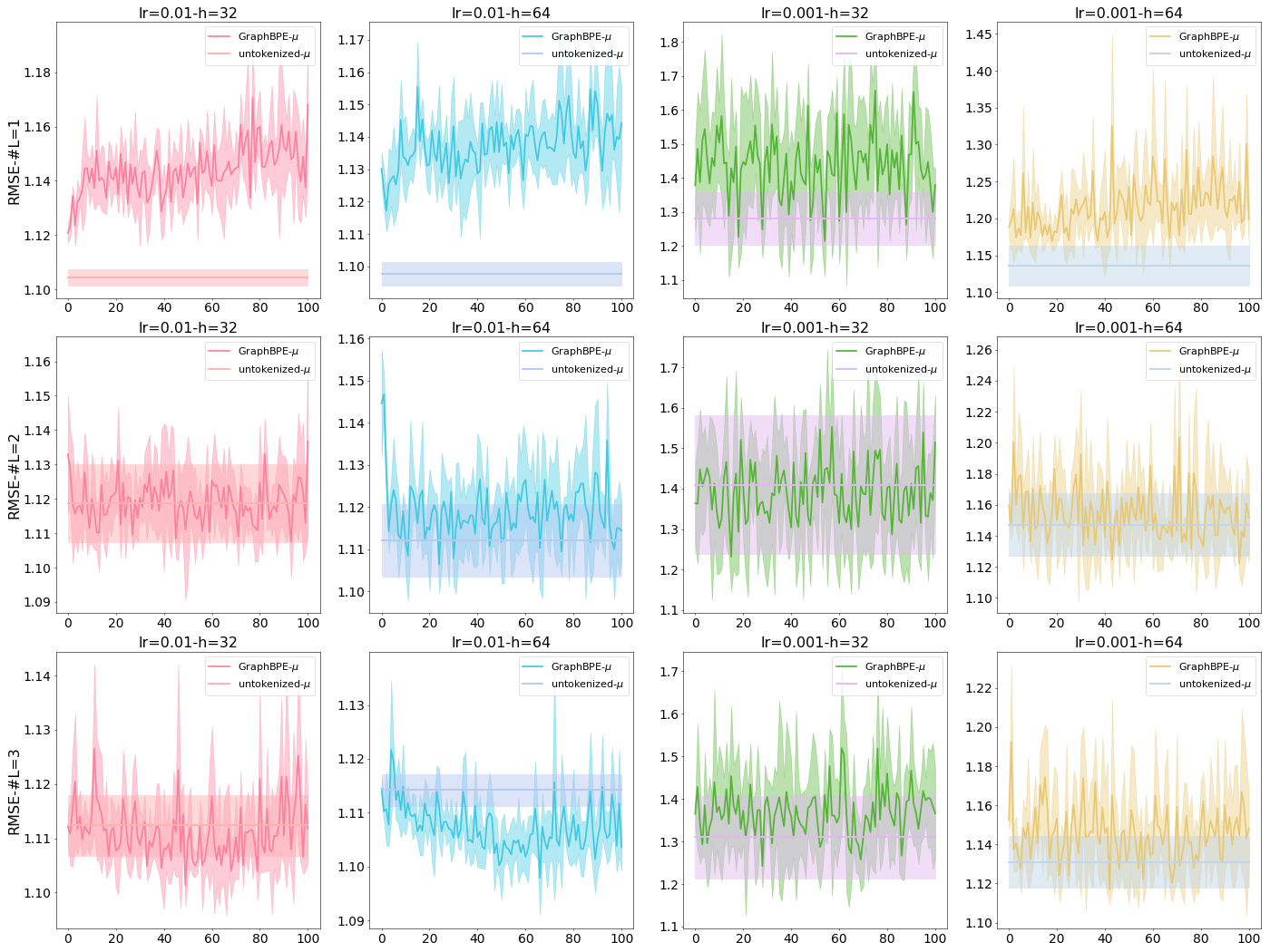}
\caption{Results of GCN on \textsc{Lipophilicity}, with \textbf{RMSE} the \textit{lower} the better} 
\label{fig:lipo_gcn}
\end{figure}

\begin{figure}[H]
\centering
\includegraphics[width=0.9\linewidth, height=0.59\textwidth]{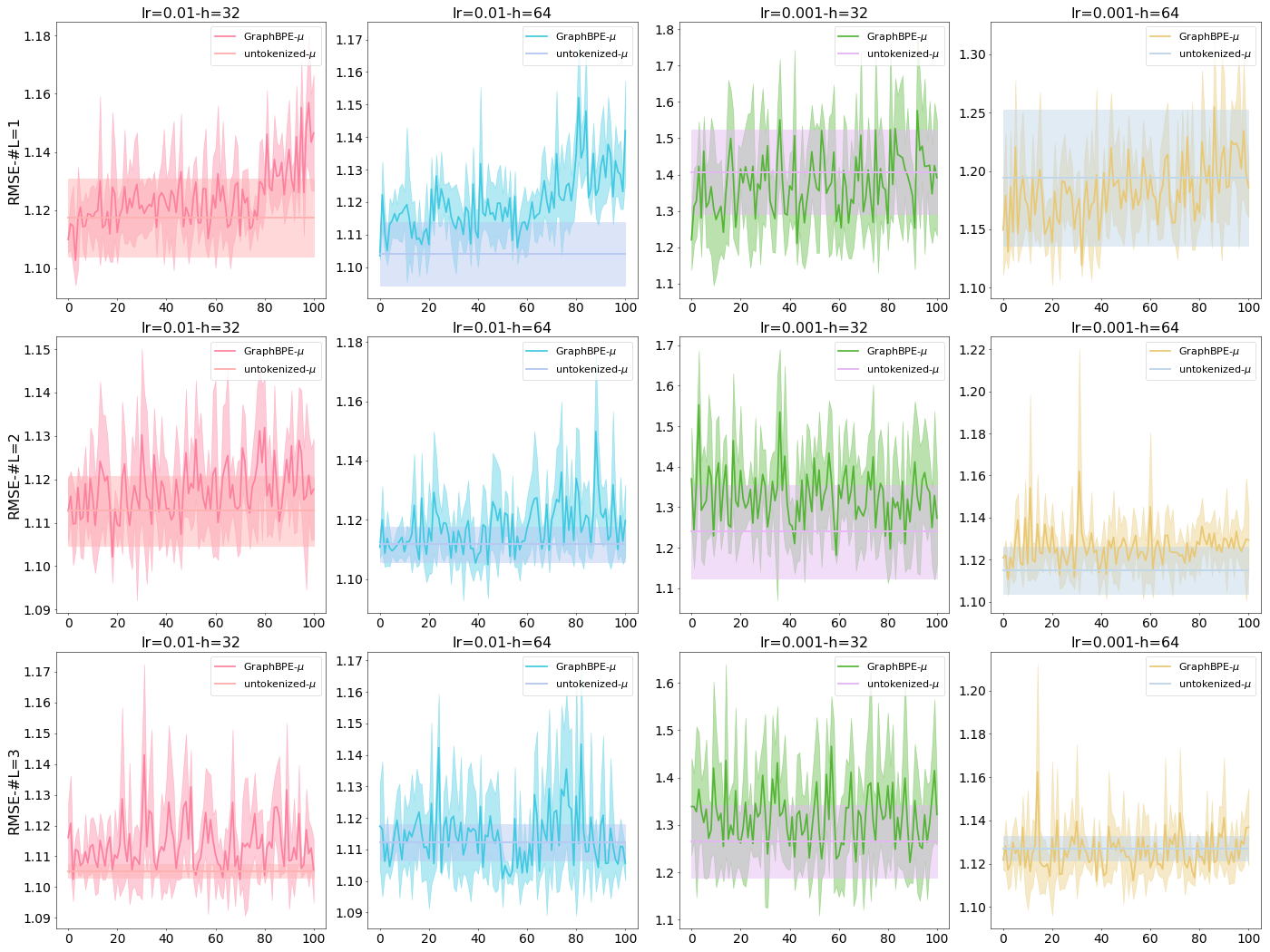}
\caption{Results of GAT on \textsc{Lipophilicity}, with \textbf{RMSE} the \textit{lower} the better} 
\label{fig:lipo_gat}
\end{figure}

\begin{figure}[H]
\centering
\includegraphics[width=0.9\linewidth, height=0.59\textwidth]{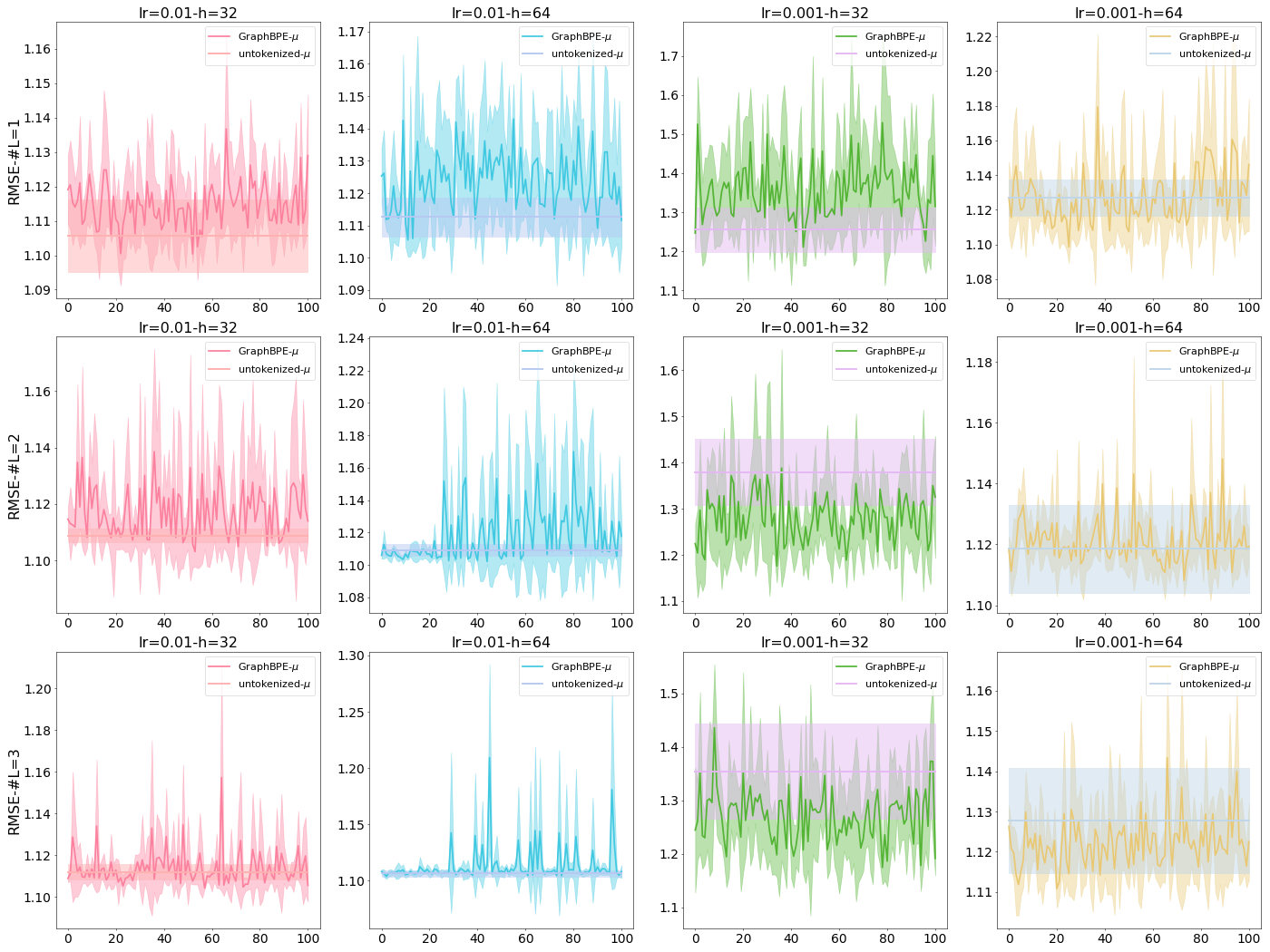}
\caption{Results of GIN on \textsc{Lipophilicity}, with \textbf{RMSE} the \textit{lower} the better} 
\label{fig:lipo_gin}
\end{figure}

\begin{figure}[H]
\centering
\includegraphics[width=0.9\linewidth, height=0.59\textwidth]{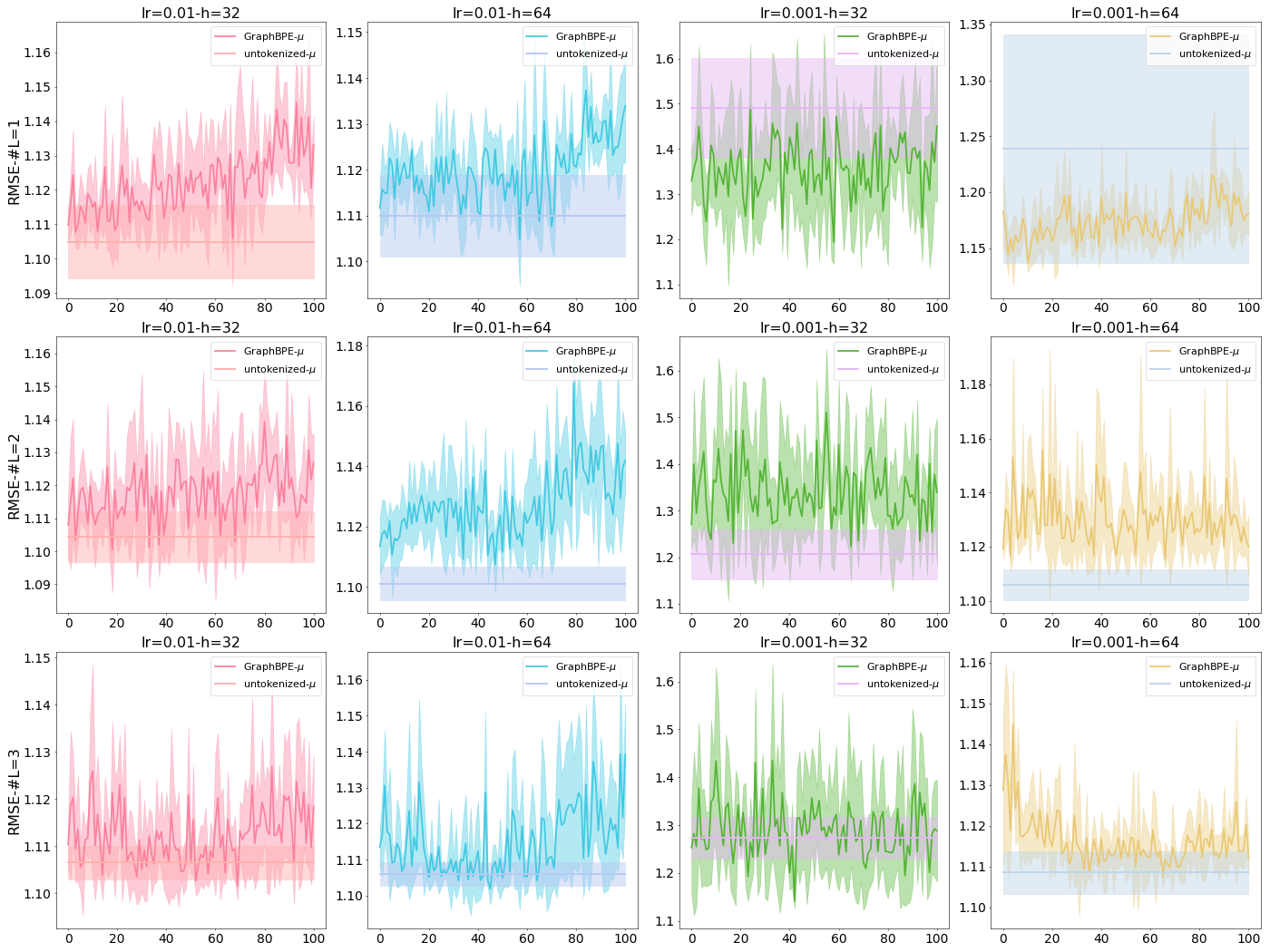}
\caption{Results of GraphSAGE on \textsc{Lipophilicity}, with \textbf{RMSE} the \textit{lower} the better} 
\label{fig:lipo_graphsage}
\end{figure}

For GNNs, we include the performance comparison results in Table~\ref{tab: p_metric_4gnns_on_lipophilicity}, and the visualization over different tokenization steps in Figure~\ref{fig:lipo_gcn}, ~\ref{fig:lipo_gat}, ~\ref{fig:lipo_gin}, and ~\ref{fig:lipo_graphsage} for GCN, GAT, GIN, and GraphSAGE.

\begin{figure}[H]
\centering
\includegraphics[width=0.9\linewidth, height=0.52\textwidth]{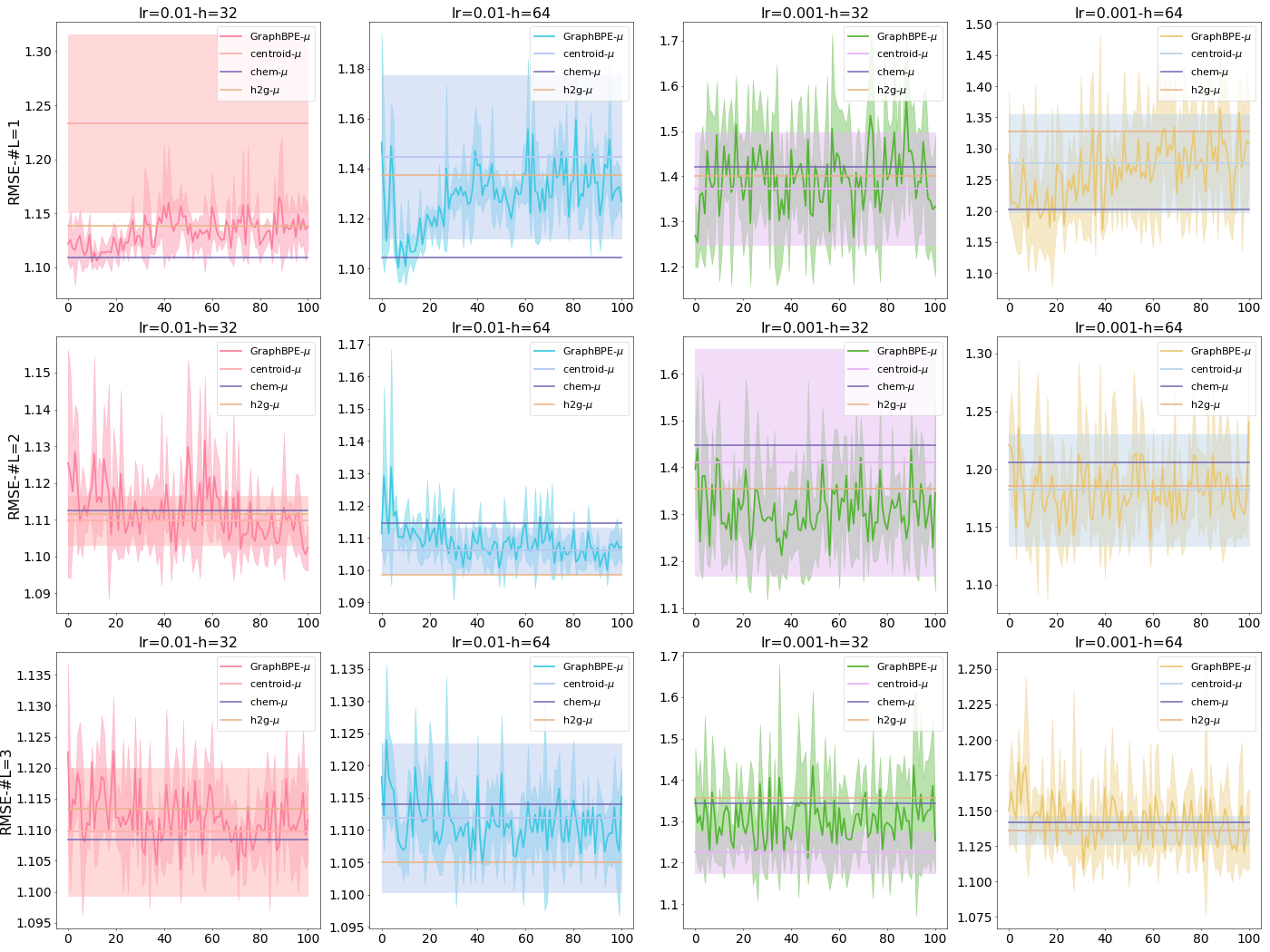}
\caption{Results of HyperConv on \textsc{Lipophilicity}, with \textbf{RMSE} the \textit{lower} the better} 
\label{fig:lipo_HyperConv}
\end{figure}

\begin{figure}[H]
\centering
\includegraphics[width=0.9\linewidth, height=0.59\textwidth]{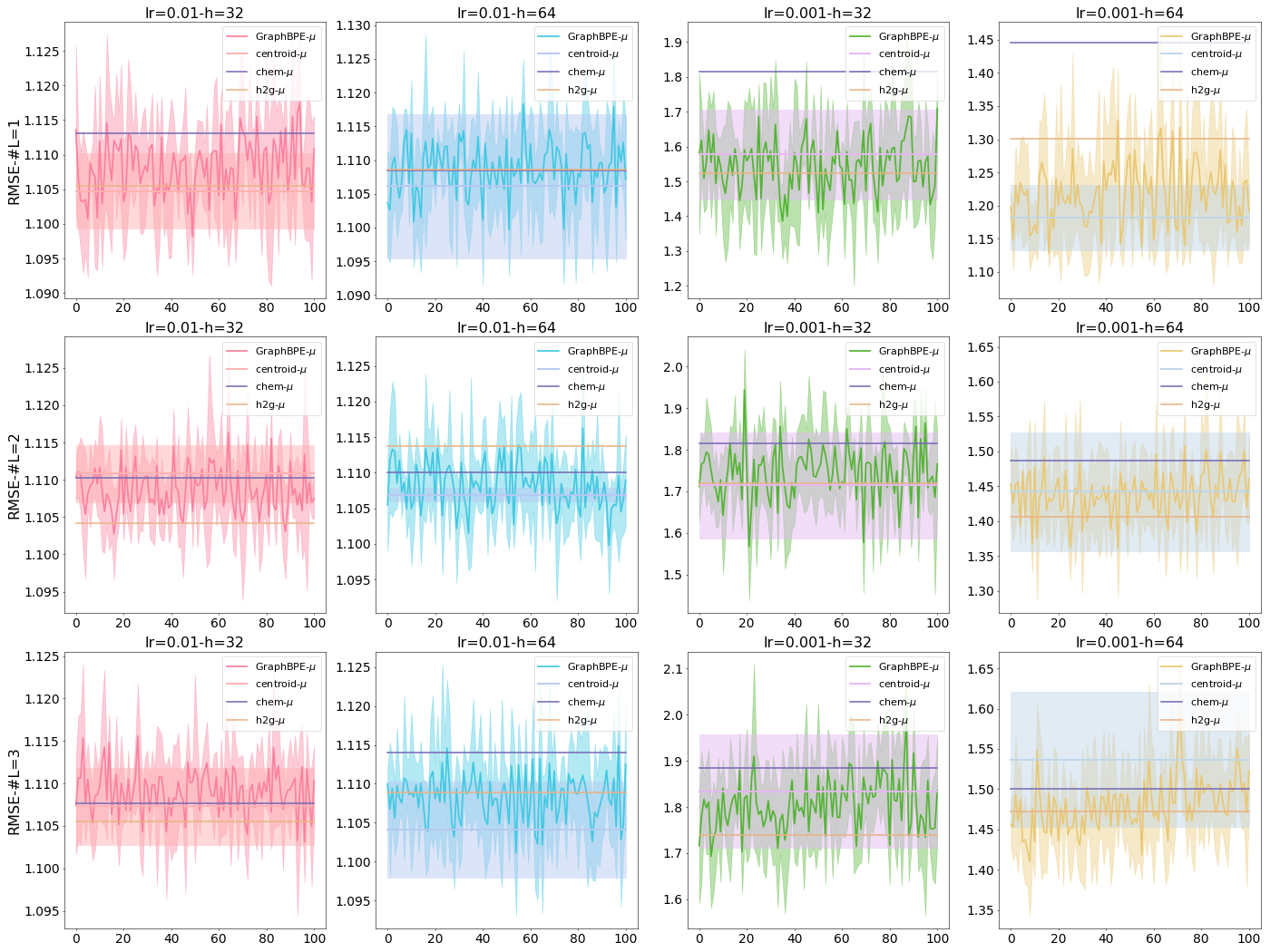}
\caption{Results of HGNN++ on \textsc{Lipophilicity}, with \textbf{RMSE} the \textit{lower} the better} 
\label{fig:lipo_HGNNP}
\end{figure}

\begin{figure}[H]
\centering
\includegraphics[width=0.9\linewidth, height=0.59\textwidth]{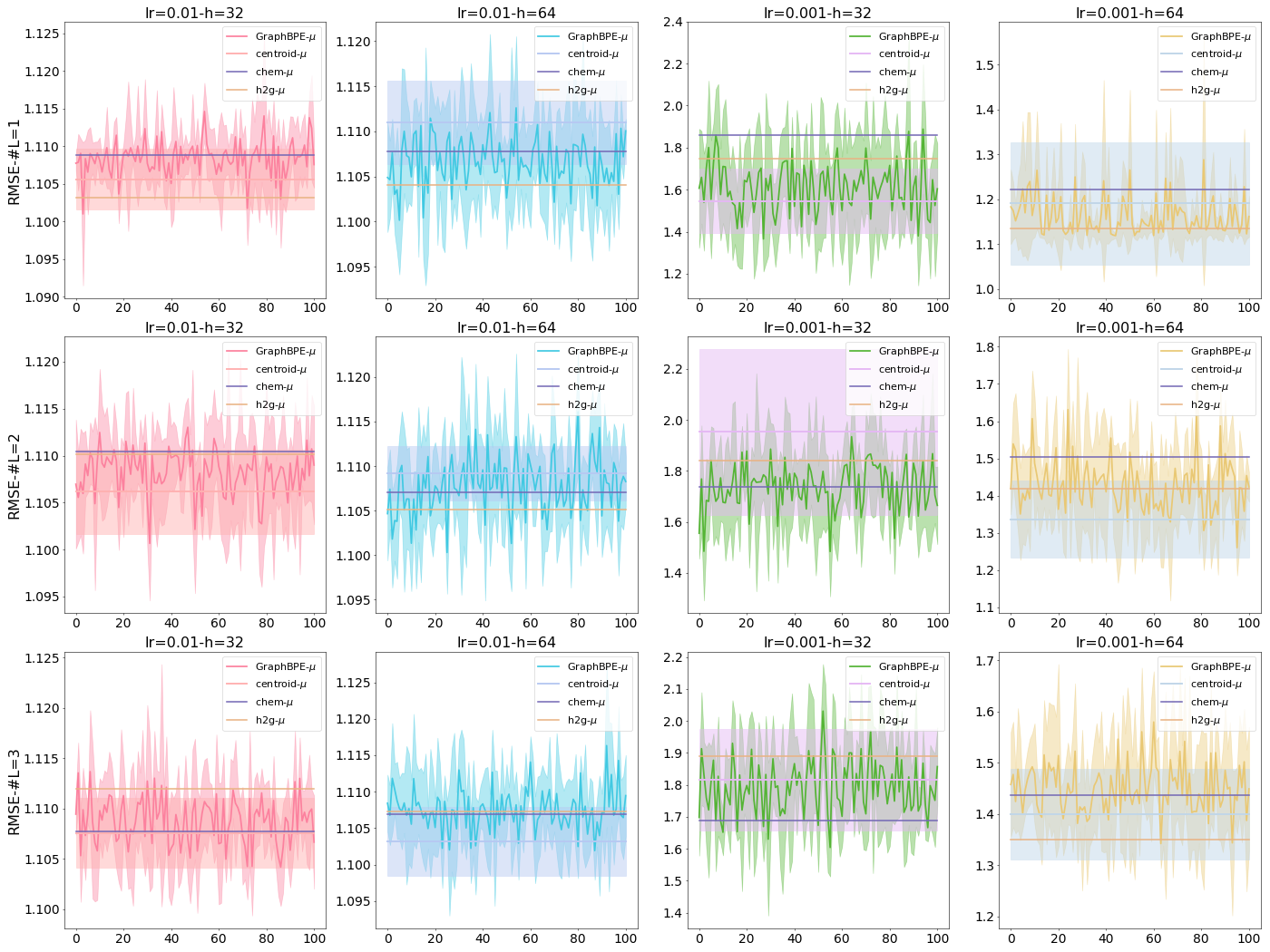}
\caption{Results of HNHN on \textsc{Lipophilicity}, with \textbf{RMSE} the \textit{lower} the better} 
\label{fig:lipo_HNHN}
\end{figure}

For HyperGNNs, we include the performance comparison results in Table~\ref{tab: p_metric_3hgnns_on_lipophilicity}, and the visualization over different tokenization steps in Figure~\ref{fig:lipo_HyperConv}, ~\ref{fig:lipo_HGNNP}, and ~\ref{fig:lipo_HNHN} for HyperConv, HGNN++, and HNHN.
\begin{table}[h]
    \centering
    \begin{minipage}[b]{0.45\linewidth}
        \centering
        \begin{subtable} % {.5\textwidth}
            \centering
	\resizebox{.99\textwidth}{!}{\begin{tabular}{lccccc}
	\toprule
	\multicolumn{2}{c}{\textbf{learning rate}} & \multicolumn{2}{c}{$10^{-2}$} & \multicolumn{2}{c}{$10^{-3}$} \\
	\multicolumn{2}{c}{\textbf{hidden size}} & $h=32$ & $h=64$ & $h=32$ & $h=64$ \\
	\midrule
	\multirow{2}{*}{$L={1}$} & p-value &0:0:\textcolor{red}{\textbf{101}} & 0:0:\textcolor{red}{\textbf{101}} & 0:\textcolor{red}{\textbf{83}}:18 & 0:37:\textcolor{red}{\textbf{64}} \\
	 & metric &0:0:\textbf{101} & 0:0:\textbf{101} & 6:0:\textbf{95} & 0:0:\textbf{101} \\
	\midrule
	\multirow{2}{*}{$L={2}$} & p-value &0:\textcolor{red}{\textbf{101}}:0 & 0:\textcolor{red}{\textbf{94}}:7 & 0:\textcolor{red}{\textbf{101}}:0 & 0:\textcolor{red}{\textbf{100}}:1 \\
	 & metric &\textbf{51}:0:50 & 11:0:\textbf{90} & \textbf{62}:0:39 & 39:0:\textbf{62} \\
	\midrule
	\multirow{2}{*}{$L={3}$} & p-value &5:\textcolor{red}{\textbf{96}}:0 & 50:\textcolor{red}{\textbf{51}}:0 & 0:\textcolor{red}{\textbf{97}}:4 & 0:\textcolor{red}{\textbf{93}}:8 \\
	 & metric &\textbf{72}:0:29 & \textbf{95}:0:6 & 16:0:\textbf{85} & 13:0:\textbf{88} \\
	\bottomrule
	\end{tabular}}
	\caption{Comparison with p-/metric value of GCN}
	\label{tab:p_metric_val_GCN_Lipophilicity}
        \end{subtable}
        
        \vspace{0.3cm} % adjust vertical space between tables
        
        \begin{subtable} %{.5\textwidth}
            \centering
	\resizebox{.99\textwidth}{!}{\begin{tabular}{lccccc}
	\toprule
	\multicolumn{2}{c}{\textbf{learning rate}} & \multicolumn{2}{c}{$10^{-2}$} & \multicolumn{2}{c}{$10^{-3}$} \\
	\multicolumn{2}{c}{\textbf{hidden size}} & $h=32$ & $h=64$ & $h=32$ & $h=64$ \\
	\midrule
	\multirow{2}{*}{$L={1}$} & p-value &0:\textcolor{red}{\textbf{98}}:3 & 0:\textcolor{red}{\textbf{93}}:8 & 0:\textcolor{red}{\textbf{88}}:13 & 5:\textcolor{red}{\textbf{95}}:1 \\
	 & metric &4:0:\textbf{97} & 10:0:\textbf{91} & 4:0:\textbf{97} & \textbf{57}:0:44 \\
	\midrule
	\multirow{2}{*}{$L={2}$} & p-value &0:\textcolor{red}{\textbf{99}}:2 & 2:\textcolor{red}{\textbf{99}}:0 & 34:\textcolor{red}{\textbf{67}}:0 & 0:\textcolor{red}{\textbf{100}}:1 \\
	 & metric &19:0:\textbf{82} & 47:0:\textbf{54} & \textbf{100}:0:1 & 40:0:\textbf{61} \\
	\midrule
	\multirow{2}{*}{$L={3}$} & p-value &6:\textcolor{red}{\textbf{95}}:0 & 0:\textcolor{red}{\textbf{100}}:1 & 15:\textcolor{red}{\textbf{86}}:0 & 1:\textcolor{red}{\textbf{100}}:0 \\
	 & metric &\textbf{52}:0:49 & 18:0:\textbf{83} & \textbf{98}:0:3 & \textbf{86}:0:15 \\
	\bottomrule
	\end{tabular}}
	\caption{Comparison with p-/metric value of GIN}
	\label{tab:p_metric_val_GIN_Lipophilicity}
        \end{subtable}
    \end{minipage}
    \hspace{0.5cm}
    \begin{minipage}[b]{0.45\linewidth}
        \centering
        \begin{subtable} %{.5\textwidth}
            \centering
	\resizebox{.99\textwidth}{!}{\begin{tabular}{lccccc}
	\toprule
	\multicolumn{2}{c}{\textbf{learning rate}} & \multicolumn{2}{c}{$10^{-2}$} & \multicolumn{2}{c}{$10^{-3}$} \\
	\multicolumn{2}{c}{\textbf{hidden size}} & $h=32$ & $h=64$ & $h=32$ & $h=64$ \\
	\midrule
	\multirow{2}{*}{$L={1}$} & p-value &0:\textcolor{red}{\textbf{95}}:6 & 0:\textcolor{red}{\textbf{62}}:39 & 4:\textcolor{red}{\textbf{97}}:0 & 0:\textcolor{red}{\textbf{101}}:0 \\
	 & metric &25:0:\textbf{76} & 1:0:\textbf{100} & \textbf{66}:0:35 & \textbf{72}:0:29 \\
	\midrule
	\multirow{2}{*}{$L={2}$} & p-value &0:\textcolor{red}{\textbf{98}}:3 & 0:\textcolor{red}{\textbf{97}}:4 & 0:\textcolor{red}{\textbf{91}}:10 & 0:\textcolor{red}{\textbf{97}}:4 \\
	 & metric &19:0:\textbf{82} & 27:0:\textbf{74} & 6:0:\textbf{95} & 5:0:\textbf{96} \\
	\midrule
	\multirow{2}{*}{$L={3}$} & p-value &0:\textcolor{red}{\textbf{91}}:10 & 5:\textcolor{red}{\textbf{94}}:2 & 0:\textcolor{red}{\textbf{99}}:2 & 8:\textcolor{red}{\textbf{93}}:0 \\
	 & metric &5:0:\textbf{96} & 47:0:\textbf{54} & 21:0:\textbf{80} & \textbf{61}:0:40 \\
	\bottomrule
	\end{tabular}}
	\caption{Comparison with p-/metric value of GAT}
	\label{tab:p_metric_val_GAT_Lipophilicity}
        \end{subtable}
        
        \vspace{0.3cm} % adjust vertical space between tables
        
        \begin{subtable} %{.5\textwidth}
            \centering
	\resizebox{.99\textwidth}{!}{\begin{tabular}{lccccc}
	\toprule
	\multicolumn{2}{c}{\textbf{learning rate}} & \multicolumn{2}{c}{$10^{-2}$} & \multicolumn{2}{c}{$10^{-3}$} \\
	\multicolumn{2}{c}{\textbf{hidden size}} & $h=32$ & $h=64$ & $h=32$ & $h=64$ \\
	\midrule
	\multirow{2}{*}{$L={1}$} & p-value &0:\textcolor{red}{\textbf{68}}:33 & 0:\textcolor{red}{\textbf{84}}:17 & 23:\textcolor{red}{\textbf{78}}:0 & 0:\textcolor{red}{\textbf{101}}:0 \\
	 & metric &0:0:\textbf{101} & 4:0:\textbf{97} & \textbf{100}:0:1 & \textbf{101}:0:0 \\
	\midrule
	\multirow{2}{*}{$L={2}$} & p-value &0:\textcolor{red}{\textbf{86}}:15 & 0:26:\textcolor{red}{\textbf{75}} & 0:\textcolor{red}{\textbf{72}}:29 & 0:30:\textcolor{red}{\textbf{71}} \\
	 & metric &5:0:\textbf{96} & 0:0:\textbf{101} & 0:0:\textbf{101} & 0:0:\textbf{101} \\
	\midrule
	\multirow{2}{*}{$L={3}$} & p-value &0:\textcolor{red}{\textbf{96}}:5 & 0:\textcolor{red}{\textbf{92}}:9 & 2:\textcolor{red}{\textbf{98}}:1 & 0:\textcolor{red}{\textbf{85}}:16 \\
	 & metric &18:0:\textbf{83} & 15:0:\textbf{86} & 40:0:\textbf{61} & 1:0:\textbf{100} \\
	\bottomrule
	\end{tabular}}
	\caption{Comparison with p-/metric value of GraphSAGE}
	\label{tab:p_metric_val_GraphSAGE_Lipophilicity}
        \end{subtable}
    \end{minipage}
    \caption{Performance comparison on accuracy with p-value $<0.05$ and metric value on \textsc{Lipophilicity}. For each triplet $a$:$b$:$c$, $a, b, c$ denote the number of times \textsc{GraphBPE} is \textcolor{red}{\textbf{statistically}}/\textbf{numerically} better/the same/worse compared with (untokenized) simple graph.}
    \label{tab: p_metric_4gnns_on_lipophilicity}
\end{table}
\newpage
\FloatBarrier  % Prevent floats from moving past this point
%%%%%%%%%%%%%%%%%%%%%%%%%%%%%%%%%%%%%%%%%
\FloatBarrier  % Prevent floats from moving past this point
%%%%%%%%%%%%%%%%%%%%%%%%%%%%%%%%%%%%%%%%%
\FloatBarrier  % Prevent floats from moving past this point
%%%%%%%%%%%%%%%%%%%%%%%%%%%%%%%%%%%%%%%%%
\FloatBarrier  % Prevent floats from moving past this point
\newpage
\begin{table}[h]
    \centering
    \begin{minipage}[b]{0.32\linewidth}
        \centering
        \begin{subtable}
            \centering
	\resizebox{.99\textwidth}{!}{\begin{tabular}{lccccc}
	\toprule
	\multicolumn{2}{c}{\textbf{learning rate}} & \multicolumn{2}{c}{$10^{-2}$} & \multicolumn{2}{c}{$10^{-3}$} \\
	\multicolumn{2}{c}{\textbf{hidden size}} & $h=32$ & $h=64$ & $h=32$ & $h=64$ \\
	\midrule
	\multirow{2}{*}{$L={1}$} & p-value &22:\textcolor{red}{\textbf{79}}:0 & 2:\textcolor{red}{\textbf{99}}:0 & 0:\textcolor{red}{\textbf{99}}:2 & 2:\textcolor{red}{\textbf{99}}:0 \\
	 & metric &\textbf{101}:0:0 & \textbf{91}:0:10 & 34:0:\textbf{67} & \textbf{61}:0:40 \\
	\midrule
	\multirow{2}{*}{$L={2}$} & p-value &1:\textcolor{red}{\textbf{97}}:3 & 0:\textcolor{red}{\textbf{98}}:3 & 0:\textcolor{red}{\textbf{101}}:0 & 0:\textcolor{red}{\textbf{100}}:1 \\
	 & metric &41:0:\textbf{60} & 25:0:\textbf{76} & \textbf{95}:0:6 & \textbf{56}:0:45 \\
	\midrule
	\multirow{2}{*}{$L={3}$} & p-value &0:\textcolor{red}{\textbf{101}}:0 & 0:\textcolor{red}{\textbf{101}}:0 & 0:\textcolor{red}{\textbf{90}}:11 & 4:\textcolor{red}{\textbf{93}}:4 \\
	 & metric &27:0:\textbf{74} & \textbf{67}:0:34 & 2:0:\textbf{99} & 35:0:\textbf{66} \\
	\bottomrule
	\end{tabular}}
	\caption{\textsc{Centroid} on HyperConv}
	\label{tab:p_metric_val_HyperConv_Lipophilicity_Centroid}
        \end{subtable}
        
        \vspace{0.25cm} % adjust vertical space between tables
        
        \begin{subtable}
            \centering
	\resizebox{.99\textwidth}{!}{\begin{tabular}{lccccc}
	\toprule
	\multicolumn{2}{c}{\textbf{learning rate}} & \multicolumn{2}{c}{$10^{-2}$} & \multicolumn{2}{c}{$10^{-3}$} \\
	\multicolumn{2}{c}{\textbf{hidden size}} & $h=32$ & $h=64$ & $h=32$ & $h=64$ \\
	\midrule
	\multirow{2}{*}{$L={1}$} & p-value &0:\textcolor{red}{\textbf{75}}:26 & 0:27:\textcolor{red}{\textbf{74}} & 0:\textcolor{red}{\textbf{101}}:0 & 0:\textcolor{red}{\textbf{93}}:8 \\
	 & metric &2:0:\textbf{99} & 2:0:\textbf{99} & \textbf{67}:0:34 & 10:0:\textbf{91} \\
	\midrule
	\multirow{2}{*}{$L={2}$} & p-value &1:\textcolor{red}{\textbf{100}}:0 & 9:\textcolor{red}{\textbf{92}}:0 & 8:\textcolor{red}{\textbf{93}}:0 & 11:\textcolor{red}{\textbf{90}}:0 \\
	 & metric &\textbf{60}:0:41 & \textbf{91}:0:10 & \textbf{101}:0:0 & \textbf{86}:0:15 \\
	\midrule
	\multirow{2}{*}{$L={3}$} & p-value &1:\textcolor{red}{\textbf{88}}:12 & 7:\textcolor{red}{\textbf{94}}:0 & 2:\textcolor{red}{\textbf{99}}:0 & 0:\textcolor{red}{\textbf{101}}:0 \\
	 & metric &17:0:\textbf{84} & \textbf{84}:0:17 & \textbf{81}:0:20 & \textbf{52}:0:49 \\
	\bottomrule
	\end{tabular}}
	\caption{\textsc{Chem} on HyperConv}
	\label{tab:p_metric_val_HyperConv_Lipophilicity_Chem}
        \end{subtable}

        \vspace{0.25cm} % adjust vertical space between tables
        
        \begin{subtable}
            \centering
	\resizebox{.99\textwidth}{!}{\begin{tabular}{lccccc}
	\toprule
	\multicolumn{2}{c}{\textbf{learning rate}} & \multicolumn{2}{c}{$10^{-2}$} & \multicolumn{2}{c}{$10^{-3}$} \\
	\multicolumn{2}{c}{\textbf{hidden size}} & $h=32$ & $h=64$ & $h=32$ & $h=64$ \\
	\midrule
	\multirow{2}{*}{$L={1}$} & p-value &0:\textcolor{red}{\textbf{101}}:0 & 13:\textcolor{red}{\textbf{88}}:0 & 2:\textcolor{red}{\textbf{98}}:1 & 0:\textcolor{red}{\textbf{101}}:0 \\
	 & metric &\textbf{67}:0:34 & \textbf{81}:0:20 & \textbf{57}:0:44 & \textbf{93}:0:8 \\
	\midrule
	\multirow{2}{*}{$L={2}$} & p-value &0:\textcolor{red}{\textbf{100}}:1 & 0:45:\textcolor{red}{\textbf{56}} & 1:\textcolor{red}{\textbf{100}}:0 & 1:\textcolor{red}{\textbf{99}}:1 \\
	 & metric &\textbf{52}:0:49 & 0:0:\textbf{101} & \textbf{84}:0:17 & \textbf{62}:0:39 \\
	\midrule
	\multirow{2}{*}{$L={3}$} & p-value &1:\textcolor{red}{\textbf{100}}:0 & 0:\textcolor{red}{\textbf{94}}:7 & 0:\textcolor{red}{\textbf{101}}:0 & 1:\textcolor{red}{\textbf{97}}:3 \\
	 & metric &\textbf{70}:0:31 & 2:0:\textbf{99} & \textbf{91}:0:10 & 35:0:\textbf{66} \\
	\bottomrule
	\end{tabular}}
	\caption{\textsc{H2g} on HyperConv}
	\label{tab:p_metric_val_HyperConv_Lipophilicity_H2g}
        \end{subtable}
    \end{minipage}
    \hfill
    \begin{minipage}[b]{0.32\linewidth}
        \centering
        \begin{subtable}
            \centering
	\resizebox{.99\textwidth}{!}{\begin{tabular}{lccccc}
	\toprule
	\multicolumn{2}{c}{\textbf{learning rate}} & \multicolumn{2}{c}{$10^{-2}$} & \multicolumn{2}{c}{$10^{-3}$} \\
	\multicolumn{2}{c}{\textbf{hidden size}} & $h=32$ & $h=64$ & $h=32$ & $h=64$ \\
	\midrule
	\multirow{2}{*}{$L={1}$} & p-value &0:\textcolor{red}{\textbf{94}}:7 & 0:\textcolor{red}{\textbf{100}}:1 & 1:\textcolor{red}{\textbf{100}}:0 & 0:\textcolor{red}{\textbf{99}}:2 \\
	 & metric &24:0:\textbf{77} & 22:0:\textbf{79} & \textbf{67}:0:34 & 24:0:\textbf{77} \\
	\midrule
	\multirow{2}{*}{$L={2}$} & p-value &4:\textcolor{red}{\textbf{97}}:0 & 1:\textcolor{red}{\textbf{94}}:6 & 1:\textcolor{red}{\textbf{98}}:2 & 0:\textcolor{red}{\textbf{101}}:0 \\
	 & metric &\textbf{79}:0:22 & 38:0:\textbf{63} & 30:0:\textbf{71} & \textbf{56}:0:45 \\
	\midrule
	\multirow{2}{*}{$L={3}$} & p-value &0:\textcolor{red}{\textbf{99}}:2 & 0:\textcolor{red}{\textbf{99}}:2 & 1:\textcolor{red}{\textbf{99}}:1 & 2:\textcolor{red}{\textbf{99}}:0 \\
	 & metric &23:0:\textbf{78} & 8:0:\textbf{93} & \textbf{73}:0:28 & \textbf{97}:0:4 \\
	\bottomrule
	\end{tabular}}
	\caption{\textsc{Centroid} on HGNN++}
	\label{tab:p_metric_val_HGNN++_Lipophilicity_Centroid}
        \end{subtable}
        
        \vspace{0.32cm} % adjust vertical space between tables
        
        \begin{subtable}
            \centering
	\resizebox{.99\textwidth}{!}{\begin{tabular}{lccccc}
	\toprule
	\multicolumn{2}{c}{\textbf{learning rate}} & \multicolumn{2}{c}{$10^{-2}$} & \multicolumn{2}{c}{$10^{-3}$} \\
	\multicolumn{2}{c}{\textbf{hidden size}} & $h=32$ & $h=64$ & $h=32$ & $h=64$ \\
	\midrule
	\multirow{2}{*}{$L={1}$} & p-value &8:\textcolor{red}{\textbf{93}}:0 & 2:\textcolor{red}{\textbf{98}}:1 & 48:\textcolor{red}{\textbf{53}}:0 & \textcolor{red}{\textbf{62}}:39:0 \\
	 & metric &\textbf{90}:0:11 & 50:0:\textbf{51} & \textbf{101}:0:0 & \textbf{101}:0:0 \\
	\midrule
	\multirow{2}{*}{$L={2}$} & p-value &0:\textcolor{red}{\textbf{101}}:0 & 8:\textcolor{red}{\textbf{93}}:0 & 8:\textcolor{red}{\textbf{93}}:0 & 8:\textcolor{red}{\textbf{93}}:0 \\
	 & metric &\textbf{71}:0:30 & \textbf{71}:0:30 & \textbf{92}:0:9 & \textbf{95}:0:6 \\
	\midrule
	\multirow{2}{*}{$L={3}$} & p-value &0:\textcolor{red}{\textbf{99}}:2 & 19:\textcolor{red}{\textbf{82}}:0 & 17:\textcolor{red}{\textbf{84}}:0 & 5:\textcolor{red}{\textbf{96}}:0 \\
	 & metric &26:0:\textbf{75} & \textbf{99}:0:2 & \textbf{93}:0:8 & \textbf{80}:0:21 \\
	\bottomrule
	\end{tabular}}
	\caption{\textsc{Chem} on HGNN++}
	\label{tab:p_metric_val_HGNN++_Lipophilicity_Chem}
        \end{subtable}

        \vspace{0.32cm} % adjust vertical space between tables
        
        \begin{subtable}
            \centering
	\resizebox{.99\textwidth}{!}{\begin{tabular}{lccccc}
	\toprule
	\multicolumn{2}{c}{\textbf{learning rate}} & \multicolumn{2}{c}{$10^{-2}$} & \multicolumn{2}{c}{$10^{-3}$} \\
	\multicolumn{2}{c}{\textbf{hidden size}} & $h=32$ & $h=64$ & $h=32$ & $h=64$ \\
	\midrule
	\multirow{2}{*}{$L={1}$} & p-value &0:\textcolor{red}{\textbf{100}}:1 & 0:\textcolor{red}{\textbf{101}}:0 & 0:\textcolor{red}{\textbf{98}}:3 & 2:\textcolor{red}{\textbf{99}}:0 \\
	 & metric &28:0:\textbf{73} & \textbf{51}:0:50 & 40:0:\textbf{61} & \textbf{96}:0:5 \\
	\midrule
	\multirow{2}{*}{$L={2}$} & p-value &0:\textcolor{red}{\textbf{93}}:8 & 32:\textcolor{red}{\textbf{69}}:0 & 1:\textcolor{red}{\textbf{97}}:3 & 0:\textcolor{red}{\textbf{96}}:5 \\
	 & metric &2:0:\textbf{99} & \textbf{99}:0:2 & 32:0:\textbf{69} & 15:0:\textbf{86} \\
	\midrule
	\multirow{2}{*}{$L={3}$} & p-value &0:\textcolor{red}{\textbf{88}}:13 & 2:\textcolor{red}{\textbf{99}}:0 & 0:\textcolor{red}{\textbf{98}}:3 & 1:\textcolor{red}{\textbf{98}}:2 \\
	 & metric &10:0:\textbf{91} & \textbf{59}:0:42 & 12:0:\textbf{89} & 45:0:\textbf{56} \\
	\bottomrule
	\end{tabular}}
	\caption{\textsc{H2g} on HGNN++}
	\label{tab:p_metric_val_HGNN++_Lipophilicity_H2g}
        \end{subtable}
    \end{minipage}
    \hfill
    \begin{minipage}[b]{0.32\linewidth}
        \centering
        \begin{subtable}{}
            \centering
	\resizebox{.99\textwidth}{!}{\begin{tabular}{lccccc}
	\toprule
	\multicolumn{2}{c}{\textbf{learning rate}} & \multicolumn{2}{c}{$10^{-2}$} & \multicolumn{2}{c}{$10^{-3}$} \\
	\multicolumn{2}{c}{\textbf{hidden size}} & $h=32$ & $h=64$ & $h=32$ & $h=64$ \\
	\midrule
	\multirow{2}{*}{$L={1}$} & p-value &0:\textcolor{red}{\textbf{96}}:5 & 11:\textcolor{red}{\textbf{90}}:0 & 0:\textcolor{red}{\textbf{96}}:5 & 0:\textcolor{red}{\textbf{101}}:0 \\
	 & metric &9:0:\textbf{92} & \textbf{98}:0:3 & 29:0:\textbf{72} & \textbf{78}:0:23 \\
	\midrule
	\multirow{2}{*}{$L={2}$} & p-value &0:\textcolor{red}{\textbf{99}}:2 & 2:\textcolor{red}{\textbf{99}}:0 & 3:\textcolor{red}{\textbf{98}}:0 & 0:\textcolor{red}{\textbf{89}}:12 \\
	 & metric &16:0:\textbf{85} & \textbf{66}:0:35 & \textbf{101}:0:0 & 6:0:\textbf{95} \\
	\midrule
	\multirow{2}{*}{$L={3}$} & p-value &0:\textcolor{red}{\textbf{99}}:2 & 0:\textcolor{red}{\textbf{87}}:14 & 2:\textcolor{red}{\textbf{98}}:1 & 0:\textcolor{red}{\textbf{97}}:4 \\
	 & metric &34:0:\textbf{67} & 7:0:\textbf{94} & \textbf{61}:0:40 & 10:0:\textbf{91} \\
	\bottomrule
	\end{tabular}}
	\caption{\textsc{Centroid} on HNHN}
	\label{tab:p_metric_val_HNHN_Lipophilicity_Centroid}
        \end{subtable}
        
        \vspace{0.32cm} % adjust vertical space between tables
        
        \begin{subtable}
            \centering
	\resizebox{.99\textwidth}{!}{\begin{tabular}{lccccc}
	\toprule
	\multicolumn{2}{c}{\textbf{learning rate}} & \multicolumn{2}{c}{$10^{-2}$} & \multicolumn{2}{c}{$10^{-3}$} \\
	\multicolumn{2}{c}{\textbf{hidden size}} & $h=32$ & $h=64$ & $h=32$ & $h=64$ \\
	\midrule
	\multirow{2}{*}{$L={1}$} & p-value &2:\textcolor{red}{\textbf{97}}:2 & 2:\textcolor{red}{\textbf{99}}:0 & 10:\textcolor{red}{\textbf{91}}:0 & 0:\textcolor{red}{\textbf{101}}:0 \\
	 & metric &\textbf{56}:0:45 & \textbf{64}:0:37 & \textbf{99}:0:2 & \textbf{90}:0:11 \\
	\midrule
	\multirow{2}{*}{$L={2}$} & p-value &0:\textcolor{red}{\textbf{101}}:0 & 0:\textcolor{red}{\textbf{98}}:3 & 2:\textcolor{red}{\textbf{98}}:1 & 6:\textcolor{red}{\textbf{95}}:0 \\
	 & metric &\textbf{86}:0:15 & 39:0:\textbf{62} & 46:0:\textbf{55} & \textbf{88}:0:13 \\
	\midrule
	\multirow{2}{*}{$L={3}$} & p-value &0:\textcolor{red}{\textbf{99}}:2 & 0:\textcolor{red}{\textbf{98}}:3 & 0:\textcolor{red}{\textbf{98}}:3 & 0:\textcolor{red}{\textbf{101}}:0 \\
	 & metric &34:0:\textbf{67} & 42:0:\textbf{59} & 9:0:\textbf{92} & 43:0:\textbf{58} \\
	\bottomrule
	\end{tabular}}
	\caption{\textsc{Chem} on HNHN}
	\label{tab:p_metric_val_HNHN_Lipophilicity_Chem}
        \end{subtable}

        \vspace{0.32cm} % adjust vertical space between tables
        
        \begin{subtable}
            \centering
	\resizebox{.99\textwidth}{!}{\begin{tabular}{lccccc}
	\toprule
	\multicolumn{2}{c}{\textbf{learning rate}} & \multicolumn{2}{c}{$10^{-2}$} & \multicolumn{2}{c}{$10^{-3}$} \\
	\multicolumn{2}{c}{\textbf{hidden size}} & $h=32$ & $h=64$ & $h=32$ & $h=64$ \\
	\midrule
	\multirow{2}{*}{$L={1}$} & p-value &0:\textcolor{red}{\textbf{90}}:11 & 0:\textcolor{red}{\textbf{97}}:4 & 1:\textcolor{red}{\textbf{100}}:0 & 0:\textcolor{red}{\textbf{101}}:0 \\
	 & metric &3:0:\textbf{98} & 15:0:\textbf{86} & \textbf{91}:0:10 & 29:0:\textbf{72} \\
	\midrule
	\multirow{2}{*}{$L={2}$} & p-value &2:\textcolor{red}{\textbf{99}}:0 & 0:\textcolor{red}{\textbf{95}}:6 & 9:\textcolor{red}{\textbf{91}}:1 & 0:\textcolor{red}{\textbf{100}}:1 \\
	 & metric &\textbf{85}:0:16 & 21:0:\textbf{80} & \textbf{90}:0:11 & 48:0:\textbf{53} \\
	\midrule
	\multirow{2}{*}{$L={3}$} & p-value &13:\textcolor{red}{\textbf{88}}:0 & 0:\textcolor{red}{\textbf{101}}:0 & 1:\textcolor{red}{\textbf{100}}:0 & 0:\textcolor{red}{\textbf{88}}:13 \\
	 & metric &\textbf{93}:0:8 & 48:0:\textbf{53} & \textbf{87}:0:14 & 2:0:\textbf{99} \\
	\bottomrule
	\end{tabular}}
	\caption{\textsc{H2g} on HNHN}
	\label{tab:p_metric_val_HNHN_Lipophilicity_H2g}
        \end{subtable}
    \end{minipage}
    \caption{Performance comparison on accuracy with p-value $<0.05$ and metric value on \textsc{Lipophilicity}. For each triplet $a$:$b$:$c$, $a, b, c$ denote the number of times \textsc{GraphBPE} is \textcolor{red}{\textbf{statistically}}/\textbf{numerically} better/the same/worse compared with hypergraphs constructed by \textsc{Method} on Model (e.g., ``\textsc{Centroud} on HyperConv'' means comparing \textsc{GraphBPE} with \textsc{Centroid} on the HyperConv model).}
    \label{tab: p_metric_3hgnns_on_lipophilicity}
\end{table}
\FloatBarrier  % Prevent floats from moving past this point
\FloatBarrier  % Prevent floats from moving past this point
\FloatBarrier  % Prevent floats from moving past this point

% You can add more sections or content as needed

\end{document}